%% file: main.tex
\documentclass[letterpaper, 10 pt, journal, twoside]{IEEEtran}
\usepackage{amsmath}
\usepackage{amssymb}
\usepackage{graphicx}
\usepackage{textcomp}
\usepackage{xcolor}
\usepackage{pgfplots}
\usepackage{subfigure}
\usepgfplotslibrary{groupplots}
\usepackage{tikz}
\usepackage{tikzscale}
\usepackage{hyperref}
\usepackage{cleveref}
\usepackage{float}
\usepackage{multirow}
\usepackage{caption}
\usepackage{mathtools}
\usepackage{algorithm}
\usepackage{algpseudocode} 
\usepackage{siunitx}
\usepackage{comment}
\usepackage{booktabs}
\usepackage{alphalph}

\crefname{section}{Sec.}{Secs.}
\Crefname{section}{Sec.}{Secs.}
\Crefname{figure}{Fig.}{Figs.}
\crefname{figure}{Fig.}{Figs.}
\crefname{table}{Table.}{Tables.}
\Crefname{table}{Table.}{Tables.}
\crefname{equation}{Eq.}{Eqs.}
\Crefname{equation}{Eq.}{Eqs.}

\newcommand{\ie}{\textit{i}.\textit{e}., }
\newcommand{\eg}{\textit{e}.\textit{g}., }

\DeclareMathOperator*{\argmin}{arg\,min}

\DeclarePairedDelimiter\ceil{\lceil}{\rceil}

\title{\LARGE \bf
StableLego: Stability Analysis of Block Stacking Assembly
}

\author{Ruixuan Liu$^{1}$, Kangle Deng$^{1}$, Ziwei Wang$^{1}$ and Changliu Liu$^{1}$
\thanks{
This work is in part supported by the Manufacturing Futures Institute, Carnegie Mellon University, through a grant from the Richard King Mellon Foundation.} %Use only for final RAL version
\thanks{$^{1}$Ruixuan Liu, Kangle Deng, Ziwei Wang and Changliu Liu are with Robotics Institute,
	Carnegie Mellon University,
	Pittsburgh, PA, 15213, USA.
        {\tt\small ruixuanl, kangled, ziweiwa2, cliu6@andrew.cmu.edu}}%

}

\begin{document}

\maketitle
% \thispagestyle{empty}
% \pagestyle{empty}

%%%%%%%%%%%%%%%%%%%%%%%%%%%%%%%%%%%%%%%%%%%%%%%%%%%%%%%%%%%%%%%%%%%%%%%%%%%%%%%%
\begin{abstract}
Structural stability is a necessary condition for successful construction of an assembly.
However, designing a stable assembly requires a non-trivial effort since a slight variation in the design could significantly affect the structural stability.
To address the challenge, this paper studies the stability of assembly structures, in particular, block stacking assembly.
The paper proposes a new optimization formulation, which optimizes over force balancing equations, for inferring the structural stability of 3D block stacking structures.
The proposed stability analysis is verified on hand-crafted Lego examples.
The experiment results demonstrate that the proposed method can correctly predict whether the structure is stable.
In addition, it outperforms the existing methods since it can accurately locate the weakest parts in the design, and more importantly, solve any given assembly structures.
To further validate the proposed method, we provide \textit{StableLego}: a comprehensive dataset including 50k+ 3D objects with their Lego layouts.
We test the proposed stability analysis and include the stability inference for each corresponding object in StableLego. 
Our code and the dataset are available at \url{https://github.com/intelligent-control-lab/StableLego}.

\end{abstract}
\begin{IEEEkeywords}
Assembly; Performance Evaluation and Benchmarking; Robotics and Automation in Construction
\end{IEEEkeywords}

%%%%%%%%%%%%%%%%%%%%%%%%%%%%%%%%%%%%%%%%%%%%%%%%%%%%%%%%%%%%%%%%%%%%%%%%%%%%%%%%
\section{Introduction}

\begin{figure}
\centering
\subfigure[19-level Stairs.]{\includegraphics[width=0.33\linewidth]{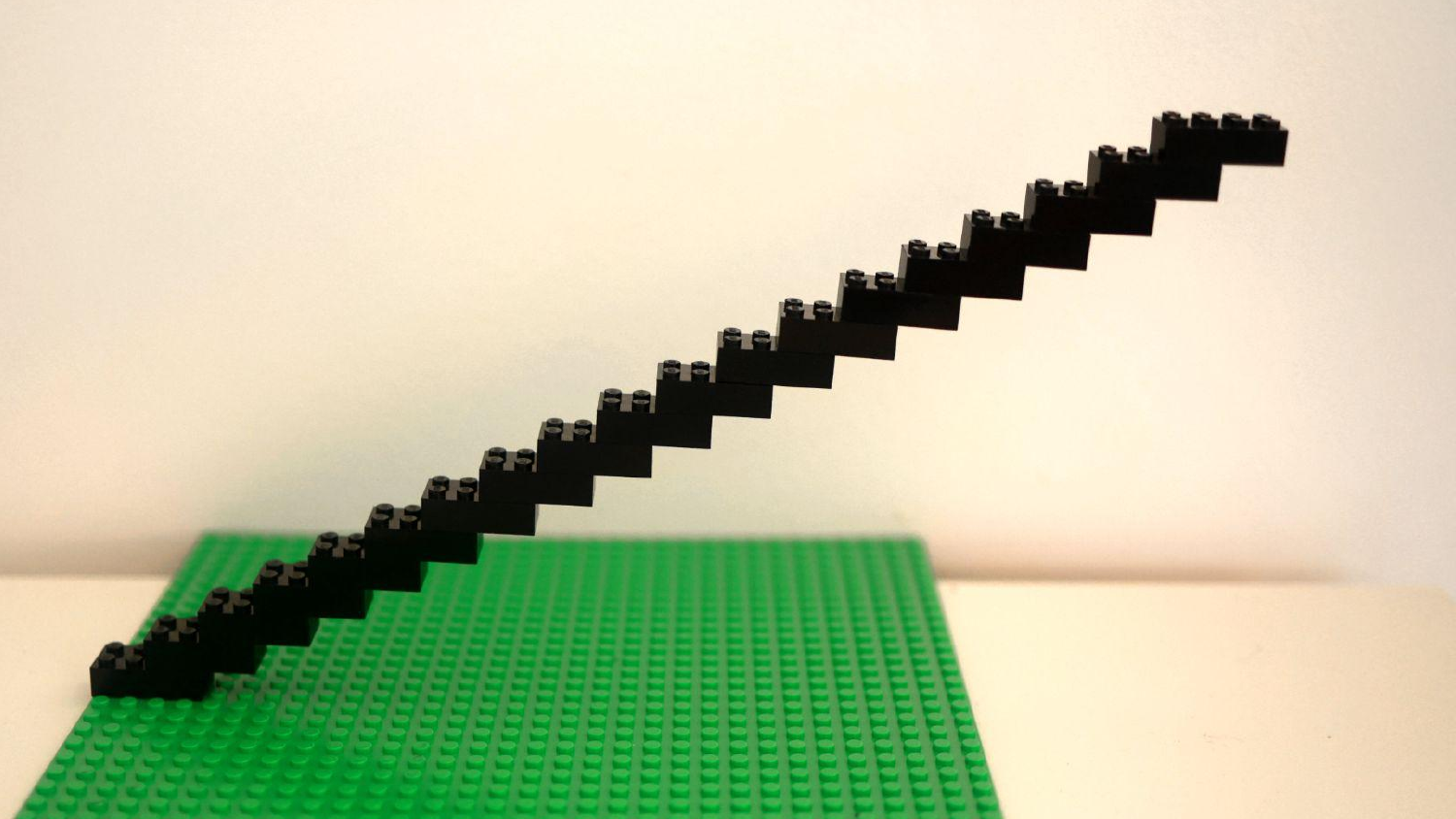}\label{fig:1}}\hfill
\subfigure[Adding one level.]{\includegraphics[width=0.33\linewidth]{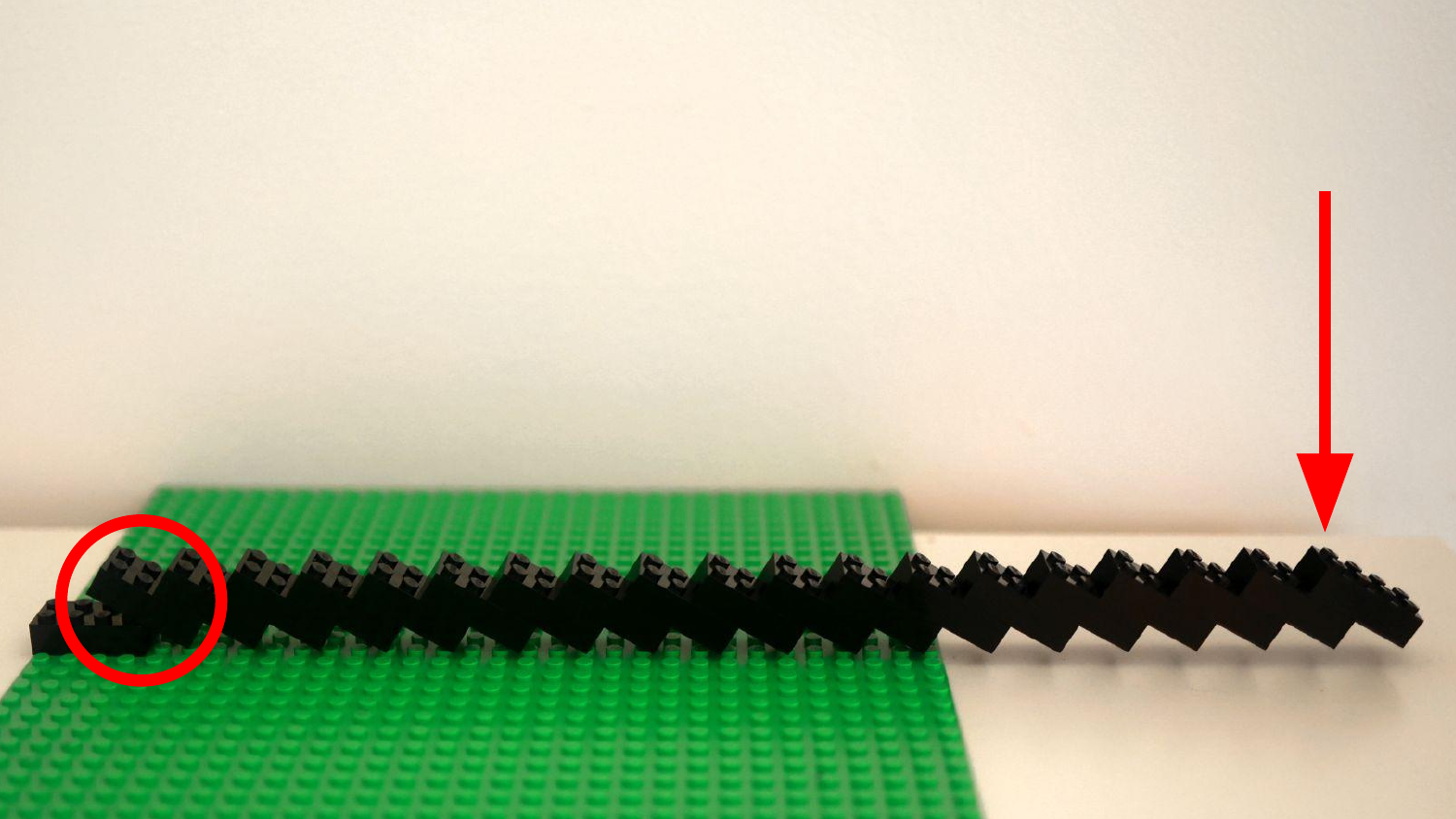}\label{fig:2}}\hfill
\subfigure[Valid 20-level Stairs.]{\includegraphics[width=0.33\linewidth]{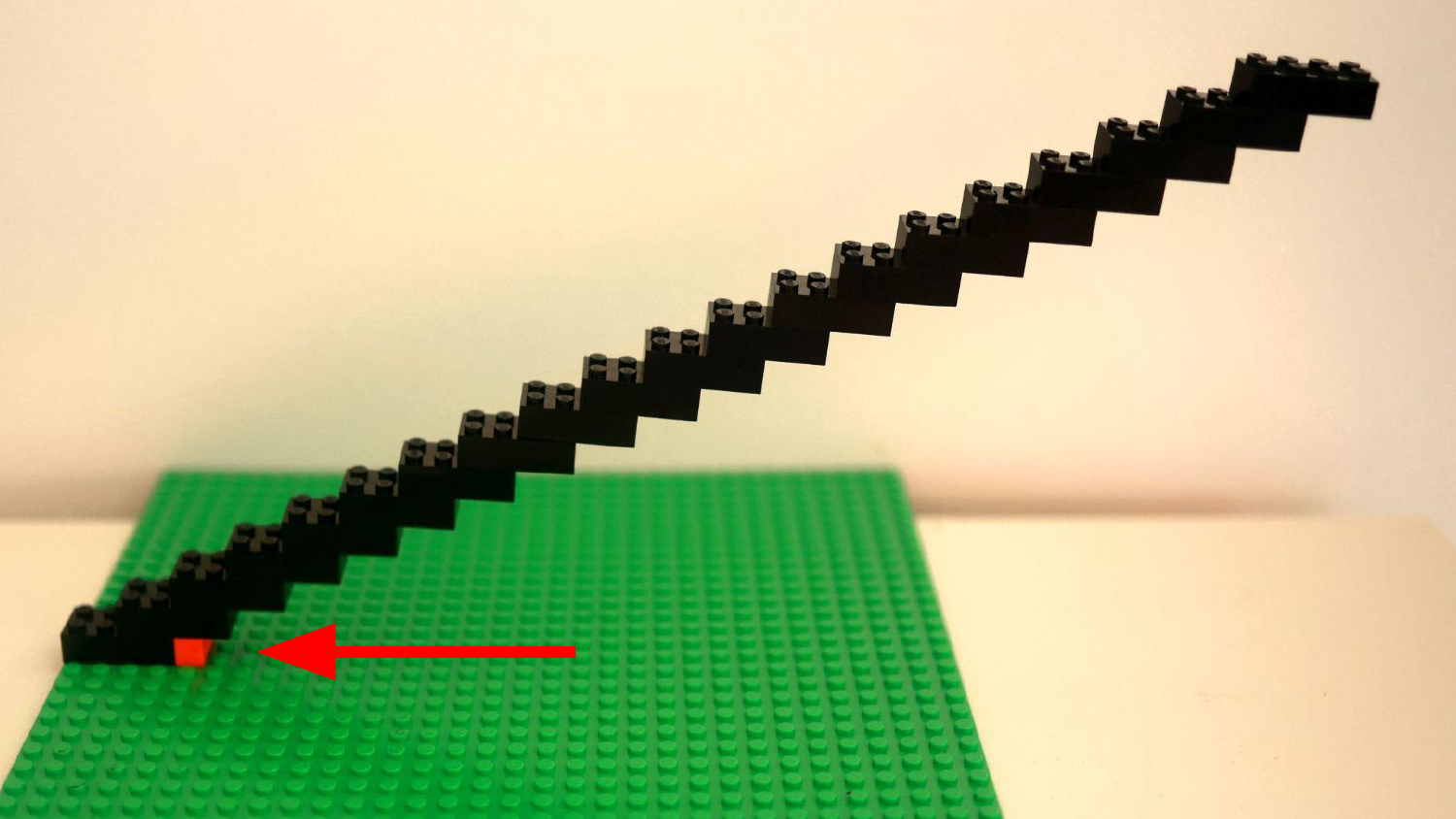}\label{fig:3}}
\\
\vspace{-5pt}
\subfigure[A lever with 2 pink loads.]{\includegraphics[width=0.33\linewidth]{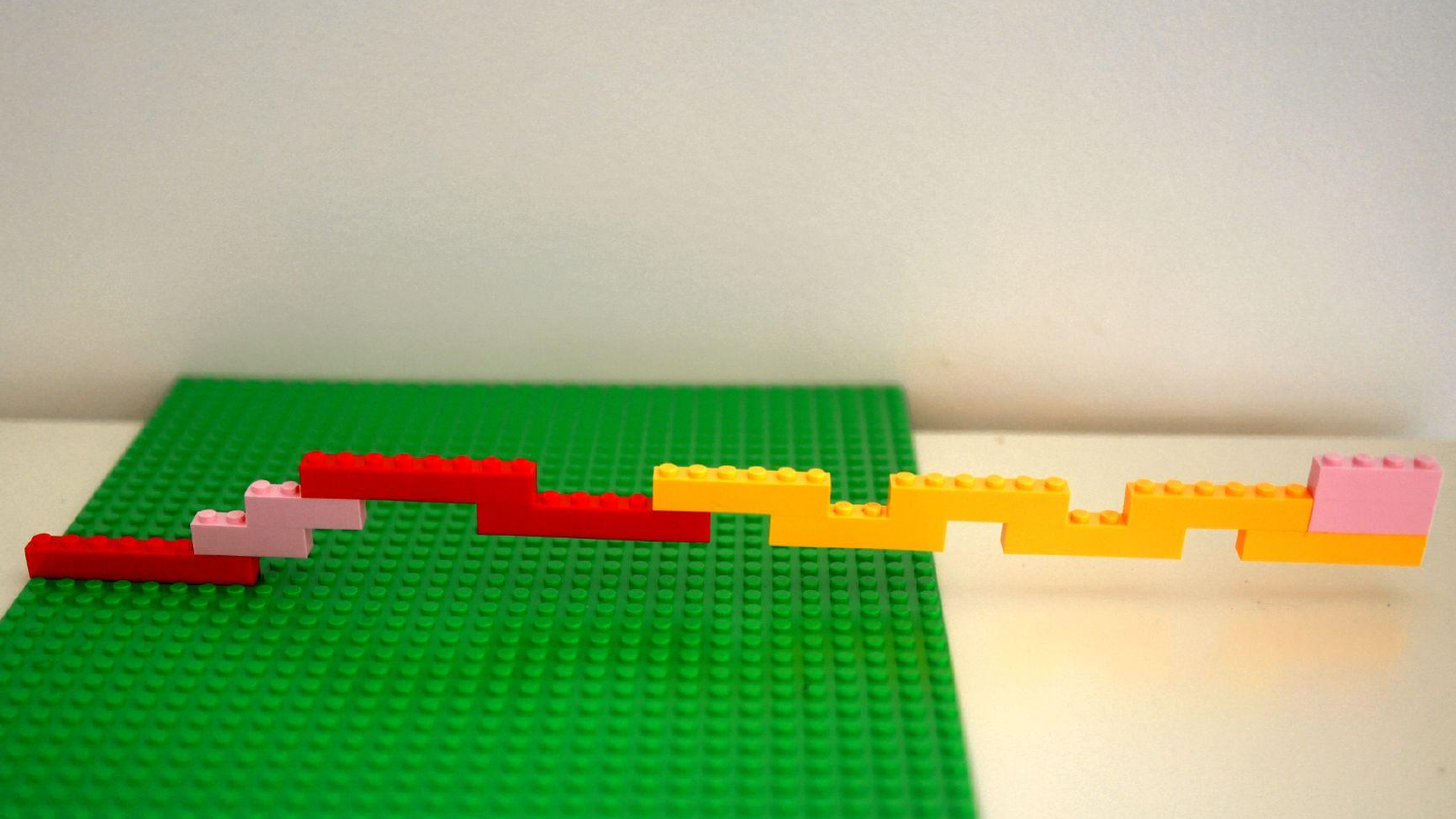}\label{fig:4}}\hfill
\subfigure[Adding one load.]{\includegraphics[width=0.33\linewidth]{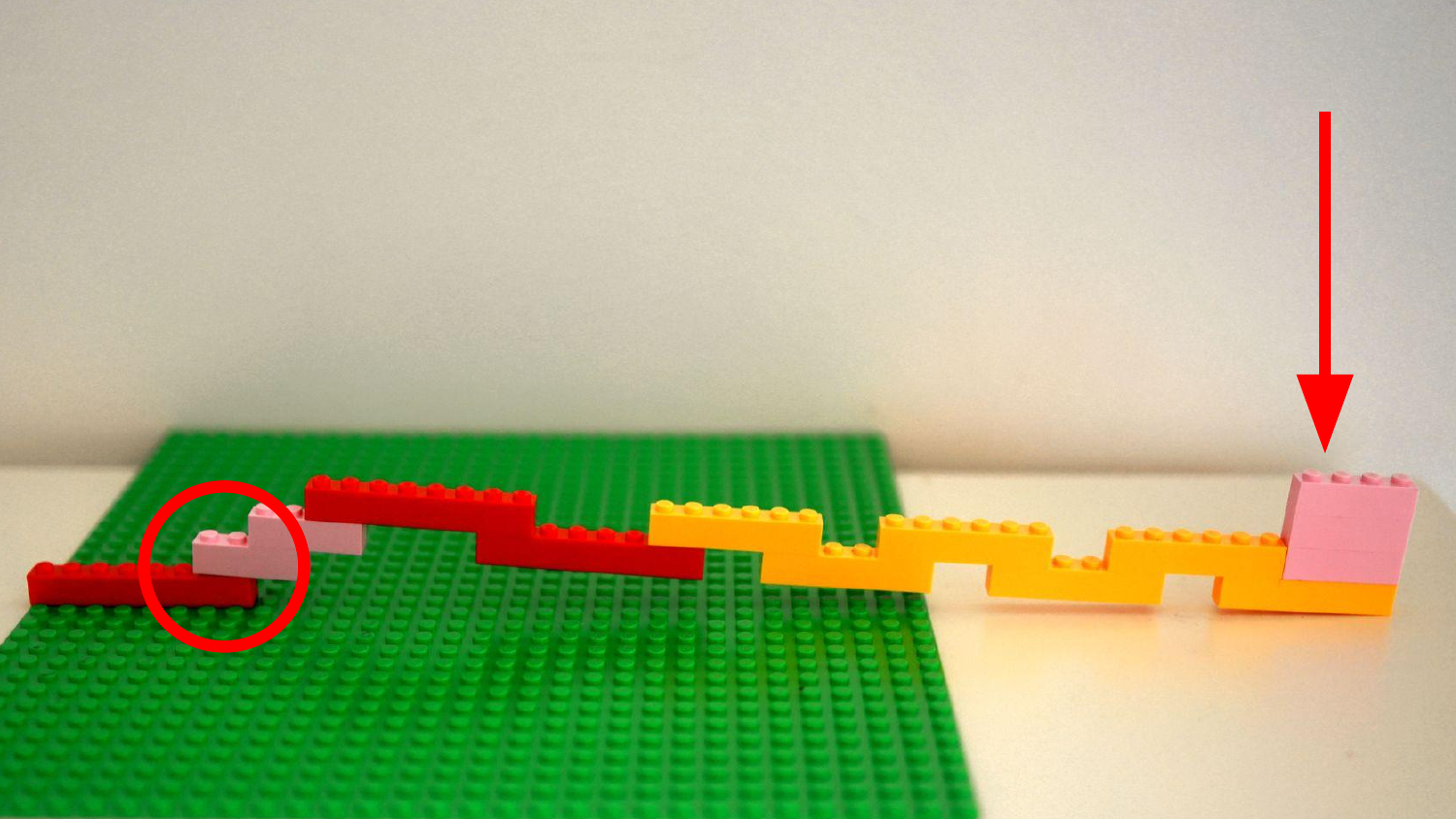}\label{fig:5}}\hfill
\subfigure[A valid lever with 3 pink loads.]{\includegraphics[width=0.33\linewidth]{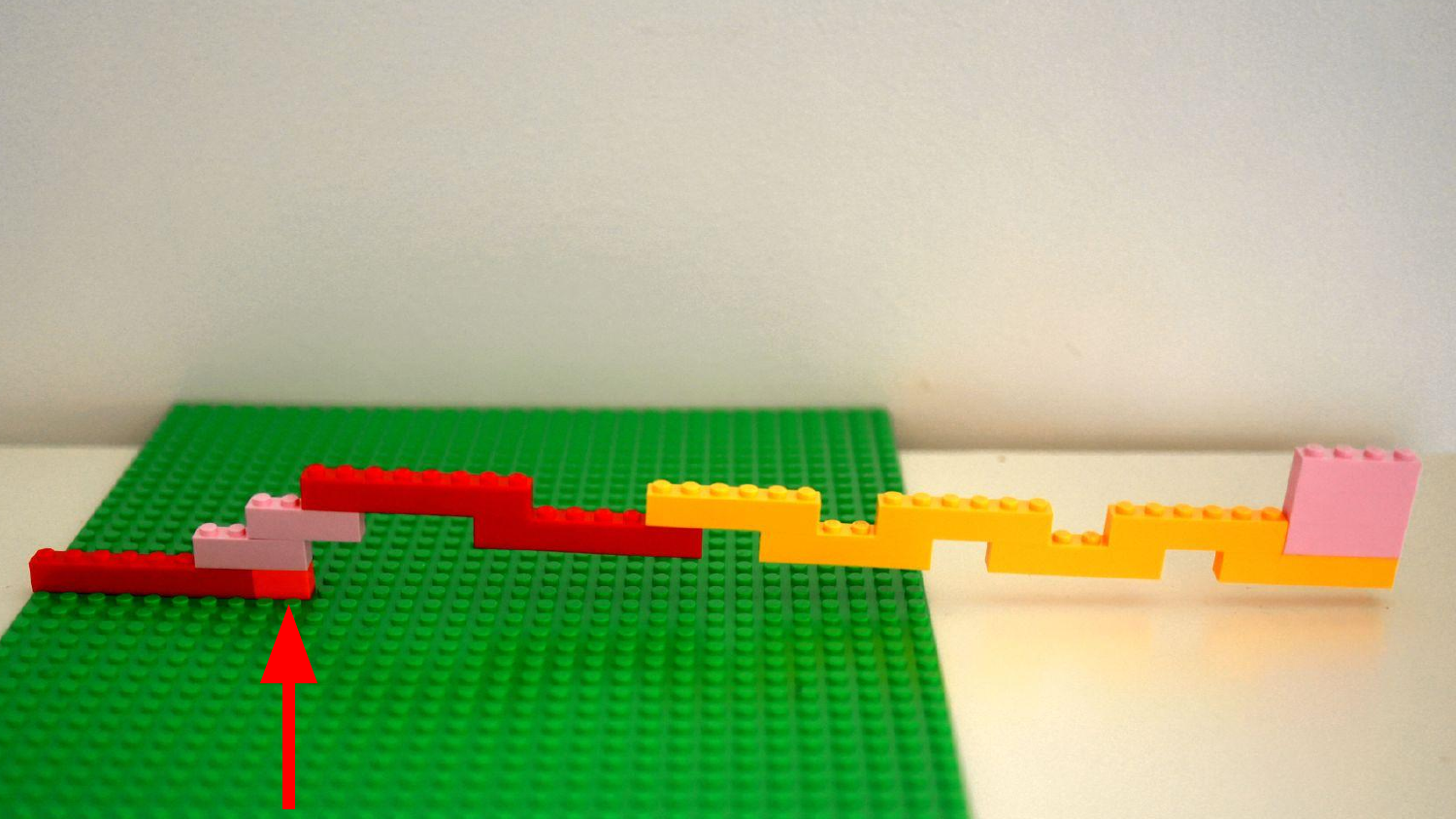}\label{fig:6}}
\vspace{-5pt}
    \caption{\footnotesize Examples of valid and invalid Lego designs. The left and right columns are valid designs and the middle column shows collapsing designs. \label{fig:block_examples}}
    \vspace{-15pt}
    % \vspace{-30pt}
\end{figure}

Recent advancements in robotics enable intelligent robots to perform assembly tasks, such as Lego construction \cite{liu2023lightweight,liu2023simulation, liu2023robotic}, toy insertion \cite{10252579}, electronic assembly \cite{mi14061126}, etc.
A good assembly design (\eg stable) is necessary for successful construction.
% Before the robots perform the actual assembly, an assembly design is necessary.
However, designing assembly requires a non-trivial effort since a slight variation could significantly influence the task.
\Cref{fig:block_examples} showcases examples of both valid and invalid designs. 
Two valid Lego designs are shown in \cref{fig:1,fig:4}. 
However, tiny modifications, \eg adding one brick as depicted in \cref{fig:2,fig:5}, can cause the structures to collapse.
Interestingly, the same small adjustment can stabilize collapsing assemblies, as seen in \cref{fig:3,fig:6}.
Despite the significant impact, these slight variations are barely perceivable to humans.
Conventional approaches leverage rapid prototyping techniques, \eg Computer-aided Design (CAD), to iteratively improve the design \cite{PHAM19981257}.
However, \textit{assembly prototyping} is usually time-consuming and the iterative process could be expensive.

\begin{figure*}
    \centering
    \includegraphics[width=\linewidth]{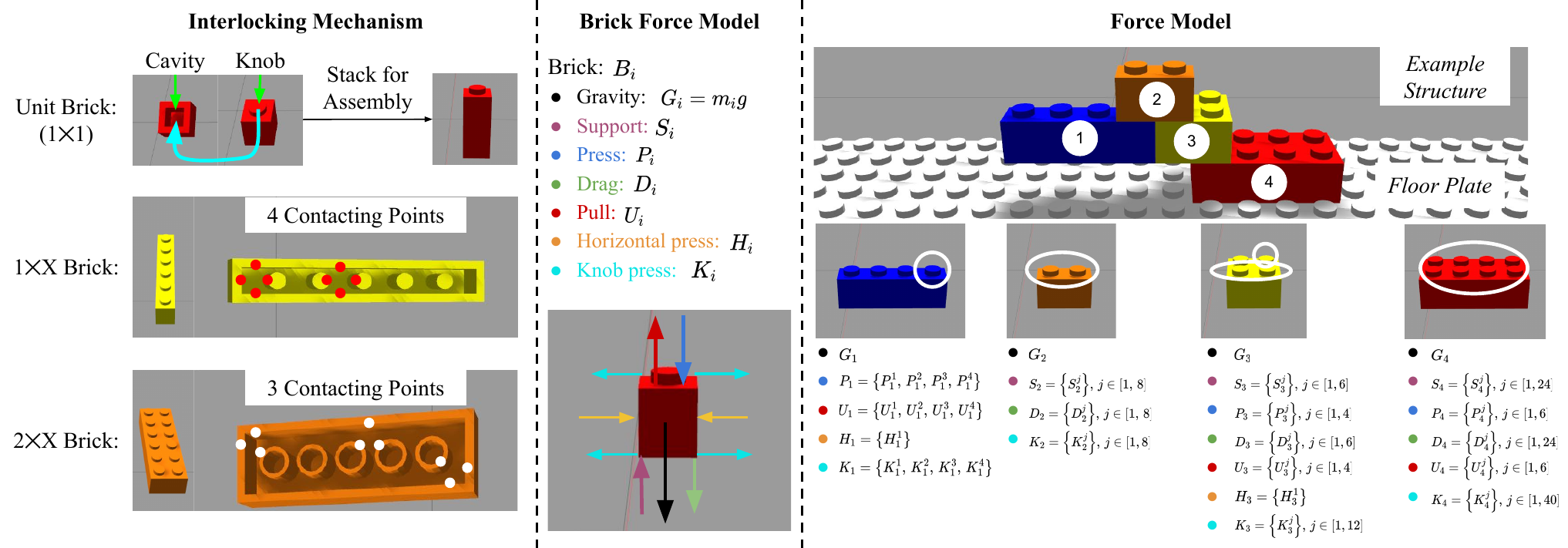}
    \vspace{-5pt}
    \caption{\footnotesize Illustration of the force model of Lego assembly.  \label{fig:force_model}}
    \vspace{-15pt}
\end{figure*}

In particular, this paper considers structural stability, which is a key factor that influences the quality of an assembly design.
It is important to ensure that the assembly design is stable so that an agent can safely perform the construction.
Specifically, this paper focuses on block stacking assembly, where people use different blocks to build 3D structures.
We will use Lego, which is a more complex type of block stacking assembly, to illustrate the concept.
% Lego consists of a wide variety of standardized cubic Lego bricks with different dimensions and colors, which allows users to freely create customized structures.
The top left diagram of \cref{fig:force_model} illustrates the interlocking mechanism of Lego assembly.
A Lego brick is stacked on top of another to form an assembly by inserting the knob into the cavity.
The tight fit of the insertion causes deformation, which generates friction to hold the assembly stable.
Recent works leverage simulations to predict the structural stability of assembly designs \cite{kim2020combinatorial,cannizzaro2023towards}, which is applicable to regular block stacking assembly (\ie blocks with smooth surfaces).
However, to the best of our knowledge, existing simulations are not able to simulate the interlocking mechanism of Lego. 
Therefore, it is challenging to evaluate the stability of a given Lego structure.

To address the challenge, this paper proposes a new optimization formulation to infer the structural stability of block stacking assembly. 
This formulation leverages the rigid block equilibrium (RBE) method and optimizes over force-balancing equations.
The proposed method is tested and verified on hand-crafted Lego examples.
The experiment results demonstrate that the proposed stability analysis can correctly predict whether the structure is stable.
In addition, it outperforms the existing methods since it can locate the weakest parts in the design and, more importantly, solve \textit{any} given assembly structure.
To further validate our method, we provide \textit{StableLego}: a comprehensive Lego assembly dataset, which includes a wide variety of Lego assembly designs for real-world objects.
StableLego is a novel benchmark that could facilitate research in related areas.
The dataset includes more than 50k Lego structures built using standardized Lego bricks with different dimensions. 
We apply the proposed stability analysis to the dataset and include the stability inferences in the dataset.
To the best of our knowledge, StableLego is the \textit{first} Lego assembly dataset with stability inferences.
Our stability analysis implementation and the StableLego dataset are available at \url{https://github.com/intelligent-control-lab/StableLego}.

\section{Related Works}

In this paper, we mainly focus on the stability of block stacking structures \cite{https://doi.org/10.1111/cgf.142660}. 
The finite element method (FEM) is widely used in analyzing complex assembly structures \cite{ZHOU2020103282}.
However, it is usually time-consuming if accuracy is required \cite{pletzbrickfem}.
Due to its customizability, Lego assembly has been recently widely studied \cite{ChungH2021neurips,wang2022translating,10.1007/978-3-031-19815-1_6}.
Prior works design rules to intuitively evaluate the structural stability and improve the assembly design \cite{gower1998lego,smal2008automated, LEGOBuilder,testuz2013automatic,8419684,petrovic2001solving,10.1111:cgf.13603}.
Such rules can be, for instance, maximizing the number of knob-to-cavity connections; minimizing the number of bricks; and maximizing the number of brick orientation alternations.
Although these rules provide insights into Lego structural stability, they are difficult to apply to other block assembly tasks.
Moreover, these pre-defined rules only provide intuitive understanding instead of quantitative measurements with physical implications.
Recent works \cite{kim2020combinatorial,cannizzaro2023towards} leverage simulators with a physics engine to simulate the behavior of assembly structures. 
However, it is difficult to simulate the interlocking mechanism between Lego bricks with existing simulators. 
Therefore, only block stacking with smooth surfaces can be addressed.
% Furthermore, Kim et al. \cite{kim2020combinatorial} only assess whether the structure would fall as a whole without considering potential collapse triggered by internal stresses.
Other recent works \cite{groth2018shapestacks,wu2024efficient} directly train a neural network to predict stability. 
However, such learning-based approaches require a significant amount of data, which is non-trivial to generate.
% In contrast, our method does not require training and provides a quantitative stability assessment that can be applied to generic block stacking structures.

\begin{figure}
\centering
\subfigure[]{\includegraphics[width=0.25\linewidth]{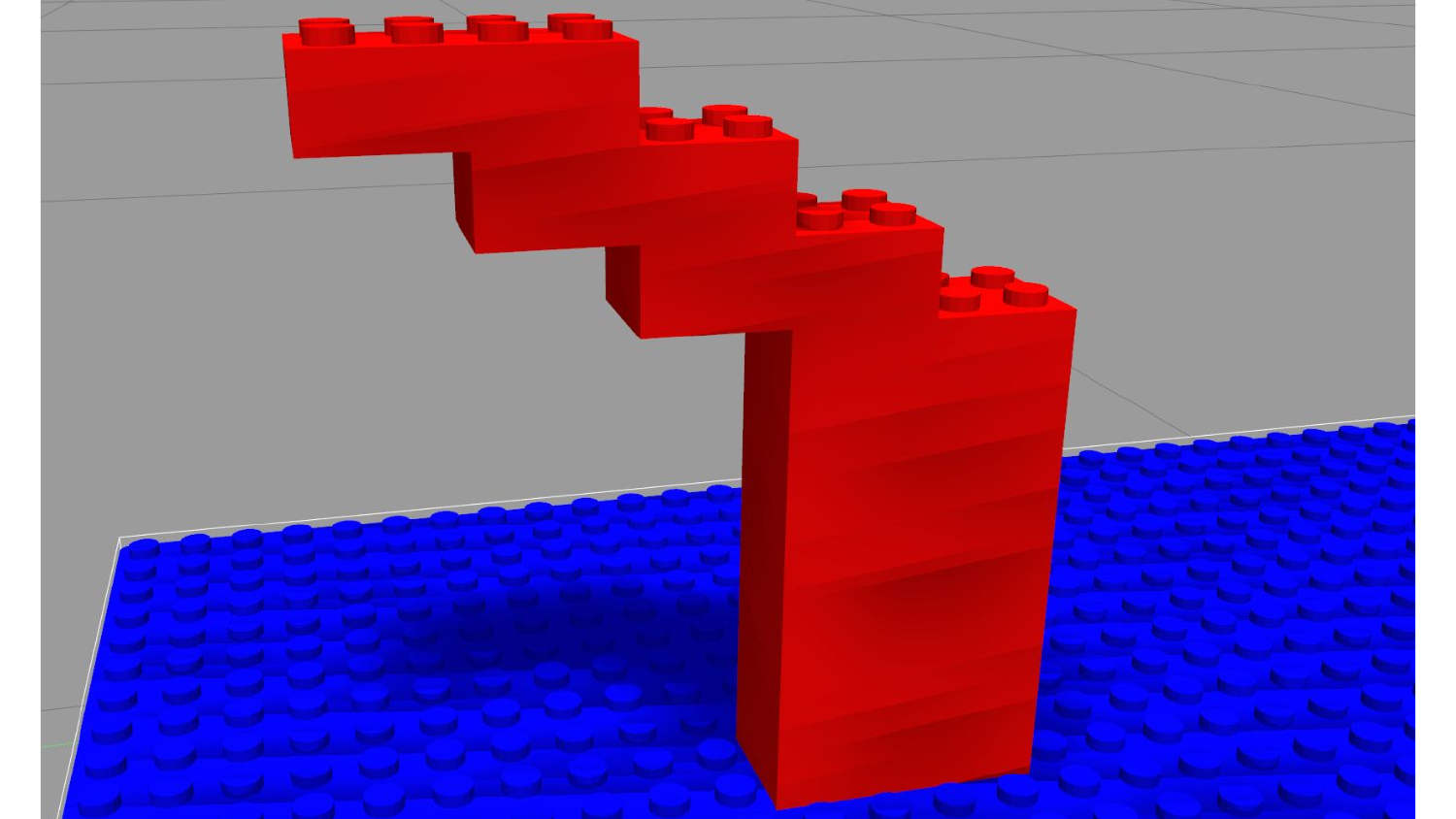}\label{fig:single_connected}}\hfill
\subfigure[]{\includegraphics[width=0.25\linewidth]{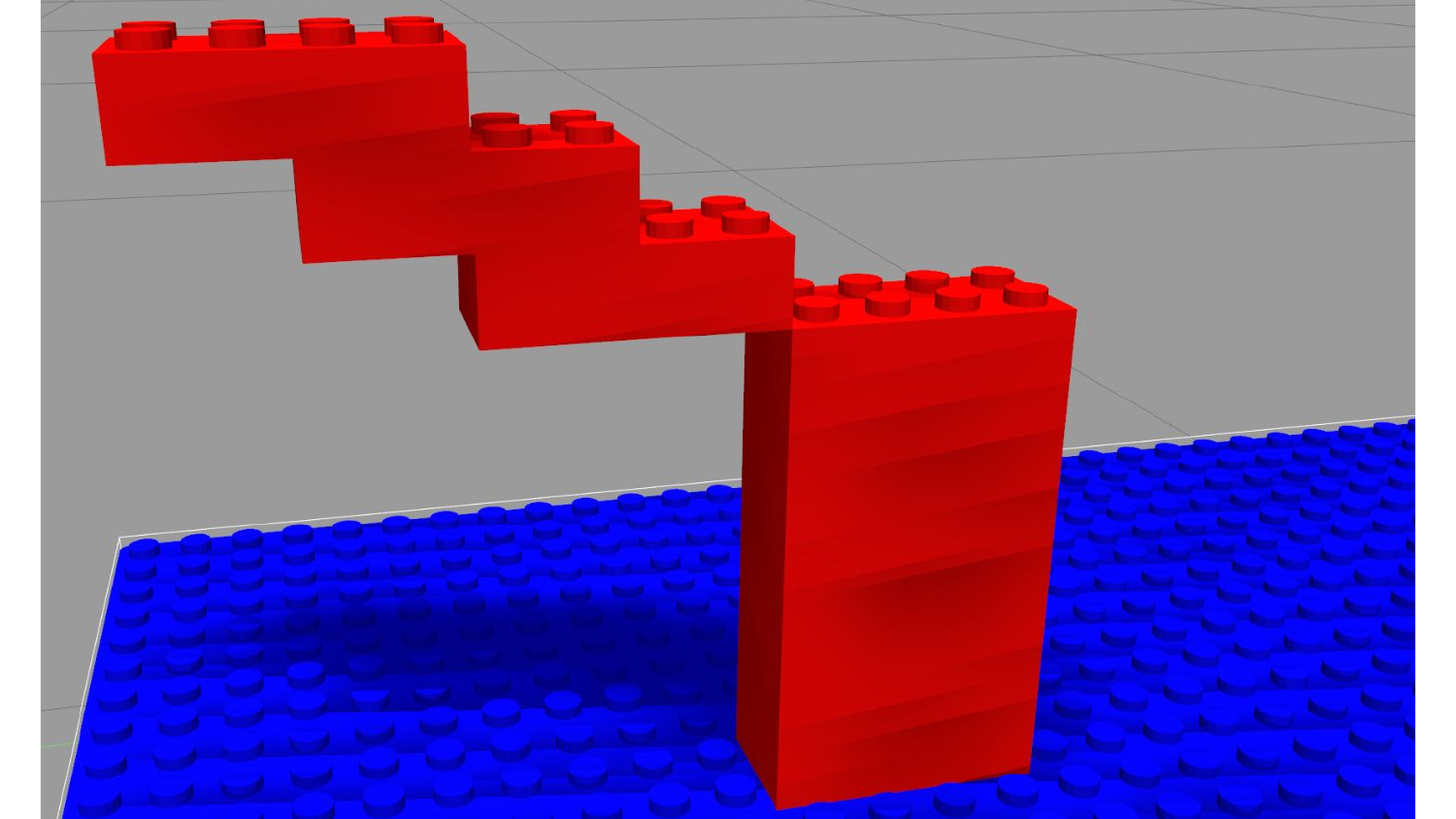}\label{fig:not_single_connected}}\hfill
\subfigure[]{\includegraphics[width=0.25\linewidth]{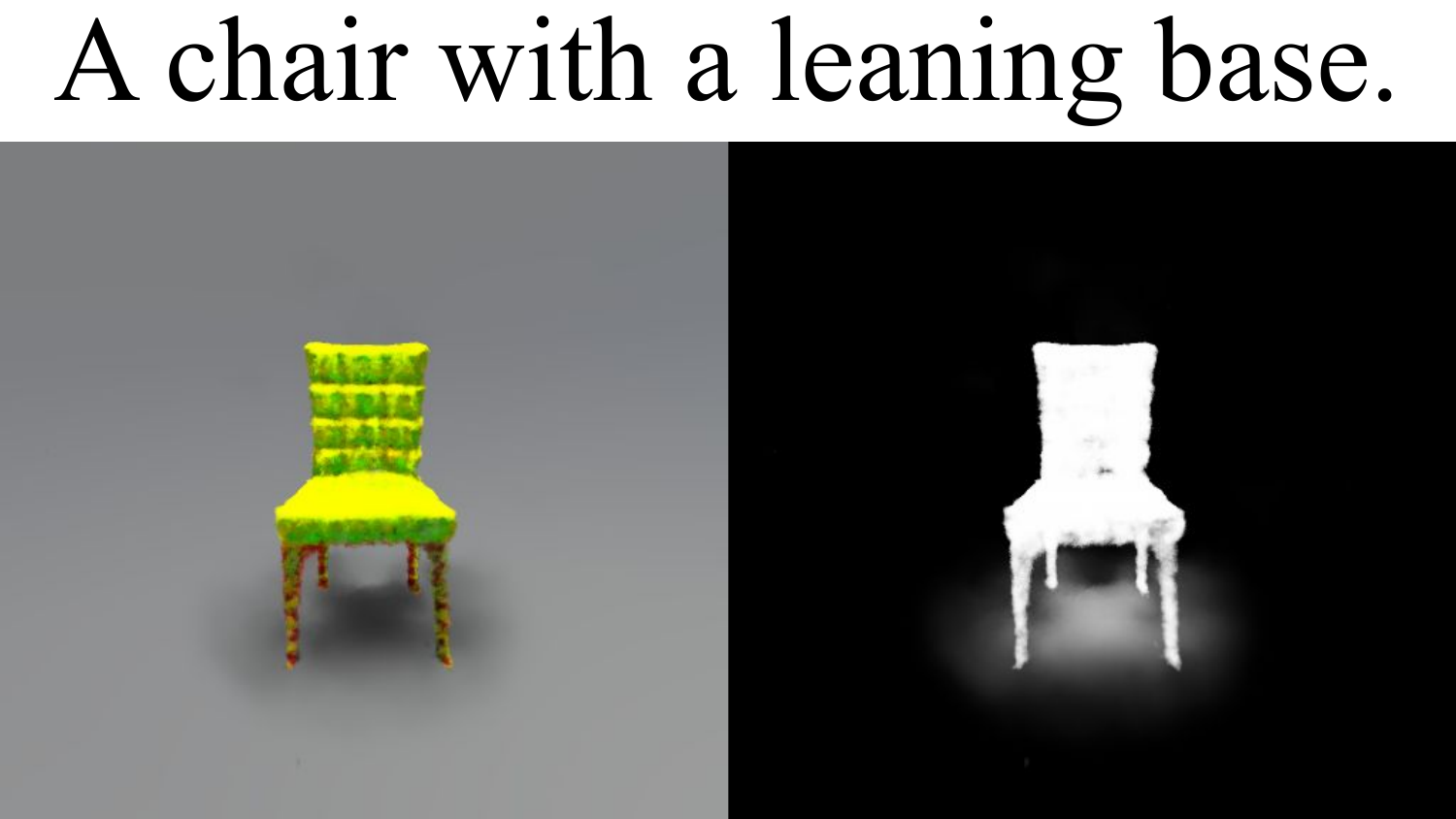}\label{fig:generative_ai}}\hfill
\subfigure[]{\includegraphics[width=0.25\linewidth]{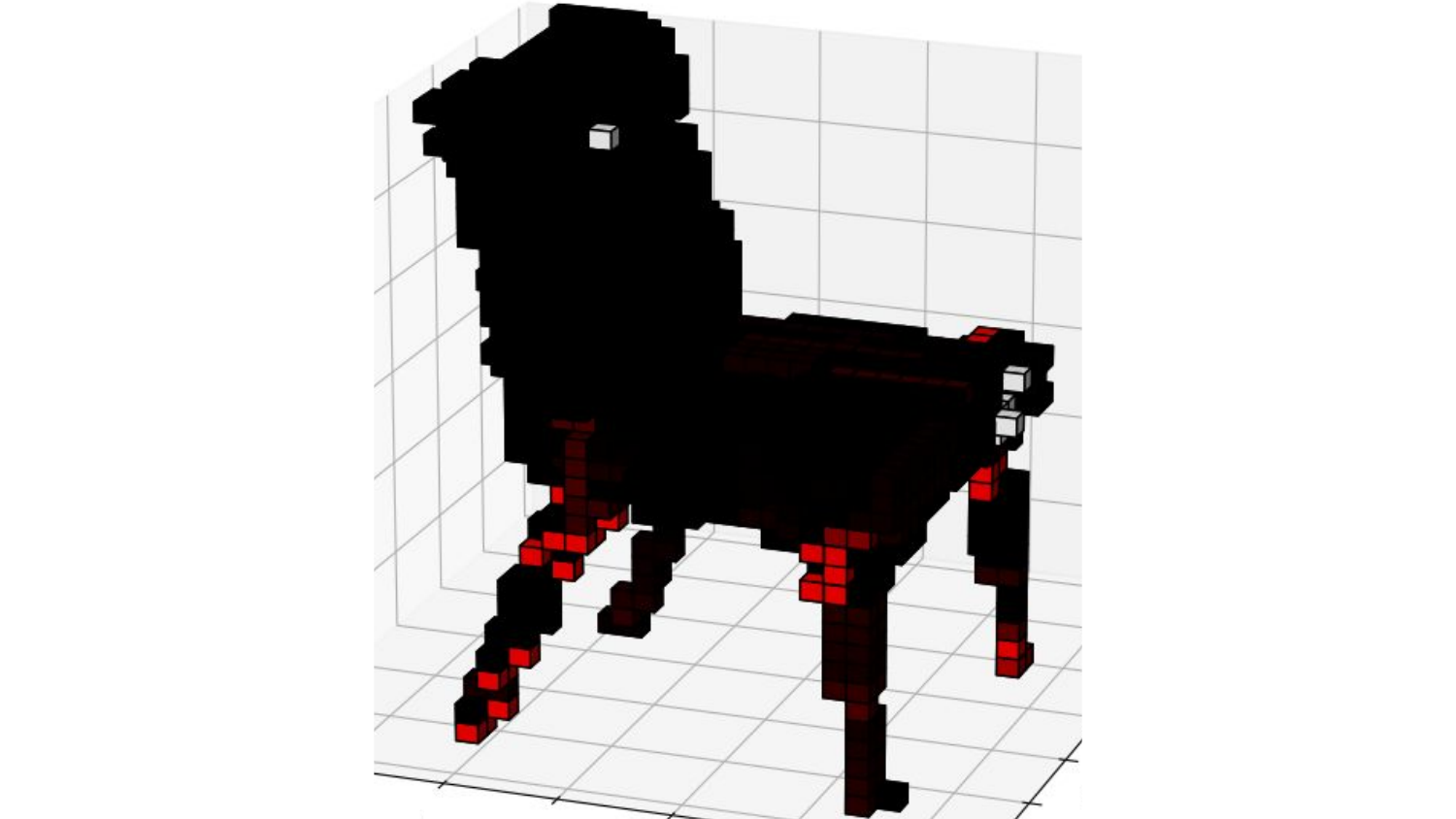}\label{fig:generative_ai_analysis}}
\vspace{-5pt}
    \caption{\footnotesize Illustrations of the single-connected assumption. (1) A single-connected design. (2) An assembly design that is not single-connected. (3) A 3D model generated by generative AI. (4) The Lego design of the generated structure. White: floating bricks.}
    \vspace{-15pt}
\end{figure}

On the other hand, the rigid block equilibrium (RBE) method \cite{10.1145/1618452.1618458} formulates the stability analysis as an optimization problem and solves a force distribution that satisfies the static equilibrium constraints.
It is widely used in evaluating the structural stability \cite{10.1145/1618452.1618458, KAO2022103216}.
Recent works \cite{10.1145/2816795.2818091, KOLLSKER2021270,kollsker2021optimisation} have utilized RBE-based techniques to evaluate and optimize Lego layouts.
However, these existing methods assume that the block assembly design is single-connected.
\Cref{fig:single_connected} illustrates a single-connected Lego design, whereas \cref{fig:not_single_connected} depicts a design that is not single-connected since the top three bricks do not have a connected path to the ground.
Existing methods would fail if the assumption is violated.
Although high-quality assembly designs usually assert single-connectivity, preliminary raw designs, \eg a design from generative AI, may violate this assumption. 
\Cref{fig:generative_ai} illustrates an example 3D structure from generative AI \cite{poole2022dreamfusion} with its corresponding prompt.
Despite the promising overall 3D shape, the corresponding Lego design could be imperfect as shown in \cref{fig:generative_ai_analysis} since it contains floating bricks (\ie the white bricks).
Such a design violates the single-connected assumption and is not solvable by existing methods. 
% Contrarily, our method can effectively address these cases.

\section{Stability Analysis \\ of Block Stacking Structures}

Following the idea of RBE, this paper formulates the stability analysis as an optimization problem and solves a force distribution by optimizing over force balancing equations.
Unlike the prior works, the key difference is that our formulation encodes the static equilibrium conditions in the objective function while imposing additional physical constraints on the optimization.
% We will introduce our analysis formulation using Lego since it presents additional complexities due to the interlocking mechanism.  
Similar to existing RBE methods, we assume all assembly components (\ie Lego bricks) are rigid bodies, and factors (\eg material, temperature, etc) will only influence the friction capacity $T$ in \cref{eq:capacity}.
In addition, we assume all connections between bricks are well-established, \ie all cavities and knobs are snapped together for all connections. 
The proposed formulation can be easily reduced to regular block stacking assembly.

\subsection{Force Model}
\Cref{fig:force_model} illustrates the force model in our stability analysis, which is adopted from Luo et al. \cite{10.1145/2816795.2818091}.
The middle diagram of \cref{fig:force_model} depicts the potential forces exerted on a single Lego brick in an assembly.
Given an assembly consisting of $N$ bricks, we denote a brick as $B_i$, where $i\in[1, N]$.
For any $B_i$, it has the gravity $\Vec{G}_i=m_i\Vec{g}$ applying on it, where $m_i$ is the brick mass and $\Vec{g}\approx9.8\ \text{N/kg}$.
If there is a connection to the top knob, $B_i$ will experience pressing force  $\Vec{P}_i$ (\ie the blue arrow) pointing downward due to the weight of the structures above it, as well as pulling force  $\Vec{U}_i$ (\ie the red arrow) pointing upward due to the tight connection of the knob.
Similarly, if there is a connection to the bottom cavity, $B_i$ will experience supporting force  $\Vec{S}_i$ (\ie the purple arrow) pointing upward due to the rigid structure below it, as well as dragging force $\Vec{D}_i$ (\ie the green arrow) pointing downward due to the friction from the connection.
If there are bricks right next to $B_i$, there will also be horizontal press $\Vec{H}_i$ (\ie the yellow arrows) pointing toward $B_i$.
If a knob or a cavity of $B_i$ is connected, there will be horizontal press $\Vec{K}_i$ within the knob (\ie the cyan arrows) pointing in horizontal directions that prevent the brick from sliding.
Note that each connection will generate 4 horizontal press force components in the 4 horizontal directions, \ie $\pm X$ and $\pm Y$, pointing inward to $B_i$.
Also, note that only $\Vec{G}_i$ is a force constantly exerting on $B_i$ independent of the assembly structure.
$\Vec{S}_i, \Vec{P}_i, \Vec{D}_i, \Vec{U}_i, \Vec{H}_i, \Vec{K}_i$ are forces that may or may not exist depending on the structure.
In the following discussion, we refer to these forces as candidate forces.

The figures on the left of \cref{fig:force_model} illustrate different connections of bricks.
Depending on the different dimensions of the \textbf{top} bricks, there are different numbers of contacting points that generate friction to hold the knobs of the bottom bricks.
If the top brick is $1\times X$, where $X\in \mathbb{N}, X\geq 1$, each connected knob has 4 contact points.
If the top brick is $2\times X$, $X\geq 2$, each connected knob has 3 contact points. 
If the top brick is $Q\times X$, $Q\geq 3, X\geq Q$, the connections on the edge have 3 contact points while others have 4 contact points.
In our formulation, instead of summing up the candidate forces and assuming only one vertical candidate for each of the $\Vec{S}_i, \Vec{P}_i, \Vec{D}_i, \Vec{U}_i$ within each knob-to-cavity connection, we assume the vertical candidate forces exist at each of the contact points.

The right figure in \cref{fig:force_model} illustrates the force models for each brick in an example Lego structure.
The white contours indicate the connected knobs for each brick.
If there is no connection, either on top or below a knob, there are no candidate forces exist.
The bottom of the diagram lists all the potential forces that are exerted on the brick.
All bricks have gravity applied to them.
For $B_1$, since only the right-most knob has a $1\times 2$ brick connected on top of it, it has 4 pressing candidates $\Vec{P}_1=\{\Vec{P}_1^1, \Vec{P}_1^2, \Vec{P}_1^3, \Vec{P}_1^4\}$ and 4 pulling candidates $\Vec{U}_1=\{\Vec{U}_1^1, \Vec{U}_1^2, \Vec{U}_1^3, \Vec{U}_1^4\}$ since the connection has 4 contact points.
And there exist 4 knob pressing candidates $\Vec{K}_1=\{\Vec{K}_1^1, \Vec{K}_1^2, \Vec{K}_1^3, \Vec{K}_1^4\}$ in 4 horizontal directions.
Since there exists a brick (\ie $B_3$) right next to it, it has a horizontal press candidate $\Vec{H}_1=\{\Vec{H}_1^1\}$.
Similarly for $B_2$, since there are only connections below it, there is no $\Vec{U}_2$ or $\Vec{P}_2$.
Due to the cavity connections, there are 8 supporting candidates $\Vec{S}_2=\{\Vec{S}_2^j\ |\ j\in[1,8]\}$ since each cavity has 4 contact points. 
Similarly there are 8 dragging candidates $\Vec{D}_2=\{\Vec{D}_2^j\ |\ j\in[1,8]\}$ and 8 knob pressing candidates $\Vec{K}_2=\{\Vec{K}_2^j\ |\ j\in[1,8]\}$.
Since there is no brick right next to $B_2$, $\Vec{H}_2$ does not exist.
We can derive the force models for $B_3$ and $B_4$ following the similar rules as listed in \cref{fig:force_model}.

\subsection{Static Equilibrium}
An object reaching static equilibrium indicates that it will not fall or collapse.
To ensure a stable Lego structure, we need to ensure that each brick $B_i$ can reach static equilibrium so that the structure will not collapse.
For a given Lego structure with $N$ bricks and each candidate force $F_i$ has $M_{F_i}$ candidates, the static equilibrium enforces that $\forall B_i, i\in [1, N]$, we need to satisfy
\begin{align}\label{eq:force_eq}
    C_i^f\dot =~&\Vec{G}_i + \sum_{j=1}^{M_{F_i}}\Vec{F}_i^j = \Vec{0},
\\
\label{eq:torque_eq}
        C_i^\tau\dot =~&\Vec{L}_i^{\Vec{G}_i} \times \Vec{G}_i + \sum_{j=1}^{M_{F_i}} (\Vec{L}_i^{\Vec{F}_i^j}\times \Vec{F}_i^j) = \Vec{0},
\\
\nonumber
        \Vec{F}_i^j\in \mathbf{F}_i  =\{& \Vec{S}_i^{j_S}, \Vec{P}_i^{j_P}, \Vec{D}_i^{j_D}, \Vec{U}_i^{j_U}, \Vec{H}_i^{j_H}, \Vec{K}_i^{j_K} \mid \\ 
        \nonumber
        & j_S\in [1, M_{S_i}], j_P\in [1, M_{P_i}], j_D\in [1, M_{D_i}]\\
        \nonumber
        & j_U\in [1, M_{U_i}], j_H\in [1, M_{H_i}], j_K\in [1, M_{K_i}]\},
\end{align}
where $\times$ denotes the vector cross-product operation.
$\Vec{L}_i^{\Vec{F}}$ is the force lever of the force vector $\Vec{F}$ on brick $B_i$.
\Cref{eq:force_eq} enforces that $B_i$ reaches force equilibrium so that the brick would not have translational motion.
\Cref{eq:torque_eq} enforces that $B_i$ reaches torque equilibrium (also referred as moment equilibrium). This indicates that the brick would not have rotational motion.
Satisfying both \cref{eq:force_eq,eq:torque_eq} indicates that the bricks are static and the structure is stable.

% \subsection{Determining Brick Rotation Center}
% \ruixuan{TODO}

\subsection{Constraints}
\paragraph{Non-negativity}
We assume all components are rigid bodies.
Therefore, the value of each force should be non-negative.
Let the value of $\Vec{F}_i^j\in\mathbf{F}_i$ be $F_i^j$, we have
\begin{equation}\label{eq:non-neg}
    C_i^{+}:F_i^j \geq 0.
\end{equation}

\paragraph{Non-coexistence}
At any given contact point, the pulling force $\Vec{U}_i^j$ and the pressing force $\Vec{P}_i^j$ cannot coexist. 
If $U_i^j>0$, the top brick is pulling the bottom brick upward. 
Then there is no weight loaded on the bottom brick, and thus, $P_i^j=0$.
If $P_i^j>0$, then there is weight loaded on the bottom brick.
Therefore, the top brick cannot be pulling the bottom brick upward.
Similarly, the dragging force $\Vec{D}_i^j$ and the supporting force $\Vec{S}_i^j$ cannot coexist.
The non-coexistence property gives the constraint as
\begin{equation}\label{eq:non-coexist}
    C_i^{||}: \begin{cases}
			P_i^j\cdot U_i^j=0\\
            D_i^j\cdot S_i^j=0
		 \end{cases}.
\end{equation}

\paragraph{Equality}
Newton's third law states that for every action, there is an equal and opposite reaction.
At a given contact point $q$, let the bottom brick be $B_i$ and the upper brick be $B_j$.
The supporting force $\Vec{S}_j^q$ and the pressing force $\Vec{P}_i^q$ are such an action-reaction pair. 
Similarly, the pulling force $\Vec{U}_i^q$ and the dragging force $\Vec{D}_j^q$ are also an action-reaction pair.
Also, the knob pressing candidates $\Vec{K}_i$ and $\Vec{K}_j$ are also action-reaction pairs.
Let $B_k$ be a brick adjacent to $B_i$, then the horizontal press $\Vec{H}_i$ and $\Vec{H}_k$ are also an action-reaction pairs. 
Therefore, we have the equality constraints as
\begin{equation}\label{eq:equality}
    C^=: \begin{cases}
        S_j^q=P_i^q\\
        U_i^q=D_j^q\\
        H_i=H_k\\
        K_i=K_j.
    \end{cases}
\end{equation}

\begin{figure*}
\centering
\subfigure[Ours \cref{fig:1}]{\includegraphics[width=0.16\linewidth]{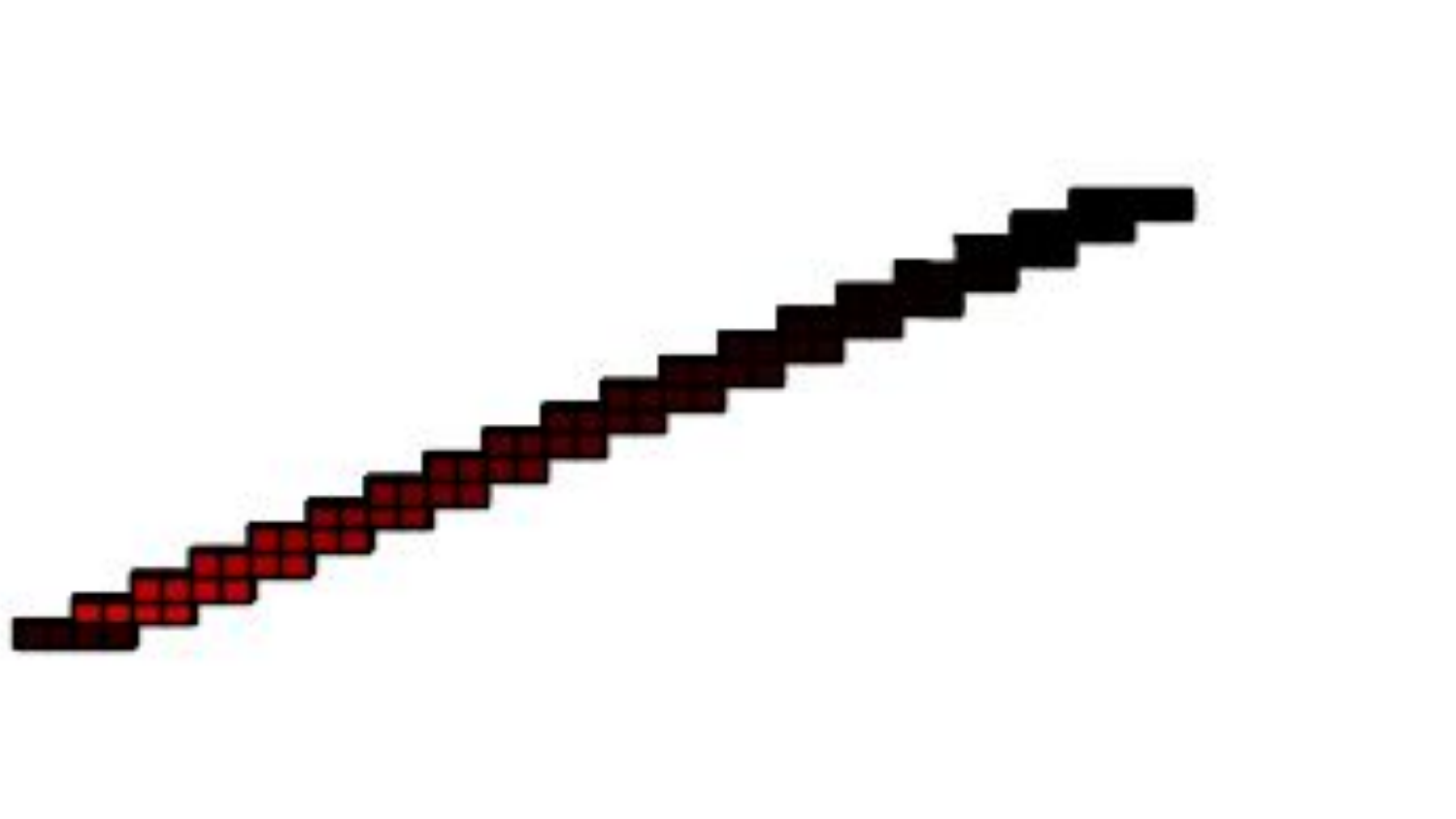}\label{fig:1_analysis}}\hfill
\subfigure[Ours \cref{fig:2}]{\includegraphics[width=0.16\linewidth]{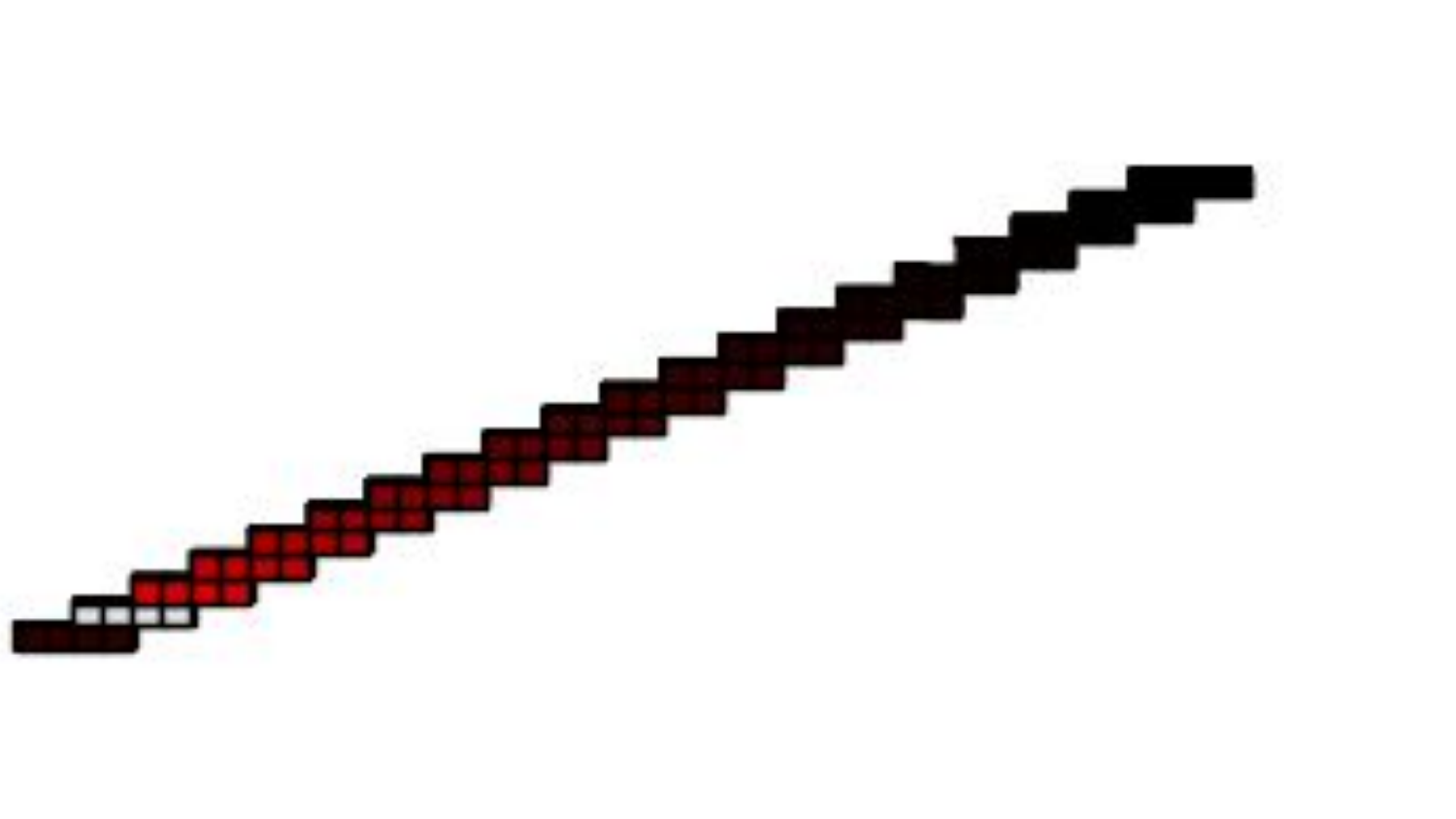}\label{fig:2_analysis}}\hfill
\subfigure[Ours \cref{fig:3}]{\includegraphics[width=0.16\linewidth]{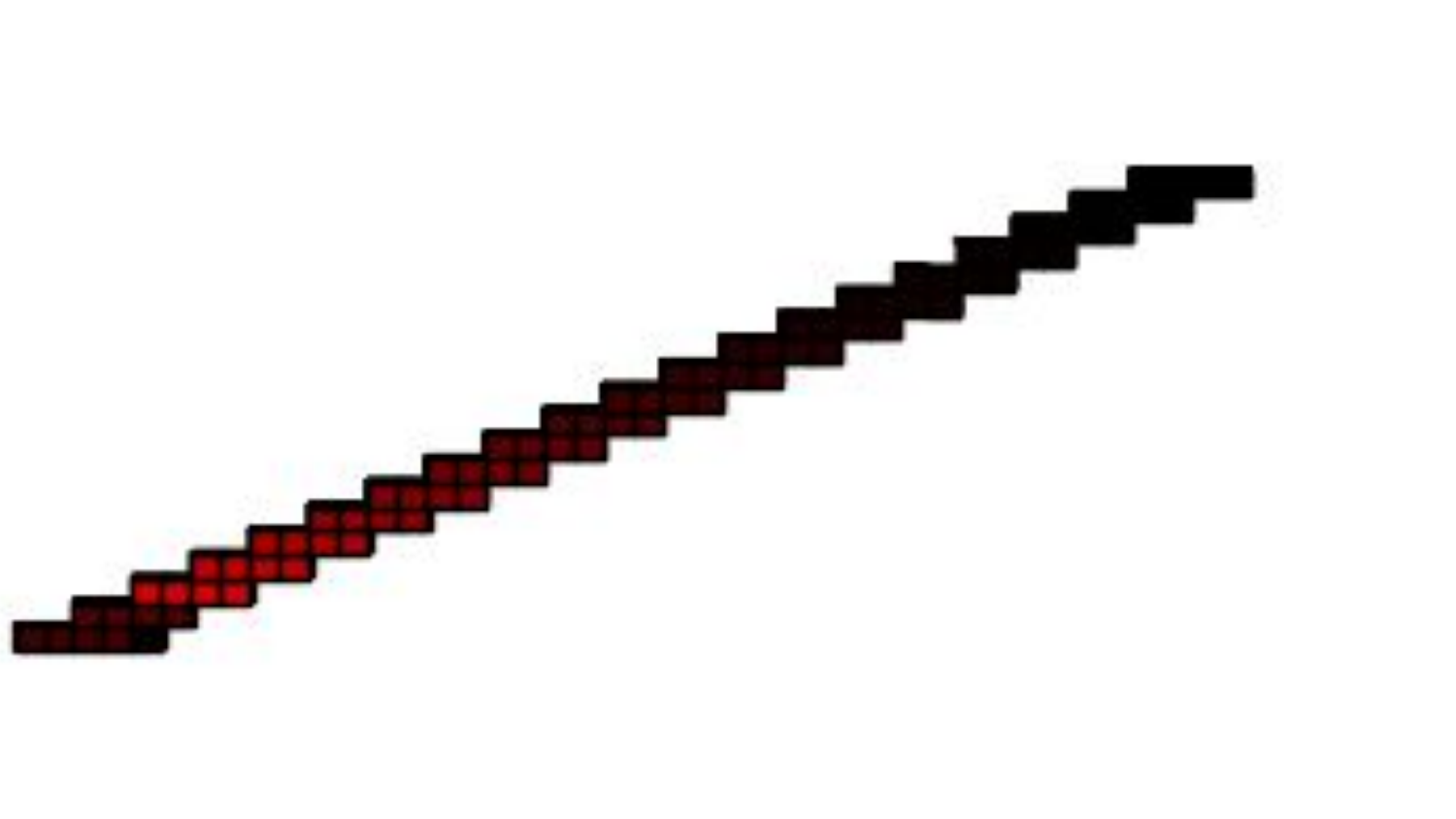}\label{fig:3_analysis}}\hfill
\subfigure[Ours \cref{fig:4}]{\includegraphics[width=0.16\linewidth]{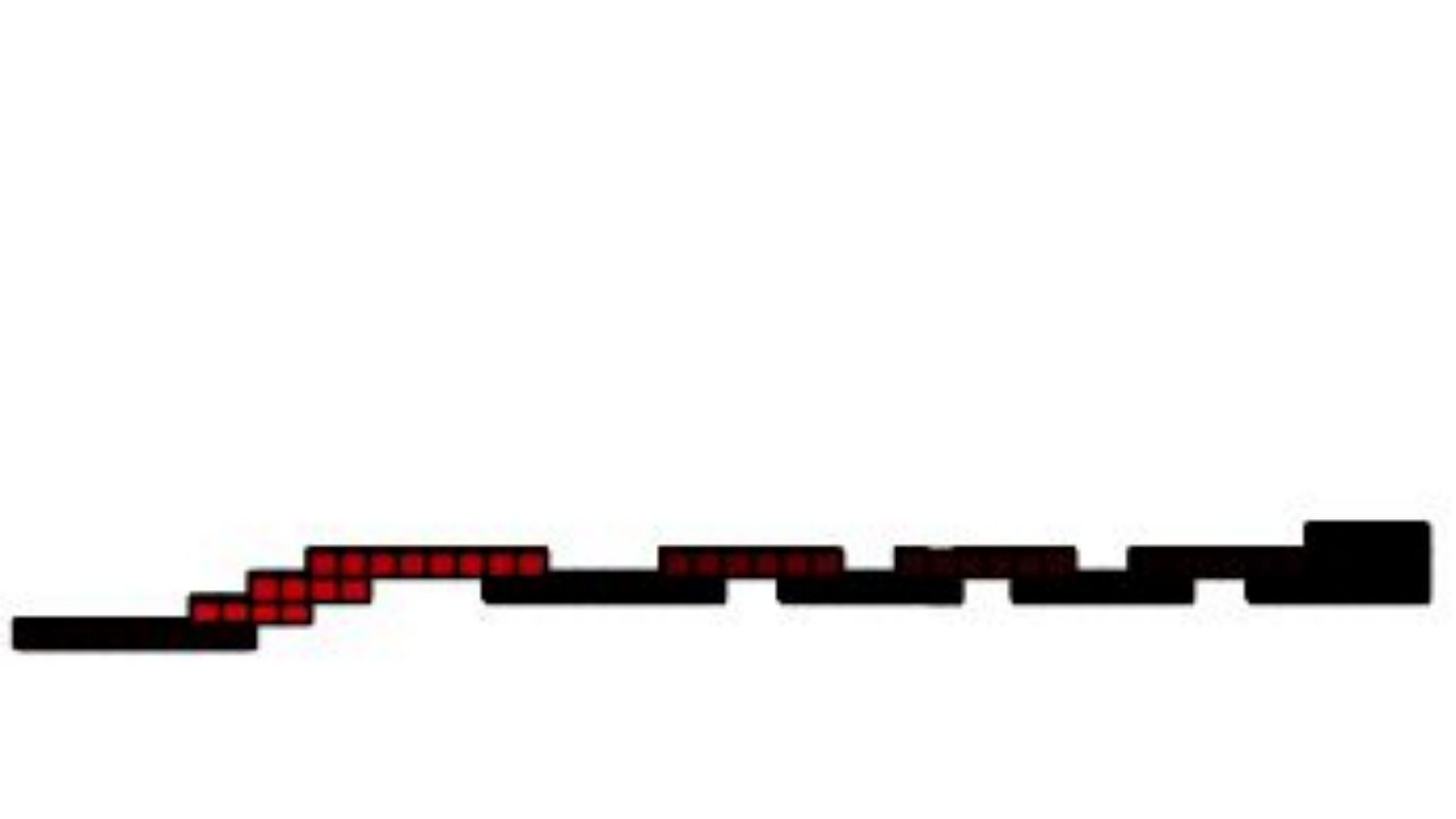}\label{fig:4_analysis}}\hfill
\subfigure[Ours \cref{fig:5}]{\includegraphics[width=0.16\linewidth]{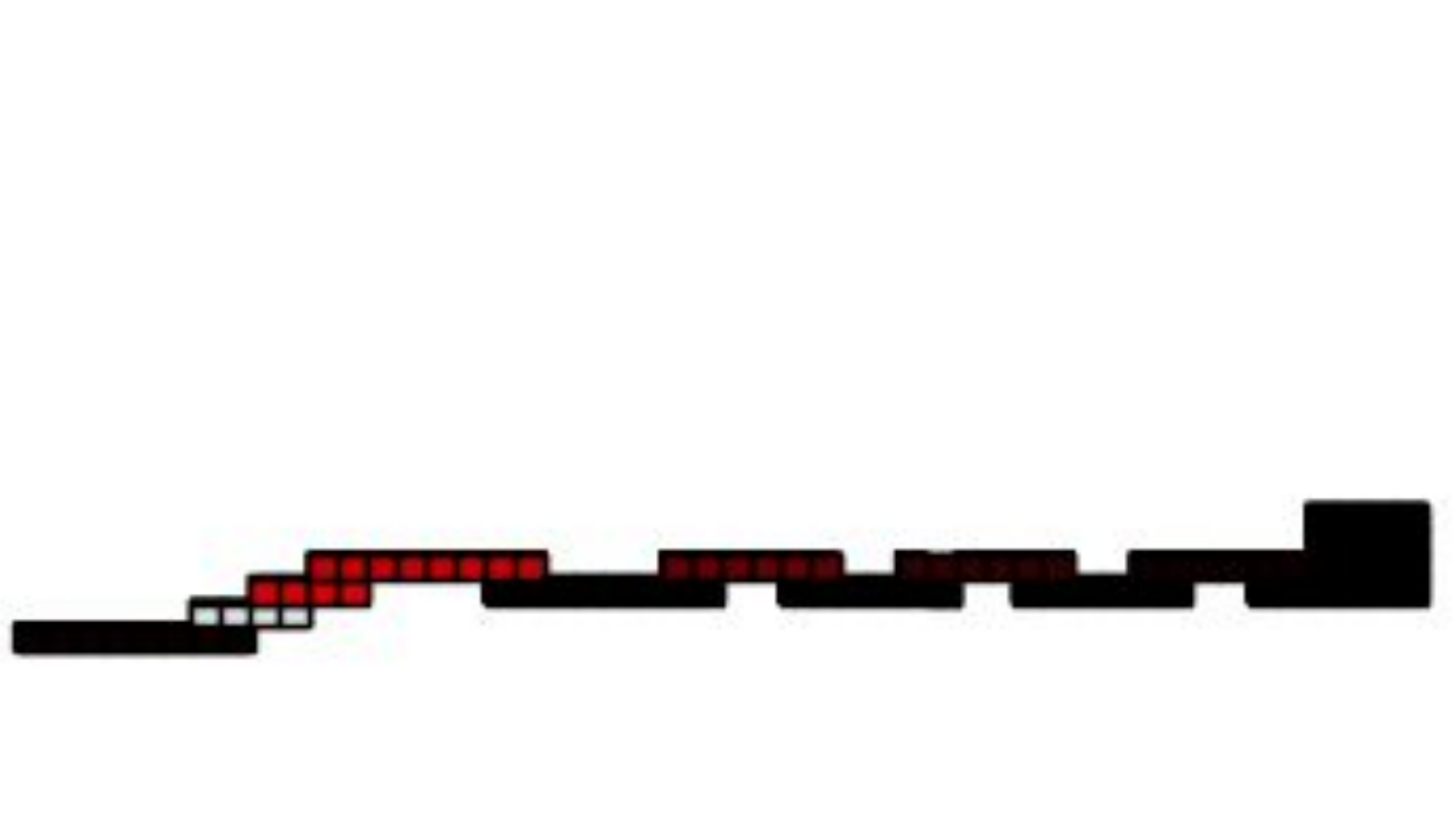}\label{fig:5_analysis}}\hfill
\subfigure[Ours \cref{fig:6}]{\includegraphics[width=0.16\linewidth]{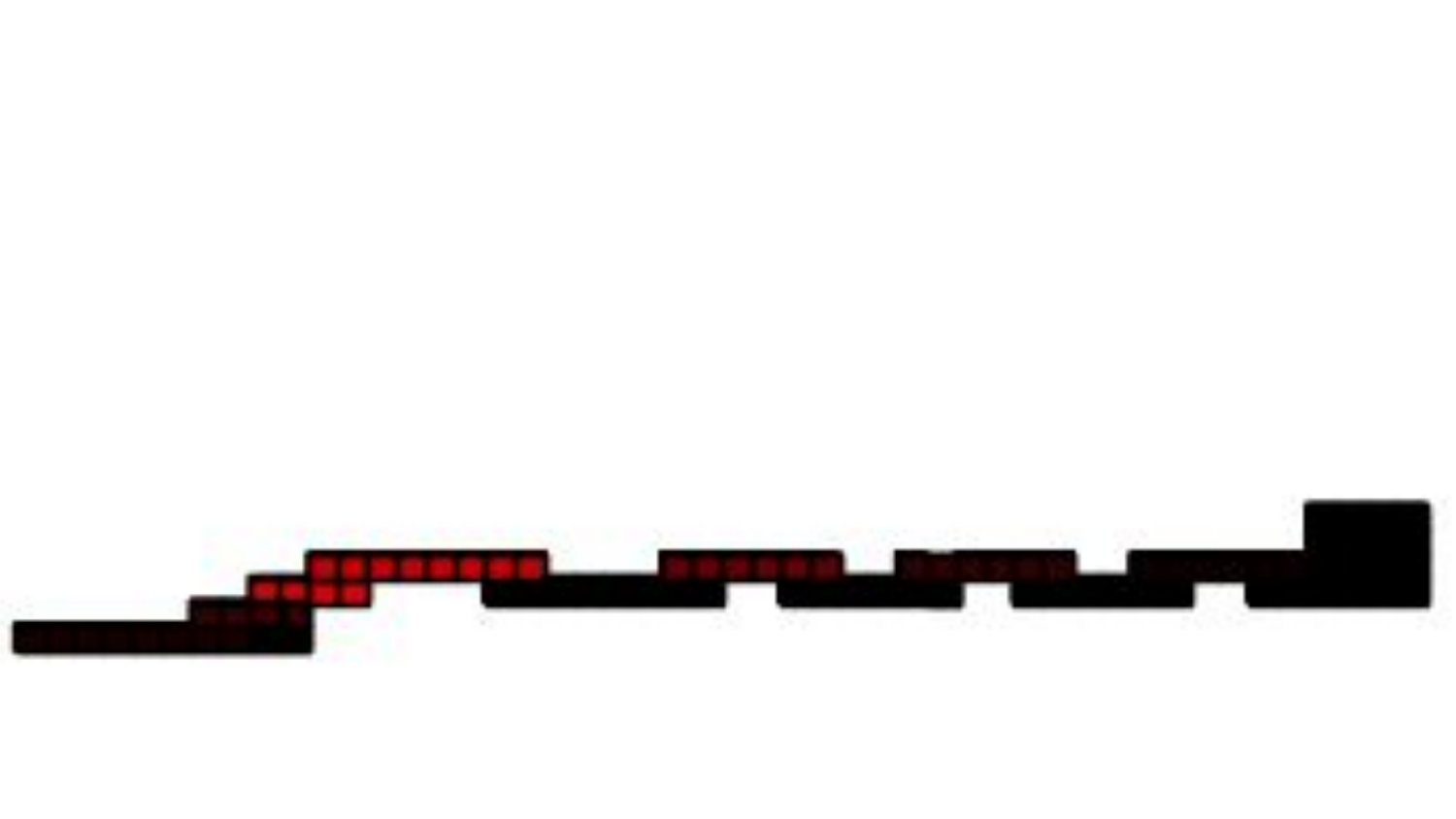}\label{fig:6_analysis}}
\\
\vspace{-5pt}
\subfigure[EB \cref{fig:1}]{\includegraphics[width=0.16\linewidth]{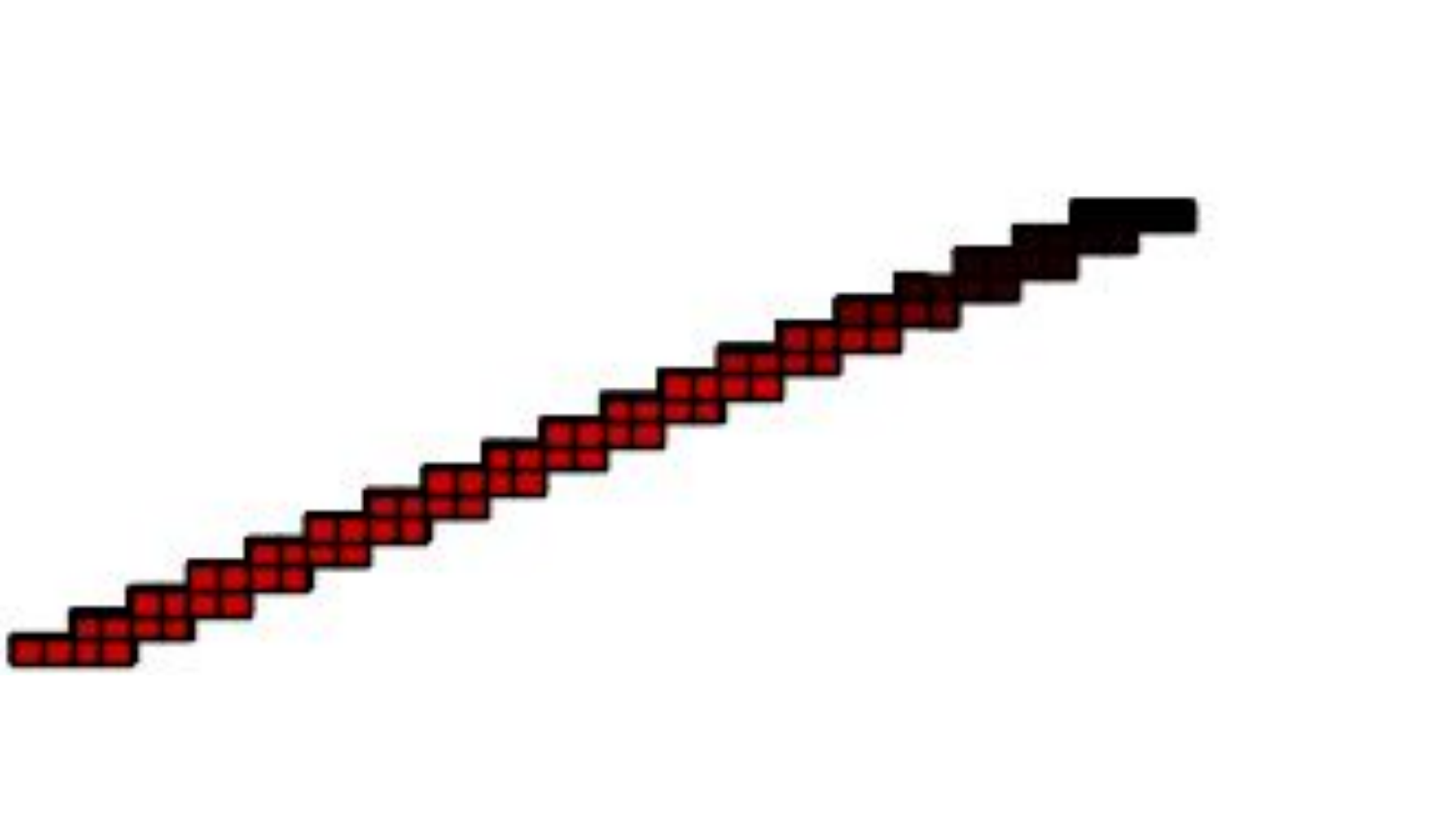}\label{fig:baseline_1_analysis}}\hfill
\subfigure[EB \cref{fig:2}]{\includegraphics[width=0.16\linewidth]{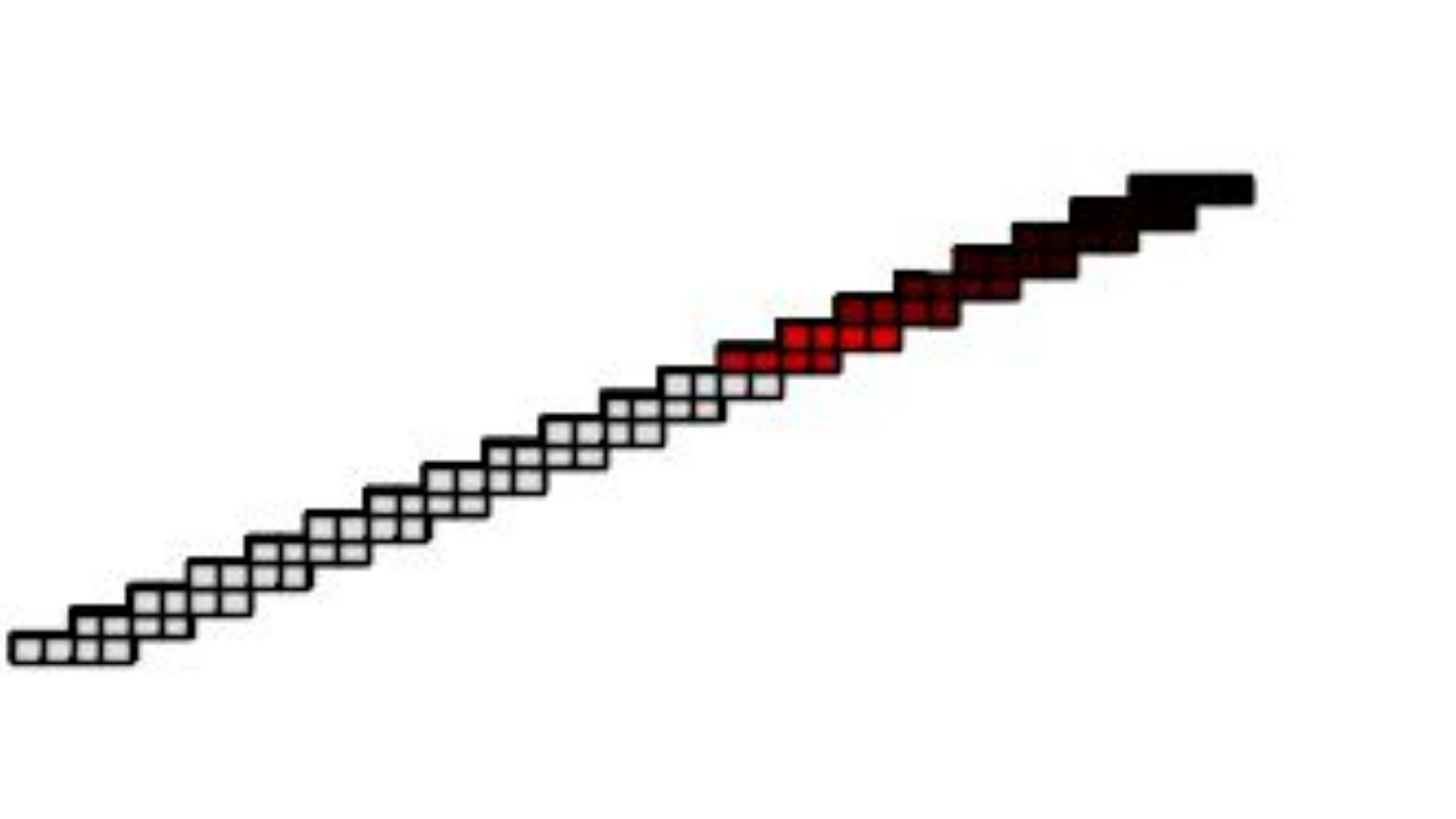}\label{fig:baseline_2_analysis}}\hfill
\subfigure[EB \cref{fig:3}]{\includegraphics[width=0.16\linewidth]{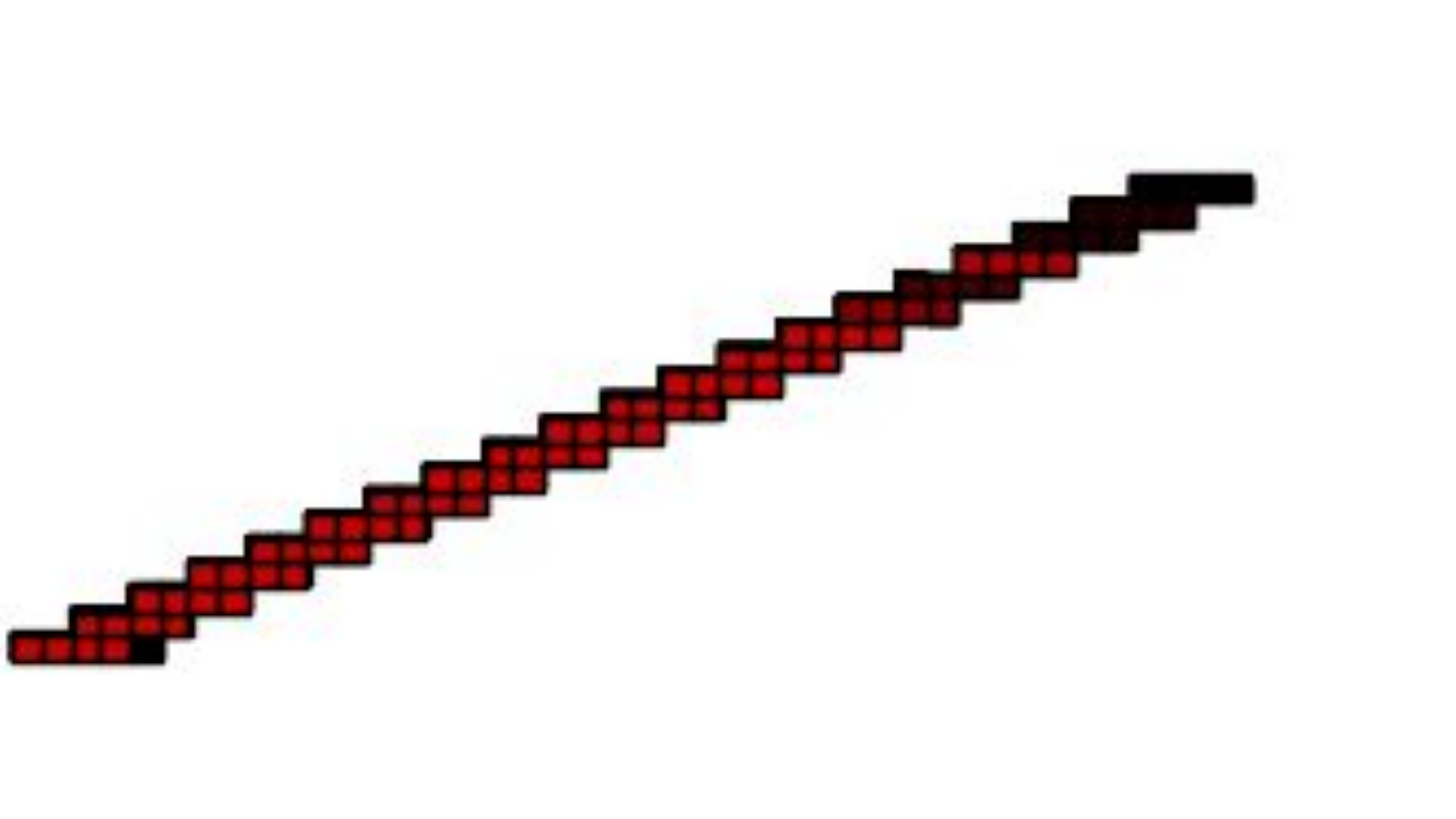}\label{fig:baseline_3_analysis}}\hfill
\subfigure[EB \cref{fig:4}]{\includegraphics[width=0.16\linewidth]{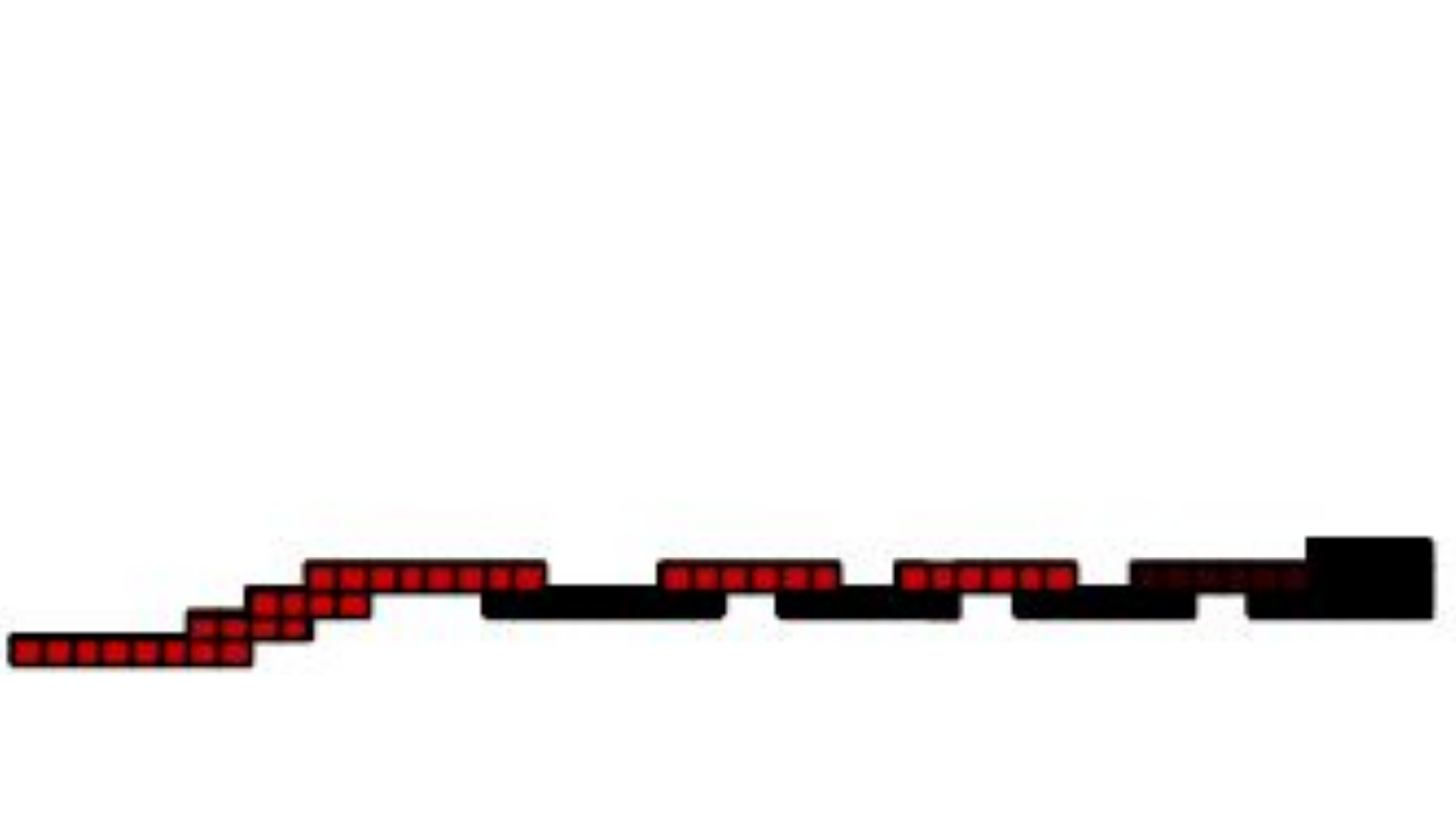}\label{fig:baseline_4_analysis}}\hfill
\subfigure[EB \cref{fig:5}]{\includegraphics[width=0.16\linewidth]{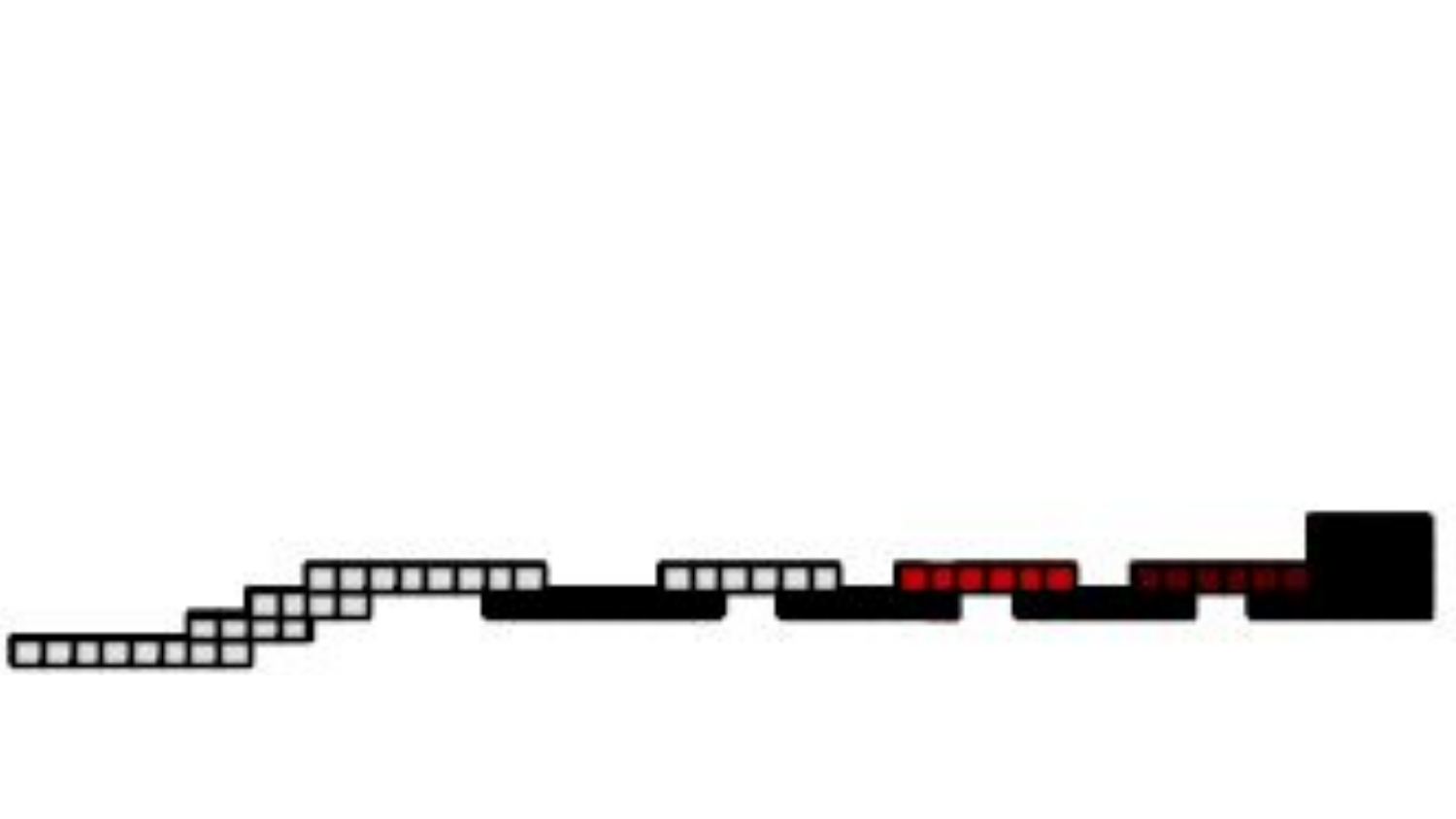}\label{fig:baseline_5_analysis}}\hfill
\subfigure[EB \cref{fig:6}]{\includegraphics[width=0.16\linewidth]{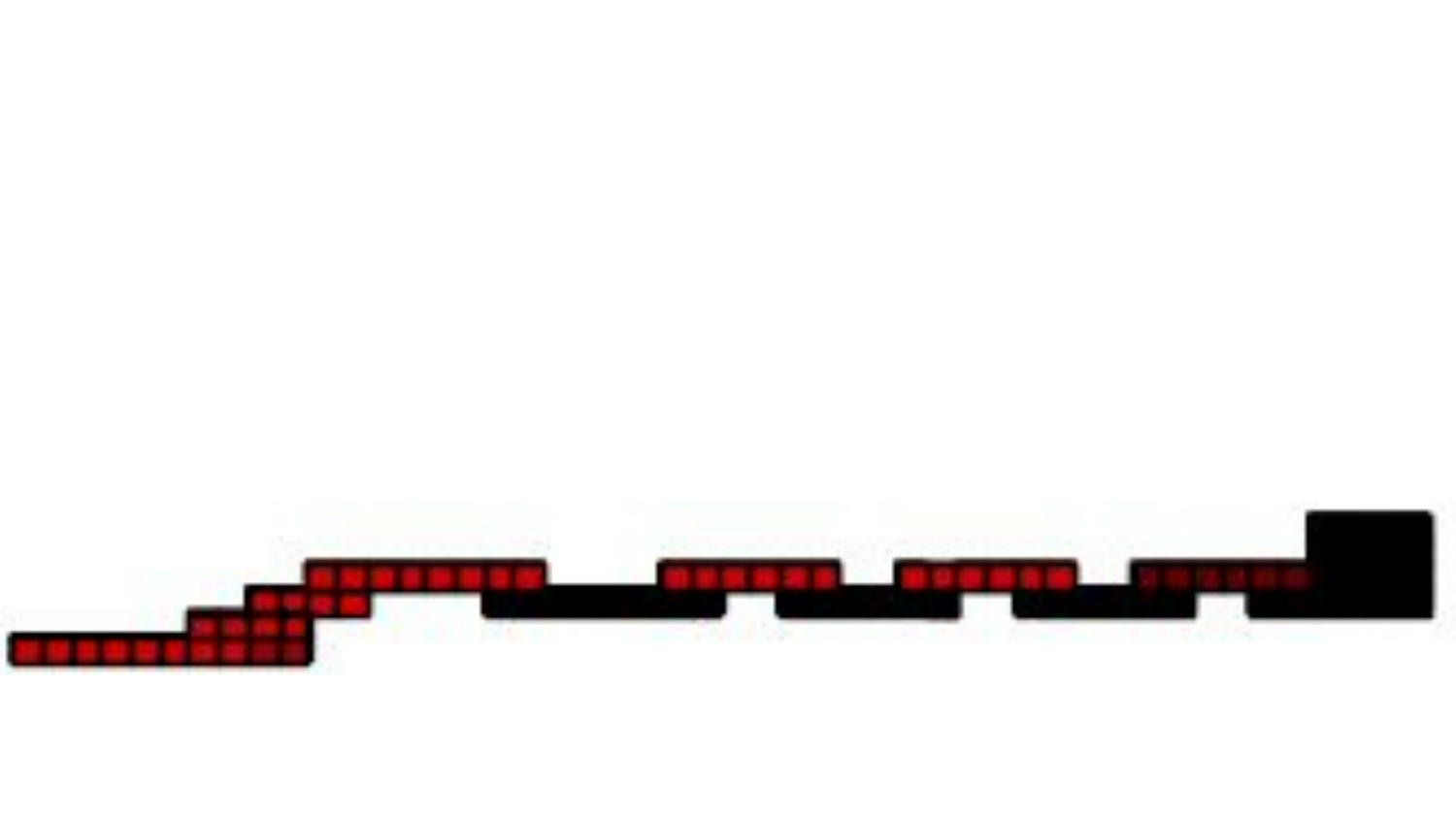}\label{fig:baseline_6_analysis}}
\vspace{-5pt}
    \caption{\footnotesize Comparison of the proposed stability analysis (\ie first row) and the EB (\ie second row). The analysis results from left to right correspond to the structures shown in \cref{fig:1,fig:2,fig:3,fig:4,fig:5,fig:6}. Black: bricks with less internal stress; Red: bricks experiencing higher stress; White: collapsing bricks.  \label{fig:accuracy}}
    \vspace{-15pt}
    % \vspace{-15pt}
\end{figure*}

\paragraph{Friction Capacity}
% Due to the rigid body assumption, it is reasonable to have $S, P\geq0$.
As shown in the left diagram of \cref{fig:force_model}, Lego bricks are held together due to the static friction (\ie $U$ and $D$) at the contact points caused by deformation.
The structure is stable if the friction is within the limit.
In our analysis, we assume all deformations are identical and all frictions share the same limit $T$.
A structure is stable if all friction forces do not exceed the limit.
Thus, we have the capacity constraint as
\begin{equation}\label{eq:capacity}
   C_i^T: \begin{cases}
       0 \leq U_i^{j_U} \leq T, ~\forall j_U\in[1, M_{U_i}]\\
       0 \leq D_i^{j_D} \leq T, ~\forall j_D\in[1, M_{D_i}]
   \end{cases}, ~ \forall i\in[1, N].
\end{equation}

\subsection{Stability Analysis Formulation}
Following the intuition in RBE \cite{10.1145/2816795.2818091}, a given structure is stable if there exists a set of forces $\mathbf{F}$ that satisfies \cref{eq:force_eq,eq:torque_eq,eq:non-neg,eq:non-coexist,eq:equality,eq:capacity}.
We can use the force distribution to estimate the stability of the structure.
%as well as improve the structure.
To solve $\mathbf{F}$, we formulate the optimization as
\begin{equation}\label{eq:optimization}
    \begin{split}
        \argmin_{\mathbf{F}} &\sum_{i=1}^N \Biggl\{ |C_i^f| + |C_i^{\tau}| + \alpha D_i^{max}+\beta \sum_{j=1}^{M_{D_i}} D_i^j  \Biggl\},\\
        &\text{subject to:} \begin{cases}
            C_i^{+}\\
            C_i^{||}\\
            C^=
        \end{cases}, \forall i\in[1, N].
    \end{split}
\end{equation}
where $D_i^{max}=\max_jD_i^j$ is the maximum dragging force for a brick $B_i$.
The objective function minimizes the static equilibrium values in \cref{eq:force_eq,eq:torque_eq} as well as the maximum friction and the total friction in each brick.
The terms $|C_i^f|$ and $|C_i^{\tau}|$ encourage the solver to solve a distribution of $\mathbf{F}$ that makes the structure to reach static equilibrium. 
$D_i^{max}$ tries to avoid extreme values among the dragging forces in $B_i$.
And $\sum_{j=1}^{M_{D_i}} D_i^j$ encourages the solver to solve $\mathbf{F}$ with minimum internal friction.
$\alpha$ and $\beta$ are tunable weights to adjust the influence of the two terms so that they do not take over the effect of the static equilibrium.
Note that the key difference between \cref{eq:optimization} and previous works is that instead of imposing static equilibrium \cref{eq:force_eq,eq:torque_eq} as equality constraints, we encode them in the objective function. 
This is critical since enforcing them as constraints is essentially assuming there exists a $\mathbf{F}$ that satisfies the static equilibrium.
If a given structure does not have such a $\mathbf{F}$, the formulation is voided.
An example could be a structure with floating bricks.
Including them in the objective function instead of as hard constraints can relax the single-connected assumption and solve the stability of any structures.
Aside from the objective function, \cref{eq:optimization} also imposes more equality constraints (\ie \cref{eq:non-coexist,eq:equality}) than prior works to improve the accuracy of predicted stability.

Given the solved $\mathbf{F}$, the stability of each brick $B_i$ is estimated as 
\begin{equation}\label{eq:score}
        V_i=\begin{cases}
            1 & \neg C_i^f \vee \neg C_i^{\tau} \vee \neg C_i^T\\
            1-\frac{\ceil{T - D_i^{max}}}{T} & \text{Otherwise}
        \end{cases}
\end{equation}
% where $\alpha$ and $\beta$ are tunable parameters.
The structure is stable if all bricks are stable, \ie $0\leq V_i<1, \forall i\in[1, N]$.
It is worth noting that the friction capacity \cref{eq:capacity} is not imposed as a constraint in \cref{eq:optimization}.
Instead, we add the friction terms in the objective function to minimize the solved internal friction and use \cref{eq:capacity} in \cref{eq:score} to determine the structural stability.

\section{Experiments}

Our experiment considers standard Lego bricks, \ie rectangular bricks with solid colors and a height of $9.6 \text{mm}$.
Bricks with dimension $Q\times X$ are uncommon on the market, and thus, we mainly consider $1\times X$ and $2\times X$ bricks.
\Cref{table:brick_weight} shows the masses for each brick.
To avoid uncertainty in manufacturing, each brick's mass is measured using the average of 10 bricks.
We use bricks that are highlighted in \cref{table:brick_weight} since they are most commonly used.
Our stability analysis is implemented using Python and Gurobi \cite{gurobi}.
We have $T=0.98 \text{N}$, $\alpha=10^{-3}$ and $\beta=10^{-6}$.
Our implementation is available at \url{https://github.com/intelligent-control-lab/StableLego}.
All results are generated on an Intel i7-13700HX with 32GB RAM.

\begin{table}
\centering
\begin{tabular}{c  c  c  c c } 
\hline
Dimension & $\mathbf{1\times 1}$ & $\mathbf{1\times 2}$ & $\mathbf{1\times 4}$ & $1\times 6$  \\    
Mass (g) &  0.43 & 0.81 & 1.57 &  2.28 \\
\hline
Dimension &   $1\times 8$ & $\mathbf{2\times 2}$ & $\mathbf{2\times 4}$ & $\mathbf{2\times 6}$ \\
Mass (g) & 3.03 & 1.15 & 2.16 & 3.23 \\
\hline
\end{tabular}
\caption{\footnotesize The masses of different Lego bricks with different dimensions. Highlighted bricks are the ones used in the experiment. \label{table:brick_weight}}
\vspace{-15pt}
\end{table}

\subsection{Stability Analysis Accuracy}
We implement Luo et al. \cite{10.1145/2816795.2818091} as our baseline. 
However, the original formulation in \cite{10.1145/2816795.2818091} only considers \cref{eq:force_eq,eq:torque_eq,eq:non-neg,eq:capacity}.
Thus, we implement an enhanced version of \cite{10.1145/2816795.2818091} as the \textit{enhanced baseline (EB)} by integrating \cref{eq:non-coexist,eq:equality}.
We first evaluate our stability analysis algorithm on several hand-crafted Lego structures as shown in \cref{fig:block_examples}.
\Cref{fig:accuracy} illustrates the comparisons between our analysis results (\ie the top row) and the EB's prediction (\ie the bottom row).
% The top row shows the stability results by solving \cref{eq:optimization}, and the bottom row displays the results generated by the EB.
We do not include the predictions from the original baseline because it fails to distinguish the structures and predicts that all six structures are stable.

\Cref{fig:1_analysis,fig:baseline_1_analysis} correspond to the structure in \cref{fig:1}.
The structure can be built in real, and both methods indicate that the structure is stable.
However, our method indicates higher internal stress at lower levels, whereas the EB cannot distinguish the stresses at different levels.
The structure in \cref{fig:2} collapses, and both methods indicate that the structure is unstable.
However, ours accurately predicts the collapsing point (\ie the white brick in \cref{fig:2_analysis}) while the EB cannot as shown in \cref{fig:baseline_2_analysis}.
To observe the actual collapsing point, we hold the structure before it is finished so it does not collapse during construction.
After all connections are established, we remove the external support and observe the collapsing point.
As shown in \cref{fig:2}, the structure collapses at the predicted location.
Similar results are shown in \cref{fig:3_analysis,fig:4_analysis,fig:5_analysis,fig:6_analysis,fig:baseline_3_analysis,fig:baseline_4_analysis,fig:baseline_5_analysis,fig:baseline_6_analysis}, which correspond to the structures in \cref{fig:3,fig:4,fig:5,fig:6}.
We can see that even though both methods can estimate structural stability, ours gives a more precise estimation, which generates more realistic internal stress distributions and predicts the weakest connection points.
% This is important since it provides insight into improving the brick layout design to construct a stable structure as shown in \cref{fig:block_examples}.

\subsection{StableLego Dataset}
\label{sec:dataset}
A large-scale dataset is essential for benchmarking various assembly tasks.
However, it is time-consuming, if not impossible, to design a large number of different Lego objects manually.
To this end, we present \textit{StableLego}, a comprehensive dataset that provides artificially generated Lego brick layouts for a wide variety of different 3D objects. 
StableLego is developed based on the ShapeNetCore dataset \cite{shapenet2015}.
It includes more than $50$k of different objects from 55 common object categories with their Lego layouts.
For each object, we downsample the original 3D object to a $20\times 20\times 20$ grid world and generate a corresponding brick layout.
In particular, we merge unit voxels (\ie $1\times 1$) into larger bricks and prioritize merging voxels that have no support under them.
The dataset contains a mix of simple and complex structures, in which the simple ones have less than 5 bricks, whereas the complex ones can include up to more than 1100 bricks.
Note that the focus of this dataset is not providing optimal brick layouts. 
Thus, the dataset contains a mix of valid and invalid brick layouts for testing the stability analysis accuracy.
% The top and third rows in \cref{fig:dataset,fig:invalid_dataset} illustrate examples of the 3D objects and the corresponding Lego structures.
The dataset could be used to inspire creativity in building Lego objects.
More importantly, it provides a novel benchmarking platform for verifying the performance of structure stability algorithms as well as facilitating research in related areas.
We include the stability estimation using the proposed formulation for each Lego structure.
Prior work \cite{Li_2023_CVPR} provides a Lego assembly dataset with over 150 designs generated from video input.
To the best of our knowledge, StableLego is the \textit{first} large-scale Lego assembly dataset with stability inferences.

\begin{figure*}
\centering
\subfigure[]{\includegraphics[width=0.095\linewidth]{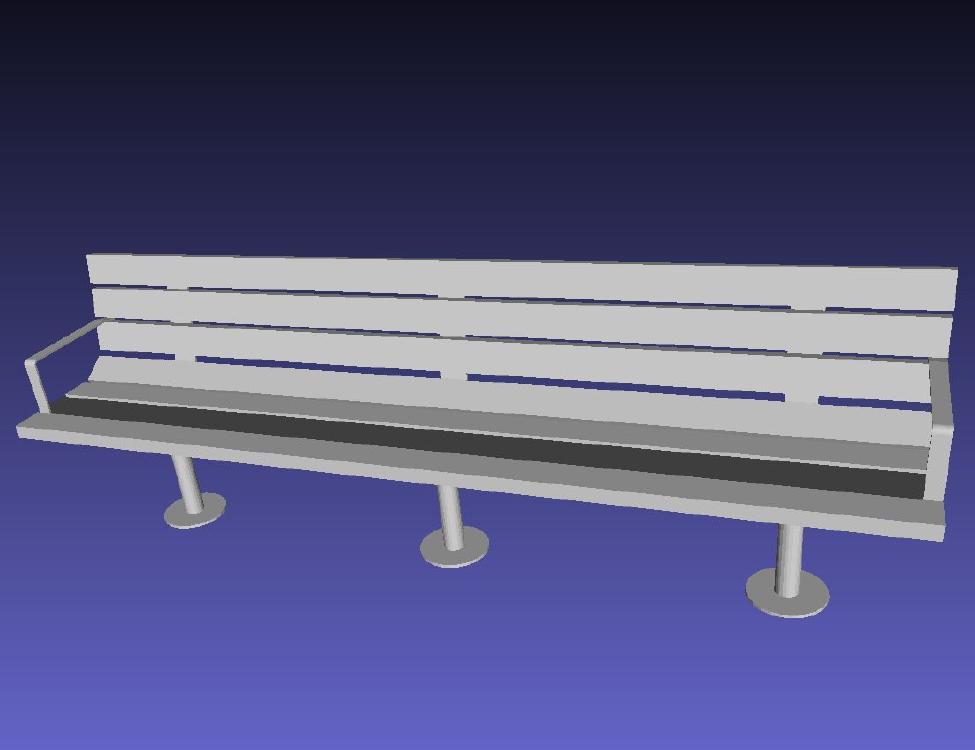}}\hfill
\subfigure[]{\includegraphics[width=0.095\linewidth]{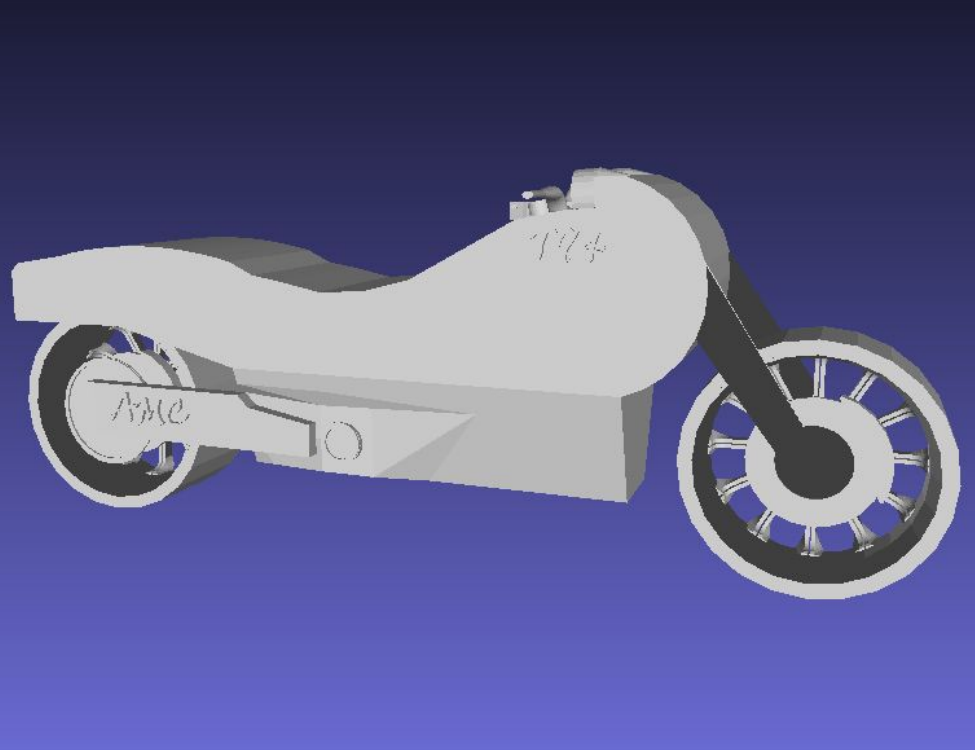}}\hfill
\subfigure[]{\includegraphics[width=0.095\linewidth]{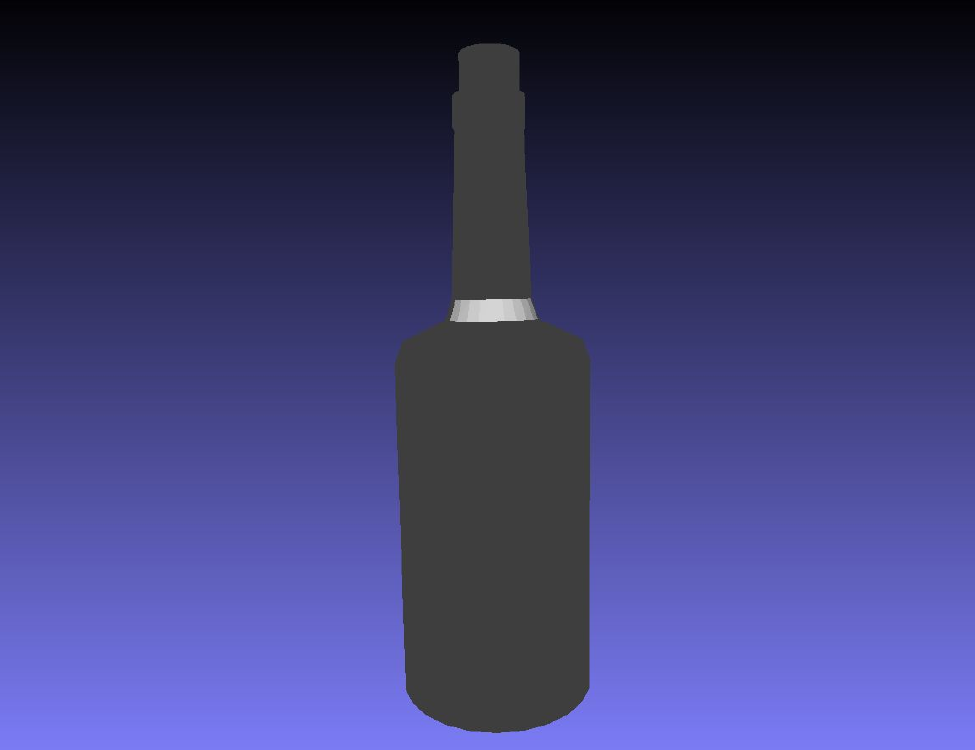}}\hfill
\subfigure[]{\includegraphics[width=0.095\linewidth]{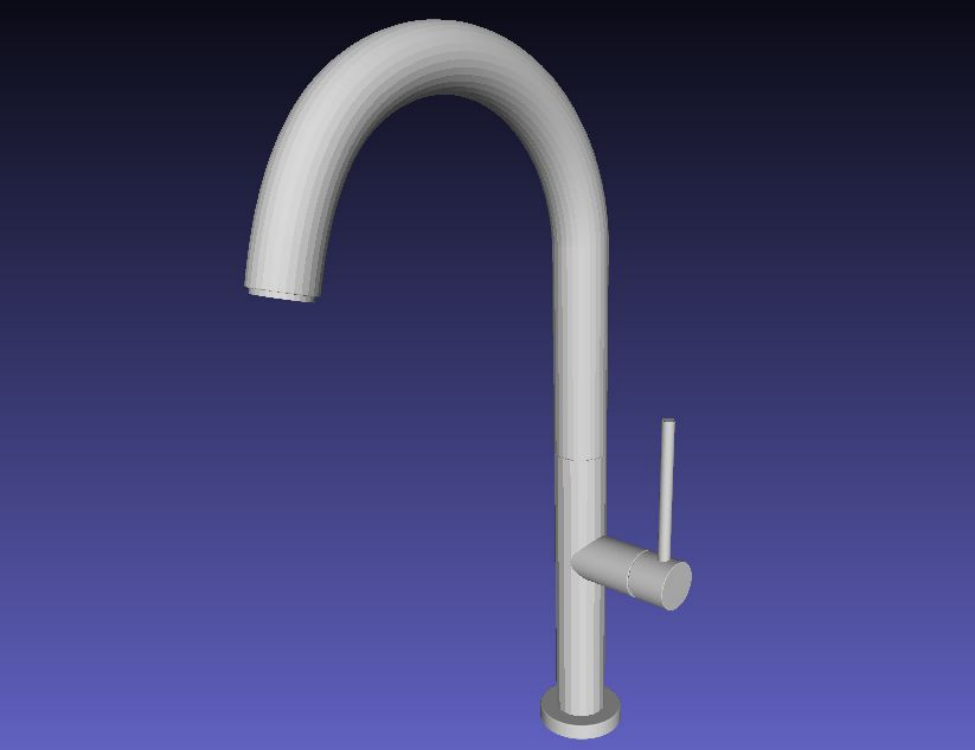}}\hfill
\subfigure[]{\includegraphics[width=0.095\linewidth]{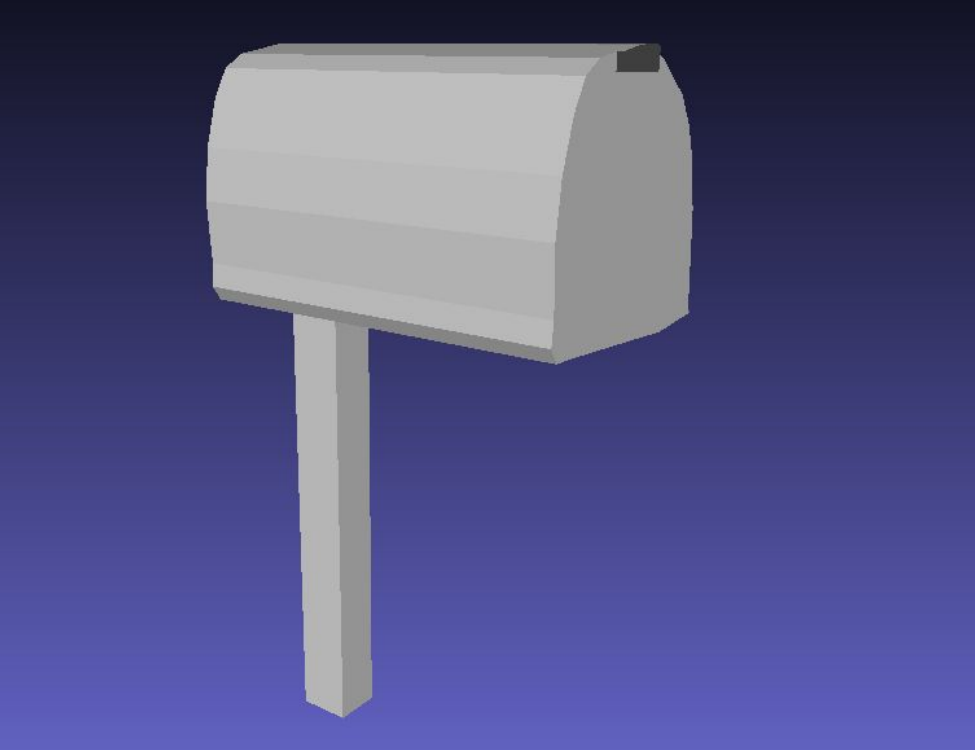}}\hfill
\subfigure[]{\includegraphics[width=0.095\linewidth]{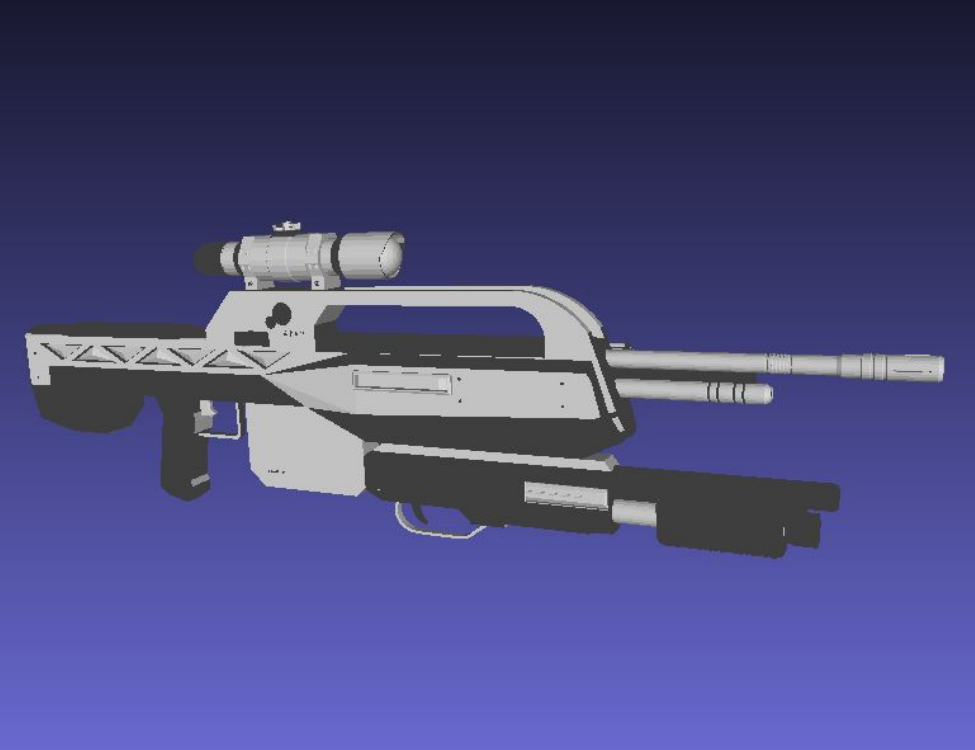}}\hfill
\subfigure[]{\includegraphics[width=0.095\linewidth]{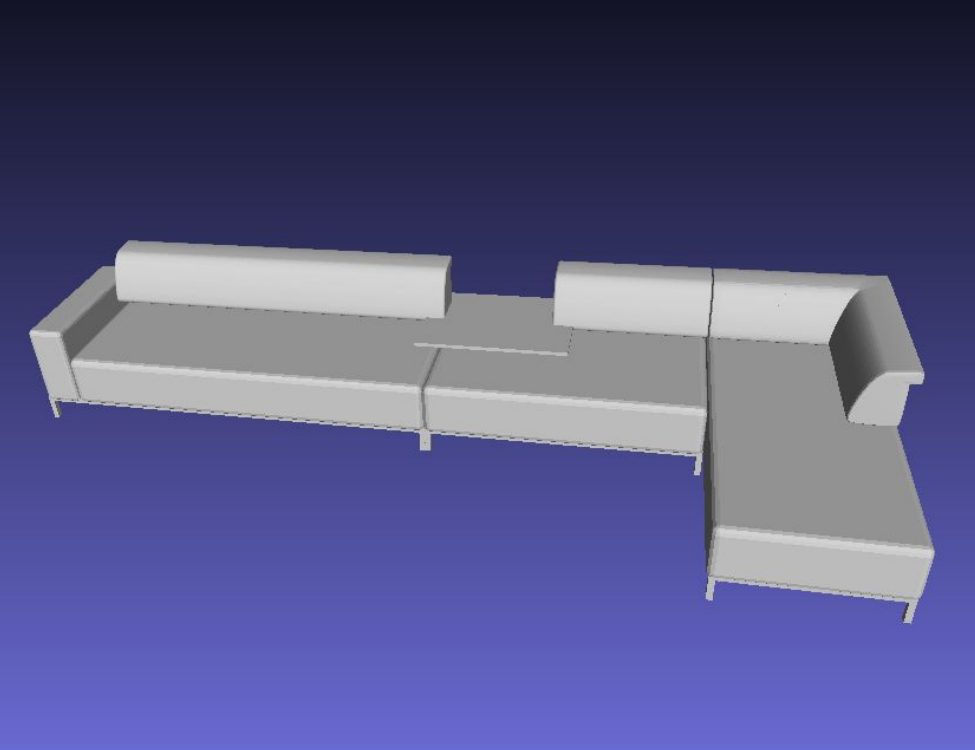}}\hfill
\subfigure[]{\includegraphics[width=0.095\linewidth]{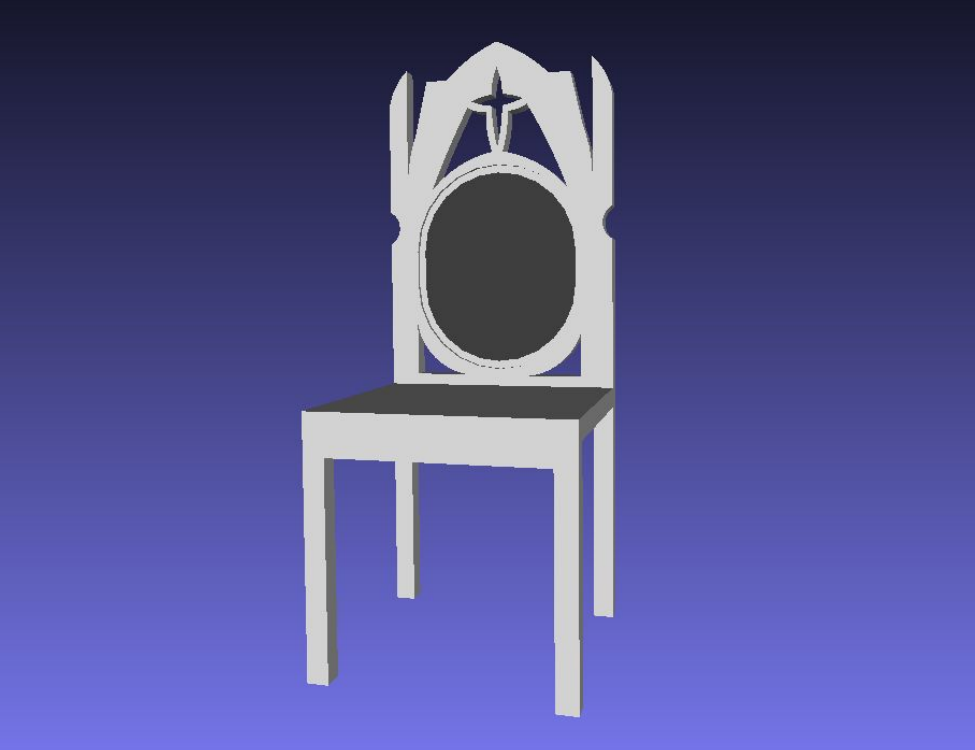}}\hfill
\subfigure[]{\includegraphics[width=0.095\linewidth]{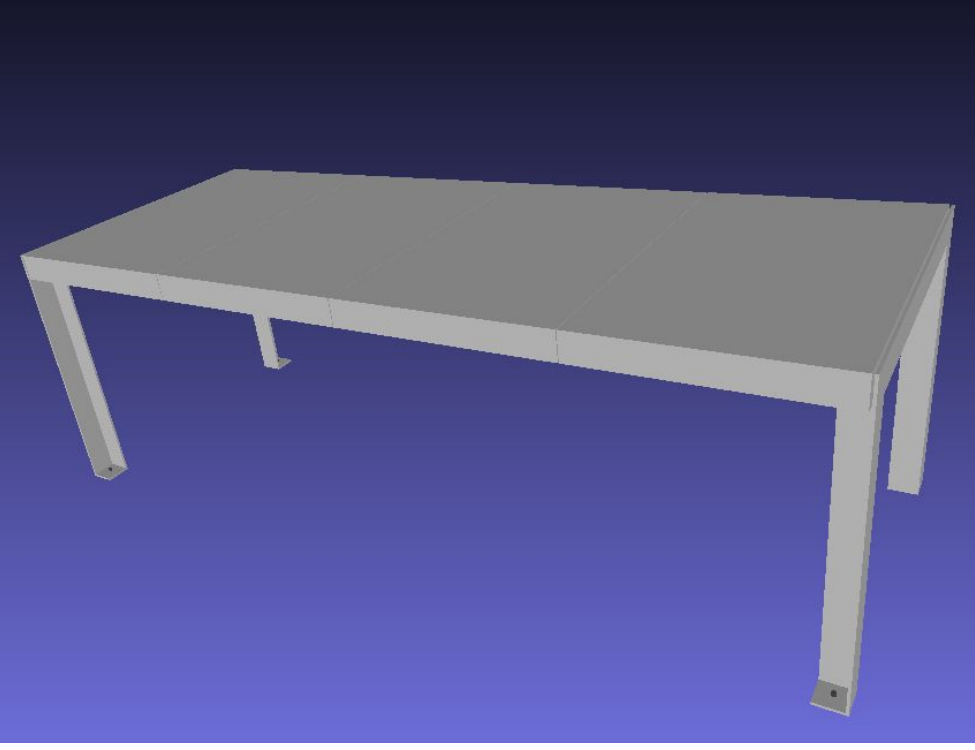}}\hfill
\subfigure[]{\includegraphics[width=0.095\linewidth]{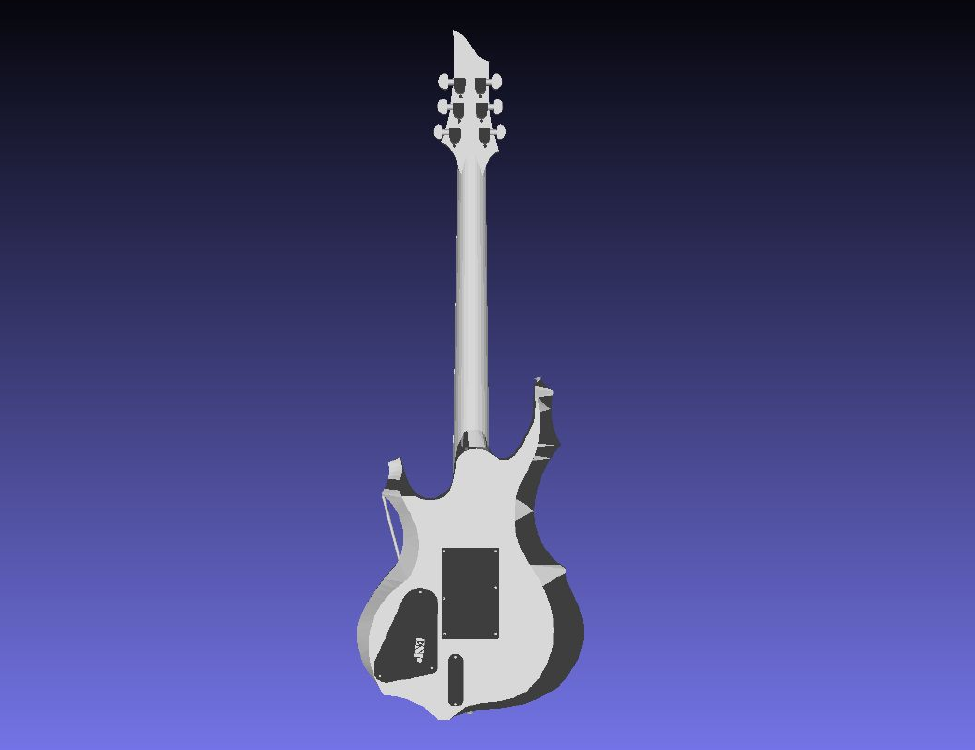}}
\\
\vspace{-5pt}
\subfigure[]{\includegraphics[width=0.095\linewidth]{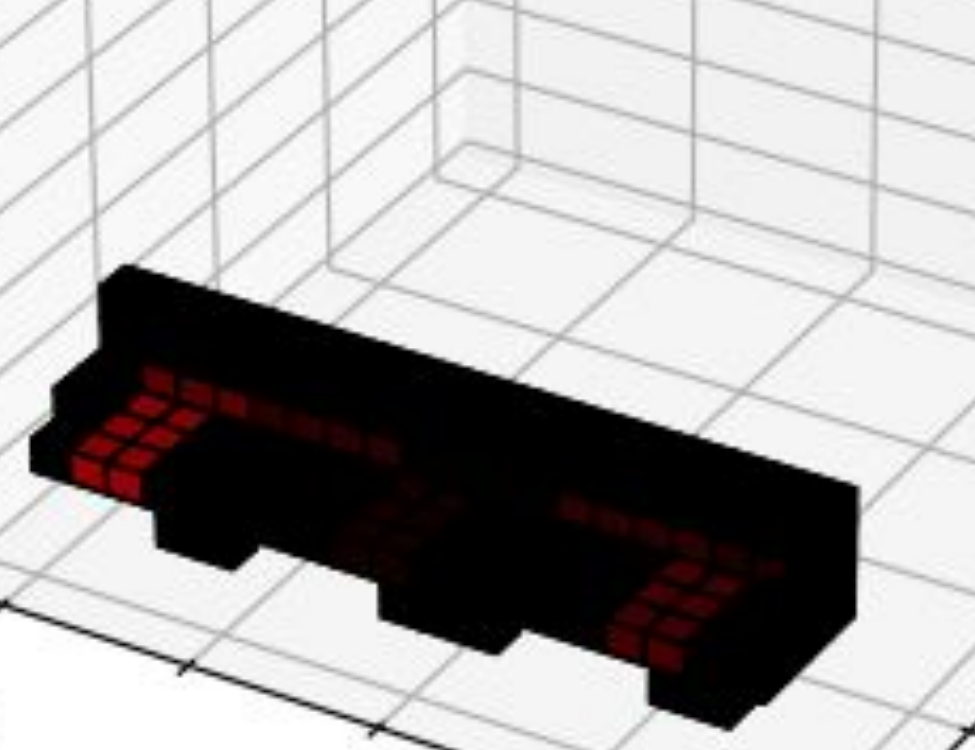}}\hfill
\subfigure[]{\includegraphics[width=0.095\linewidth]{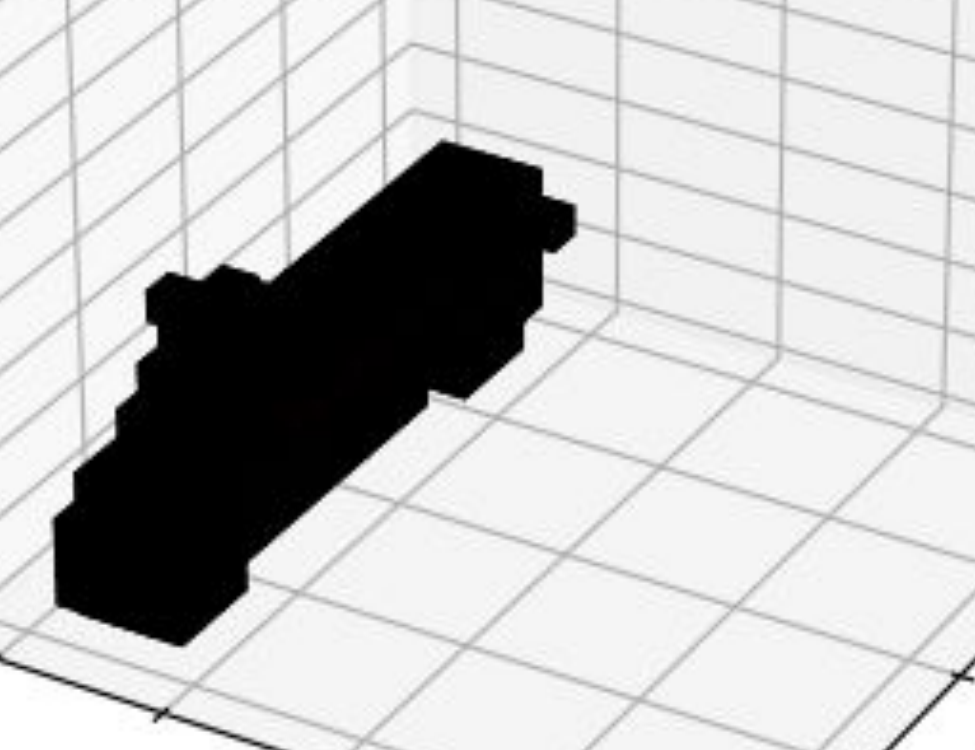}}\hfill
\subfigure[]{\includegraphics[width=0.095\linewidth]{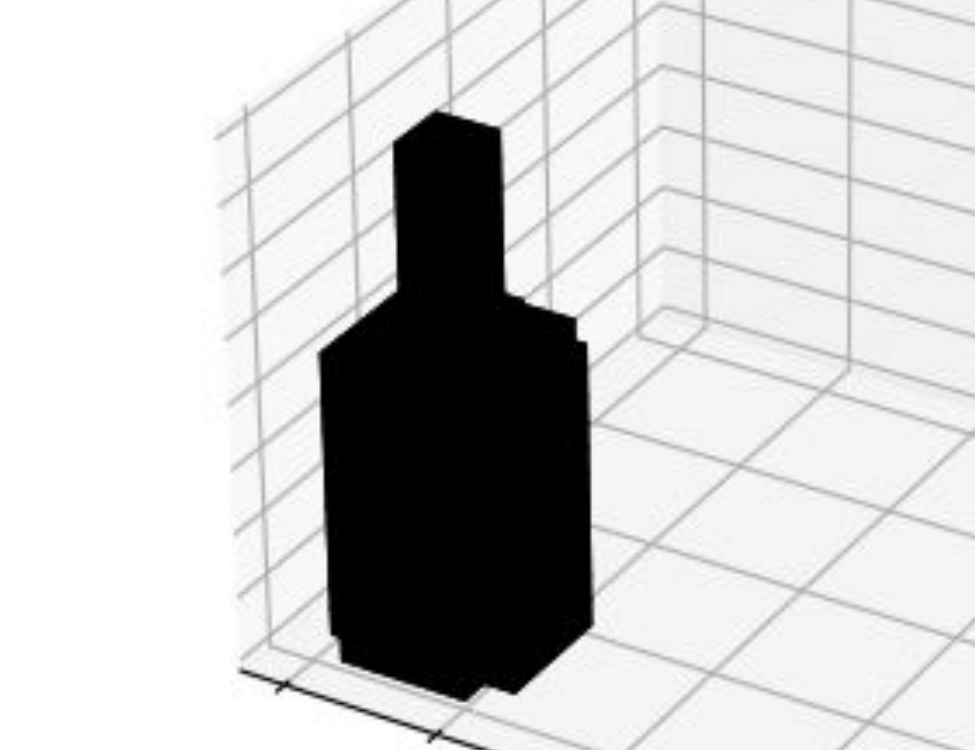}}\hfill
\subfigure[]{\includegraphics[width=0.095\linewidth]{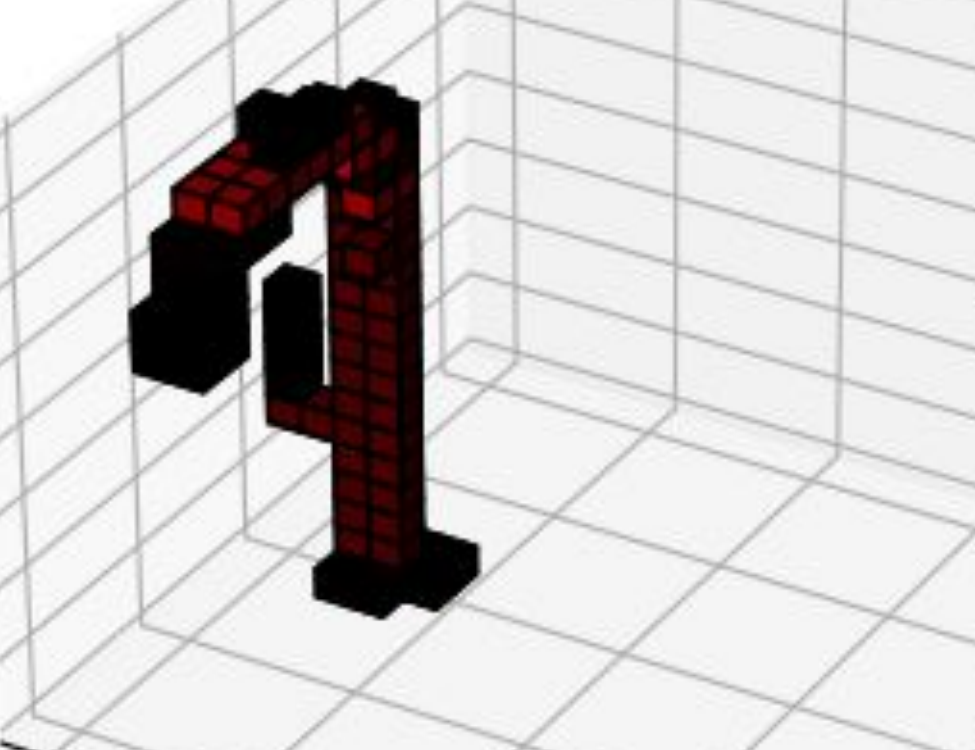}}\hfill
\subfigure[]{\includegraphics[width=0.095\linewidth]{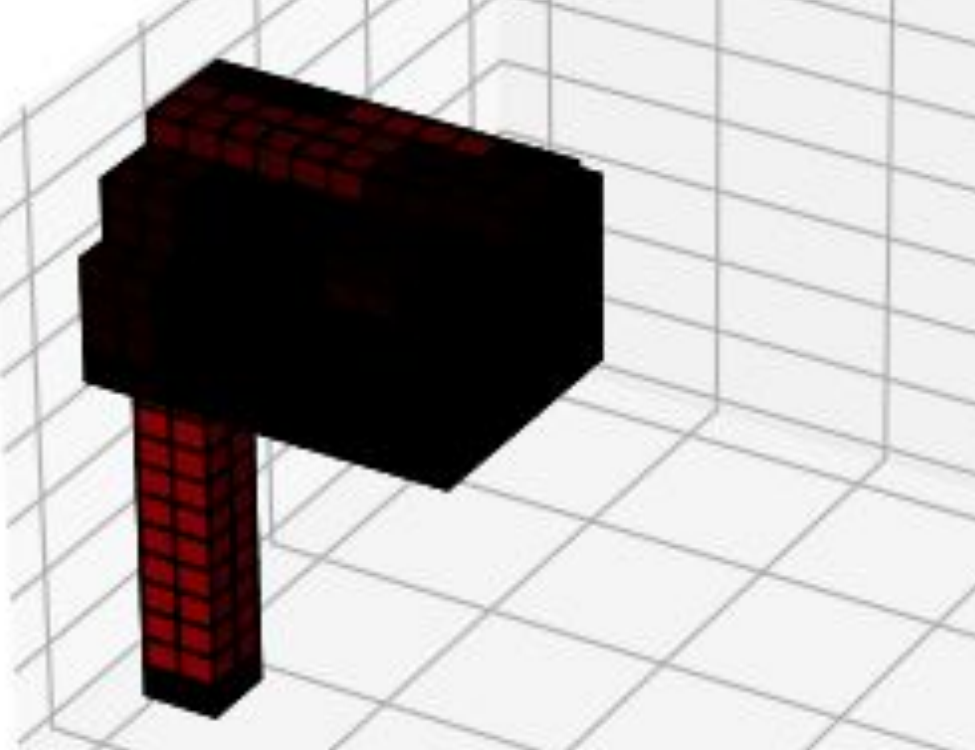}}\hfill
\subfigure[]{\includegraphics[width=0.095\linewidth]{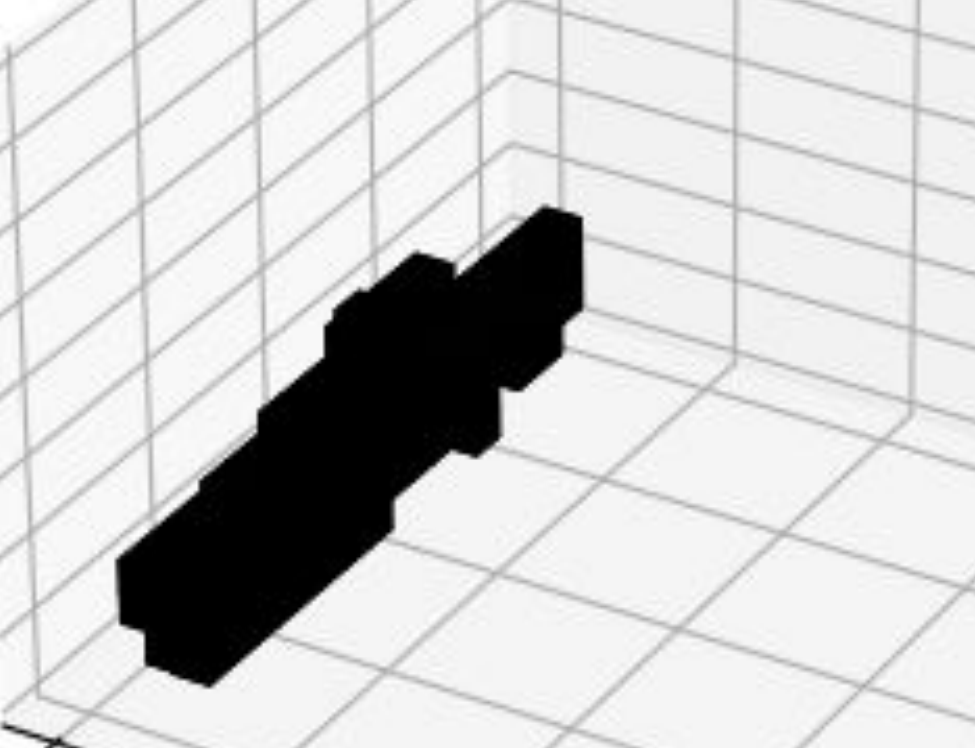}}\hfill
\subfigure[]{\includegraphics[width=0.095\linewidth]{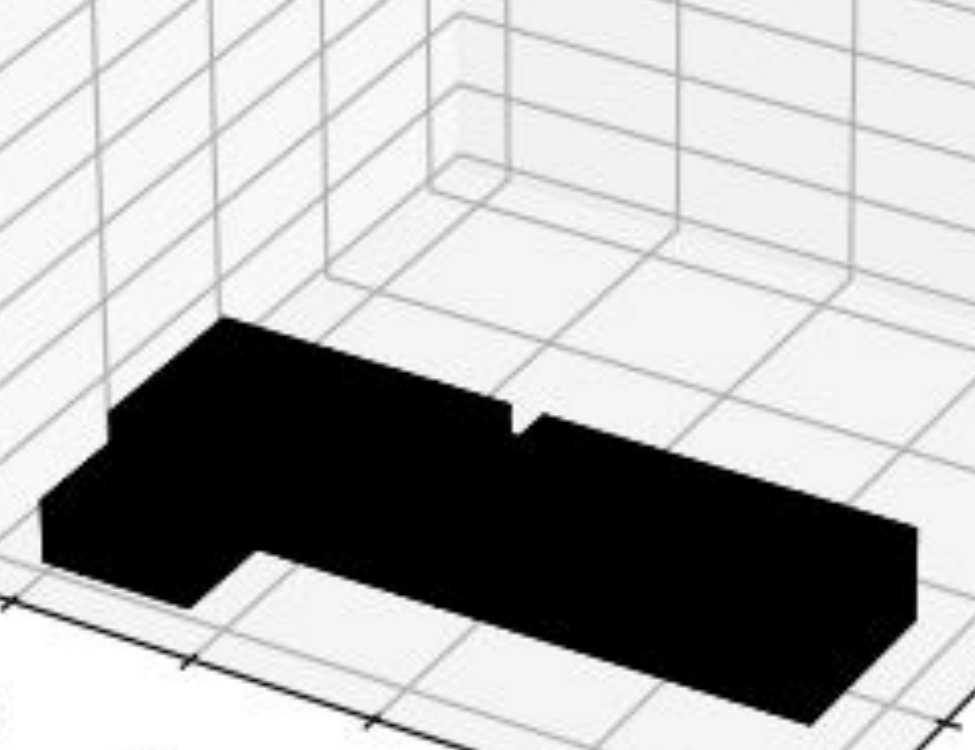}}\hfill
\subfigure[]{\includegraphics[width=0.095\linewidth]{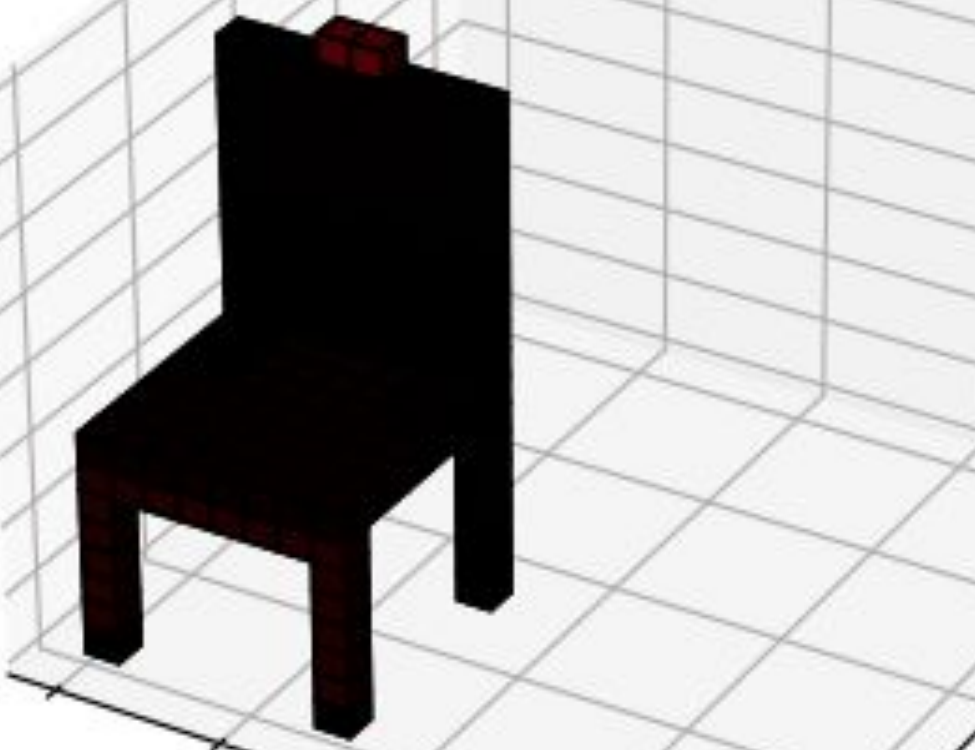}}\hfill
\subfigure[]{\includegraphics[width=0.095\linewidth]{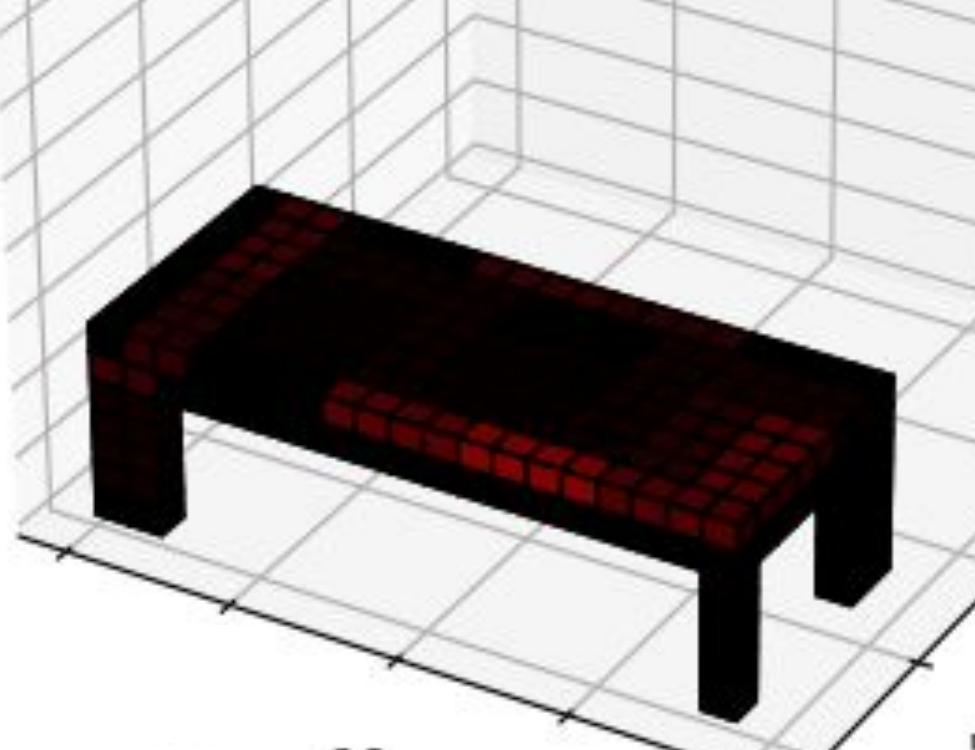}}\hfill
\subfigure[]{\includegraphics[width=0.095\linewidth]{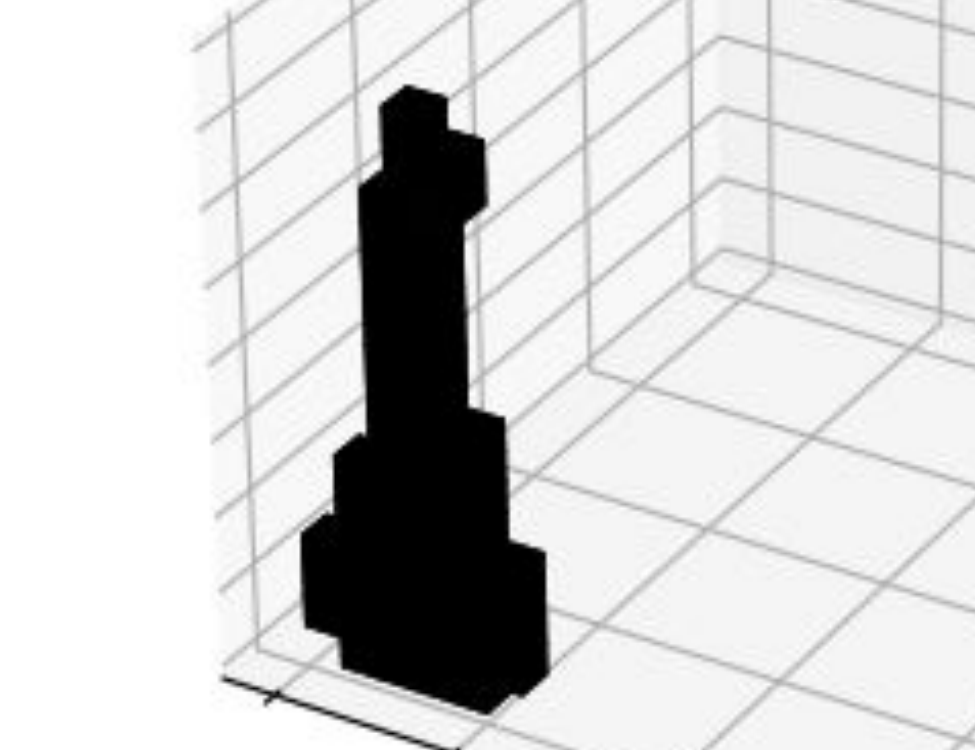}}
\vspace{-5pt}\\
\subfigure[]{\includegraphics[width=0.095\linewidth]{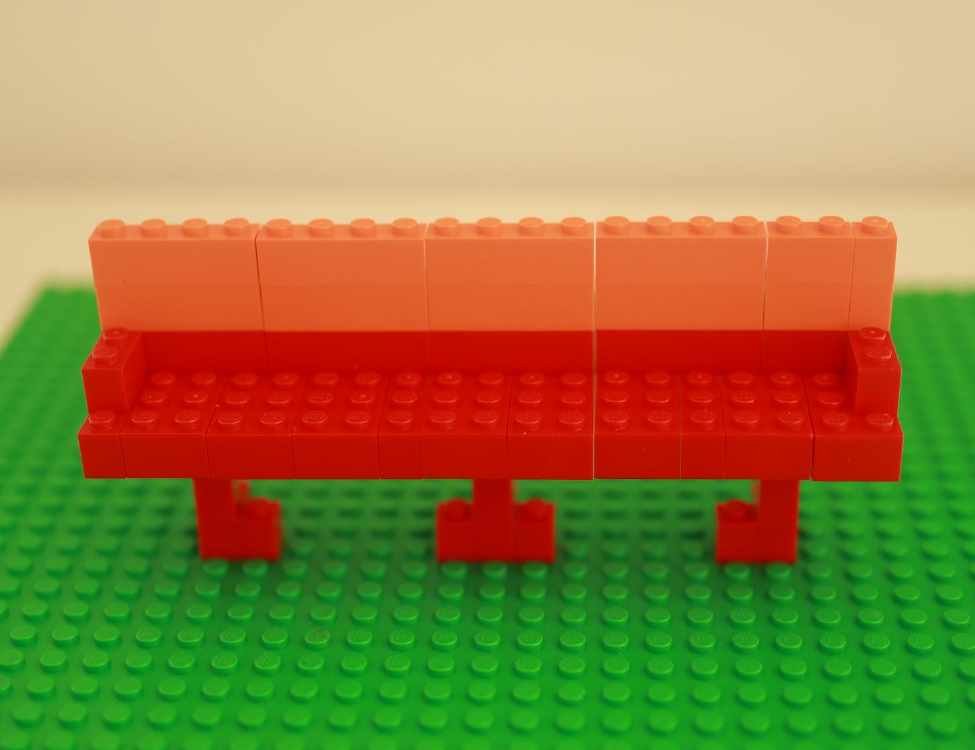}}\hfill
\subfigure[]{\includegraphics[width=0.095\linewidth]{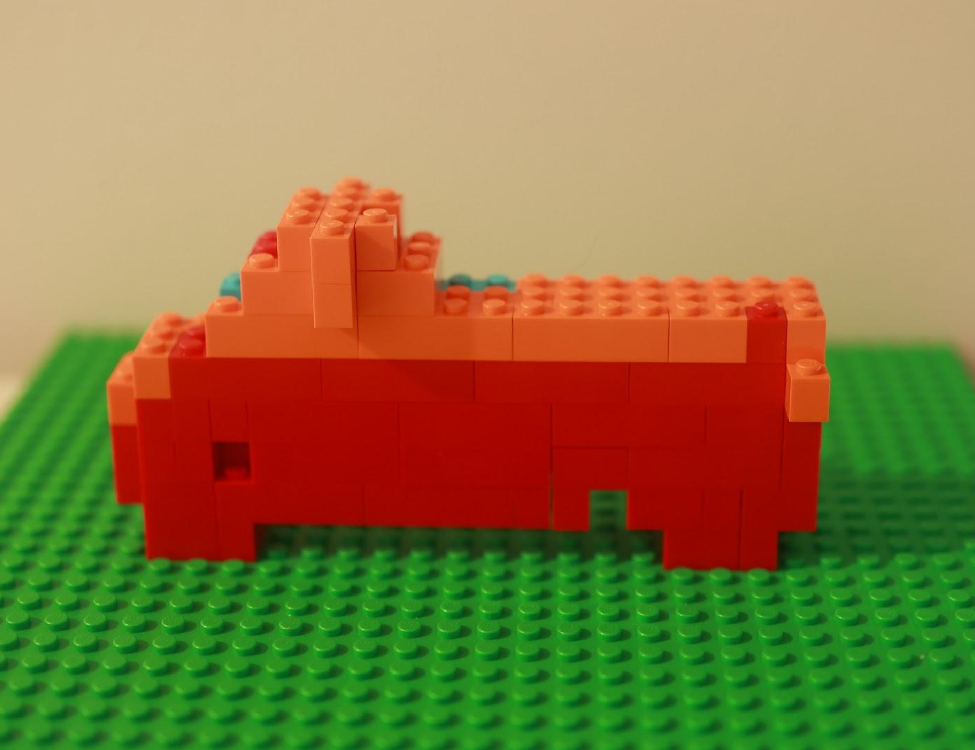}}\hfill
\subfigure[]{\includegraphics[width=0.095\linewidth]{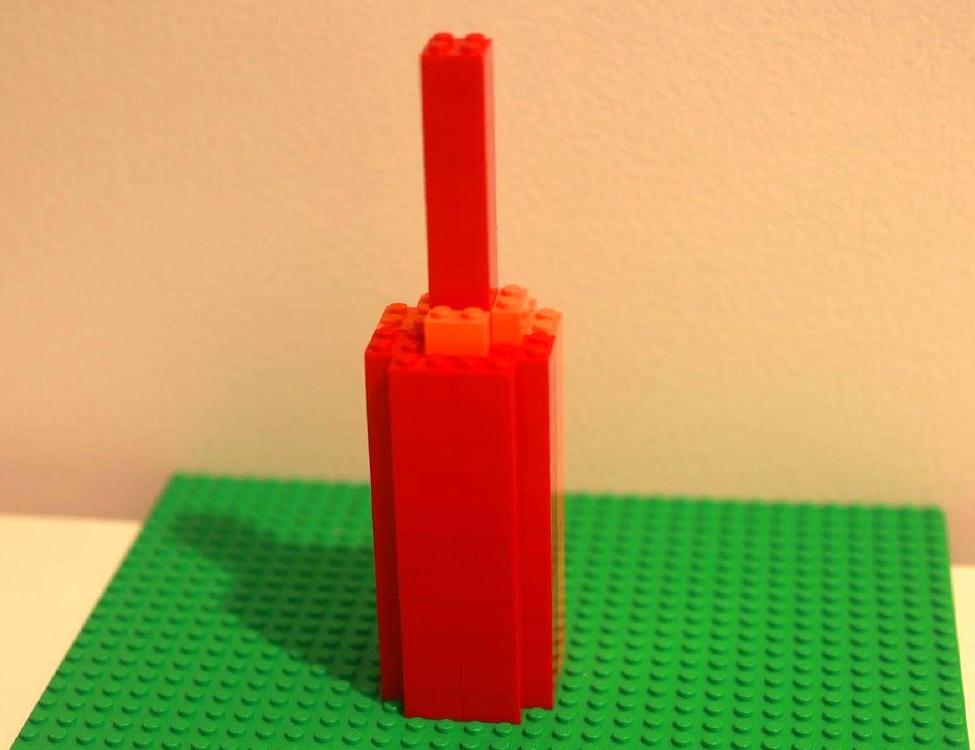}}\hfill
\subfigure[]{\includegraphics[width=0.095\linewidth]{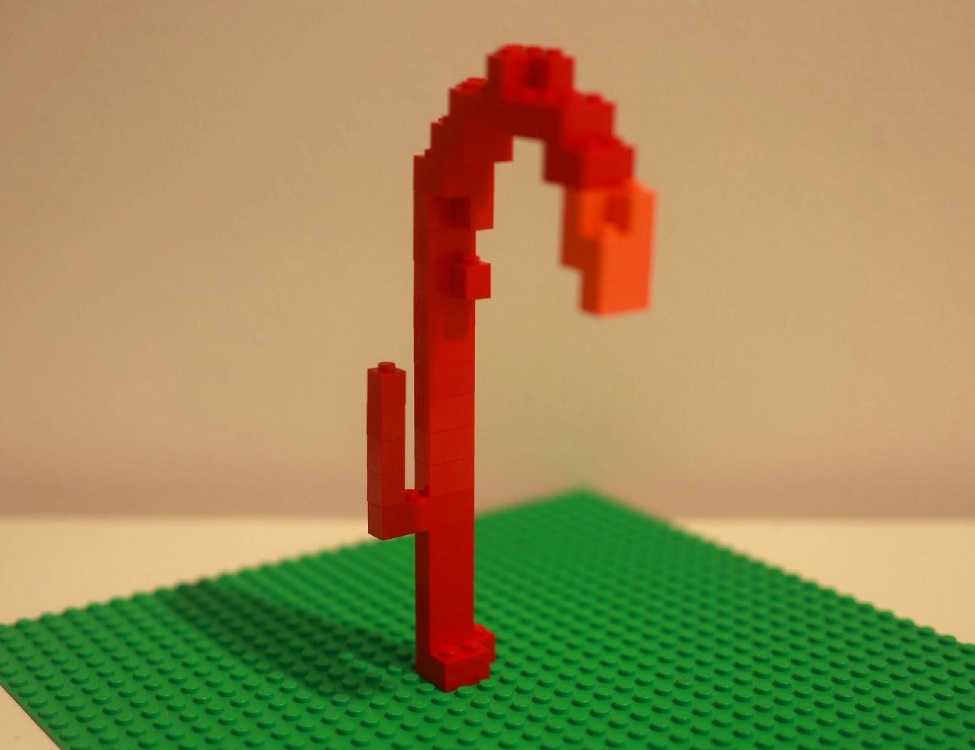}}\hfill
\subfigure[]{\includegraphics[width=0.095\linewidth]{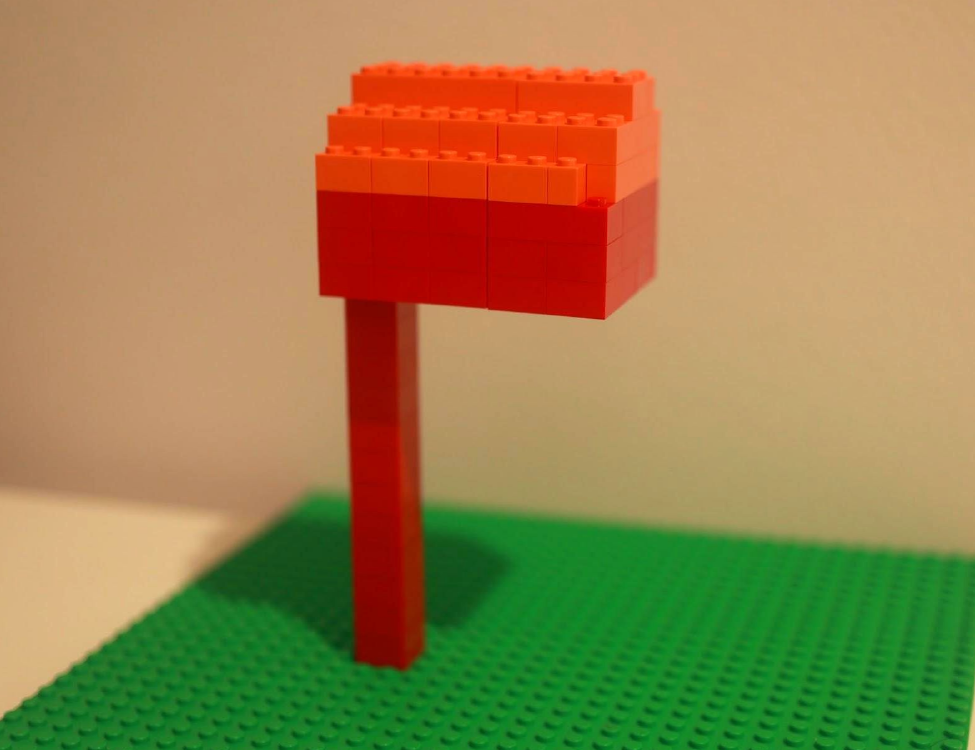}}\hfill
\subfigure[]{\includegraphics[width=0.095\linewidth]{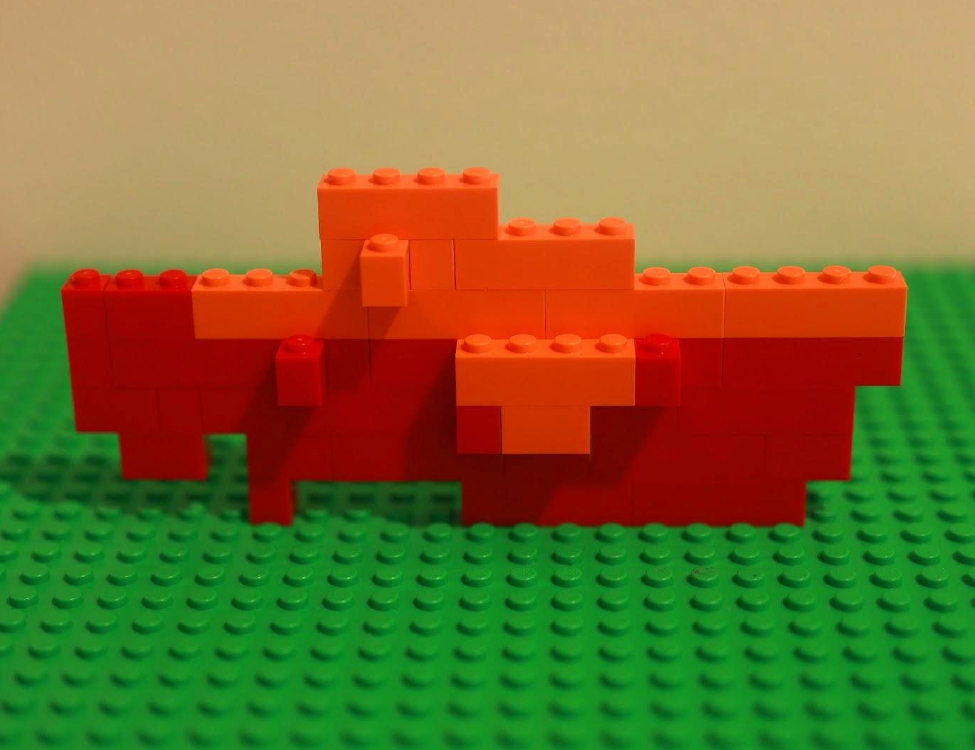}}\hfill
\subfigure[]{\includegraphics[width=0.095\linewidth]{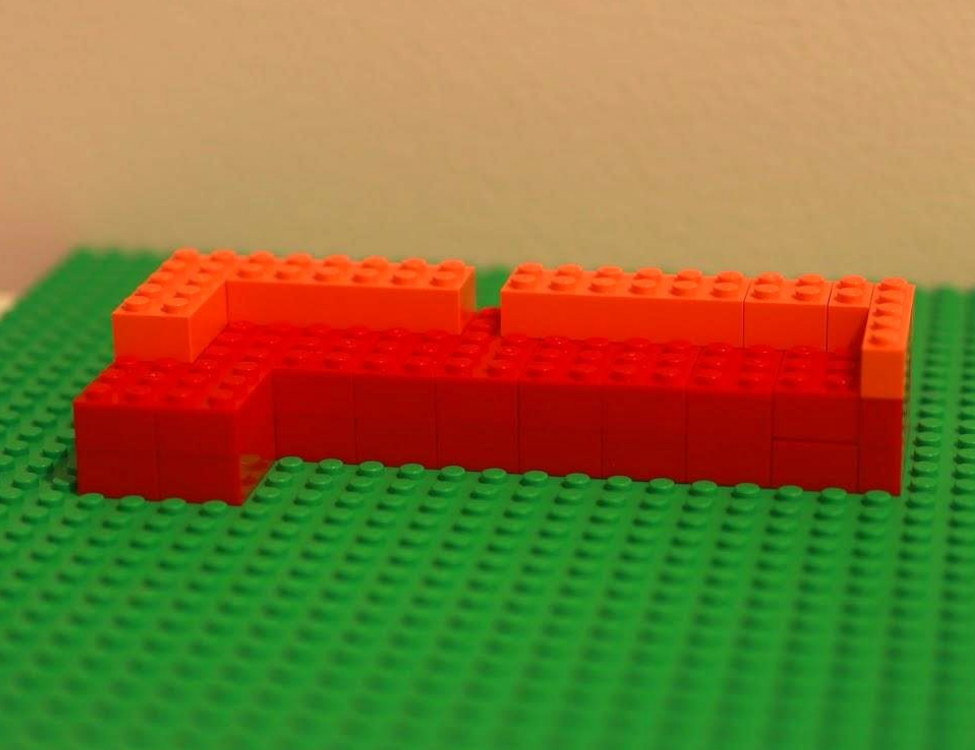}}\hfill
\subfigure[]{\includegraphics[width=0.095\linewidth]{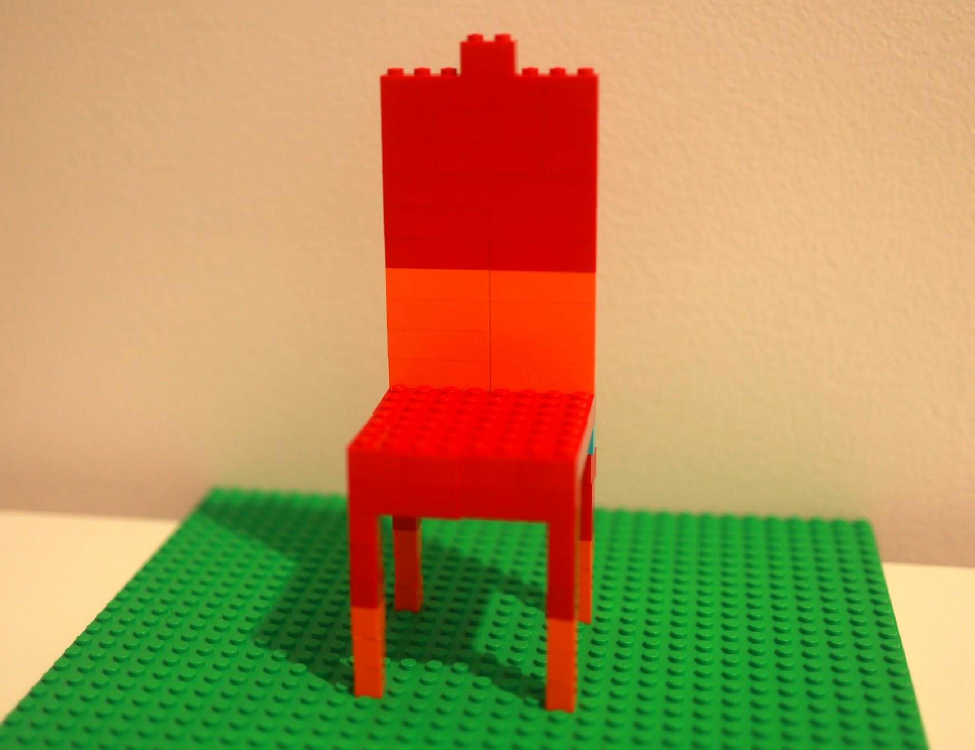}}\hfill
\subfigure[]{\includegraphics[width=0.095\linewidth]{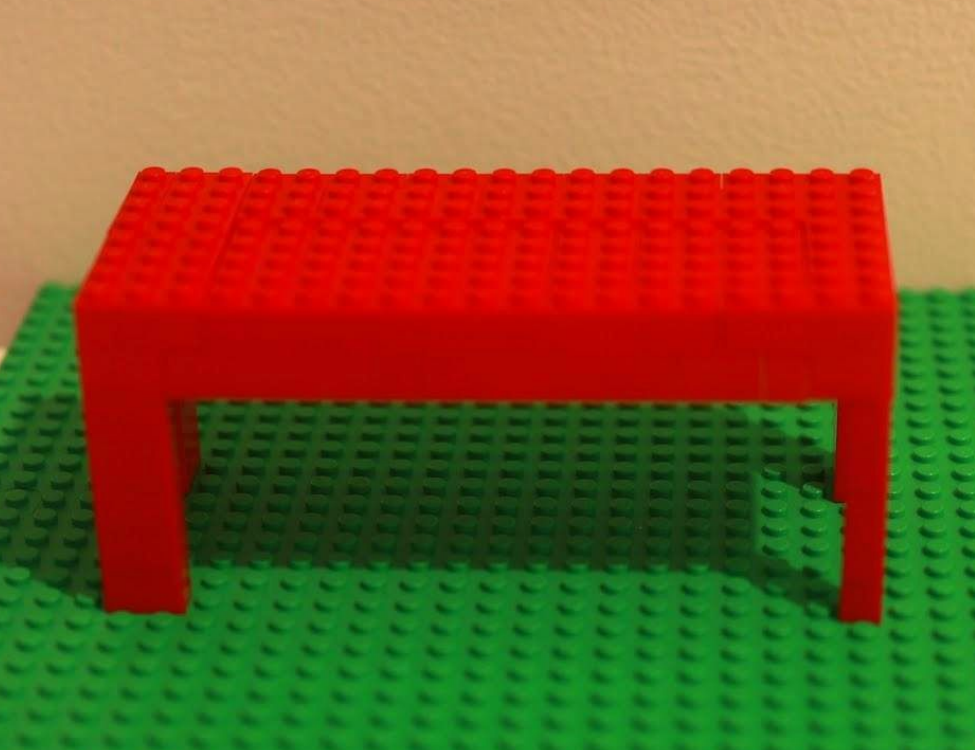}}\hfill
\subfigure[]{\includegraphics[width=0.095\linewidth]{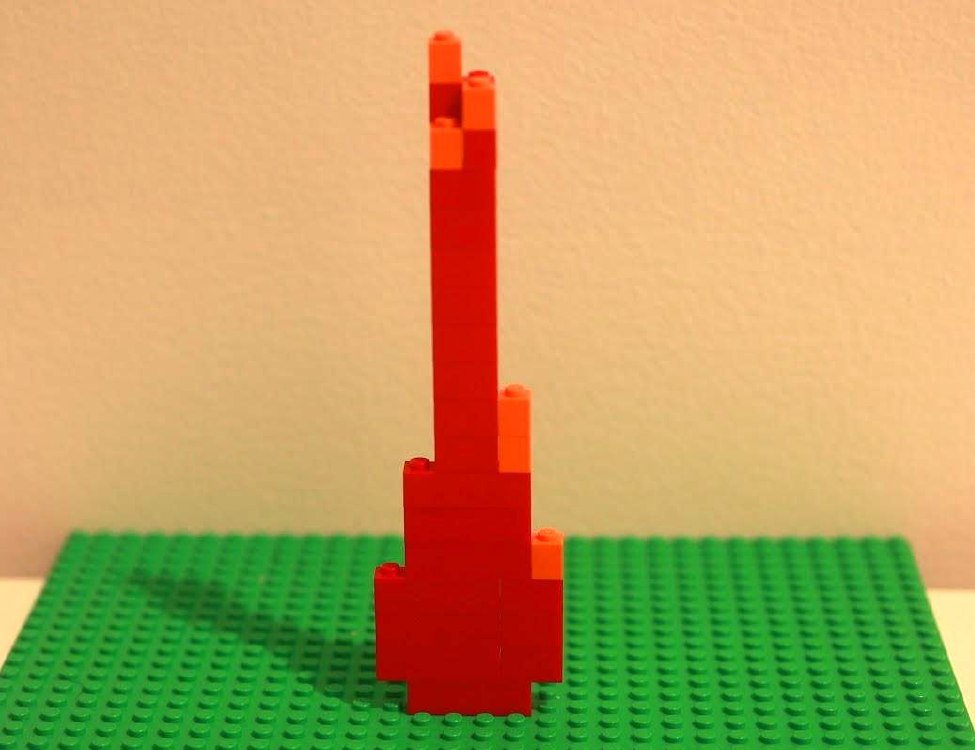}}
\vspace{-5pt}
    \caption{\footnotesize Example valid designs in the StableLEGO dataset. \label{fig:dataset}}
    \vspace{-15pt}
\end{figure*}

\begin{figure}
\centering
\subfigure[]{\includegraphics[width=0.19\linewidth]{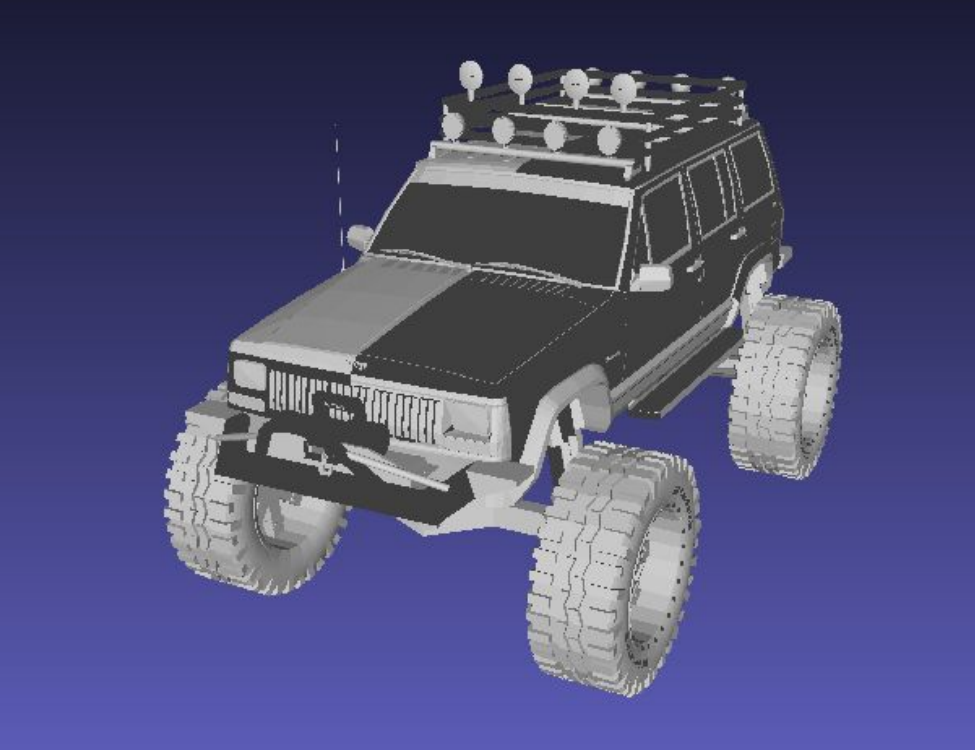}}\hfill
\subfigure[]{\includegraphics[width=0.19\linewidth]{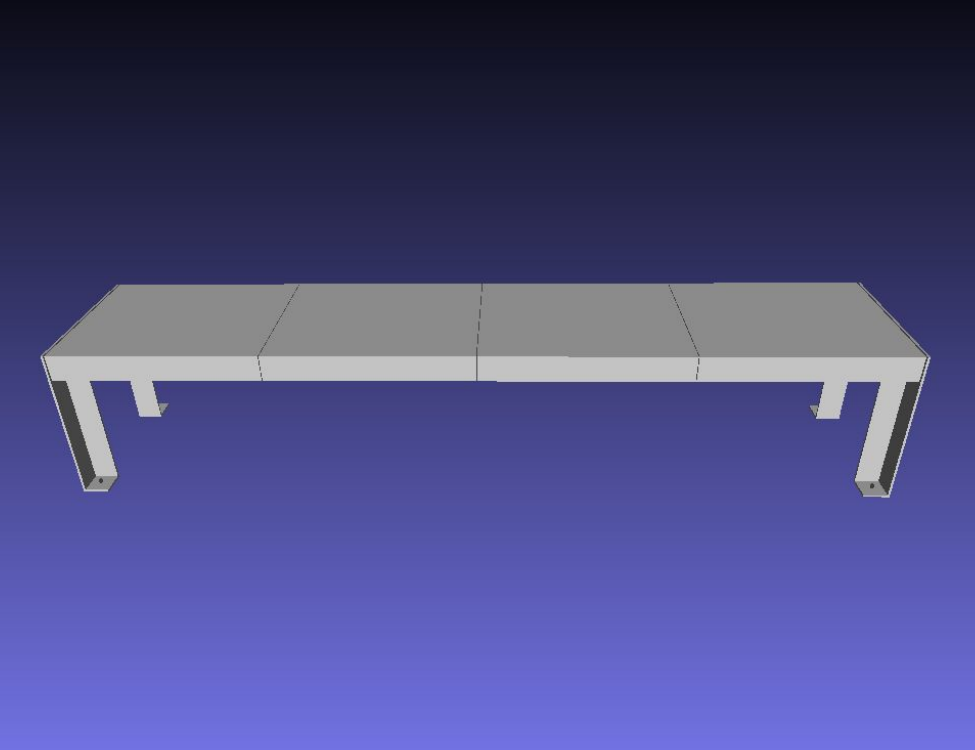}}\hfill
\subfigure[]{\includegraphics[width=0.19\linewidth]{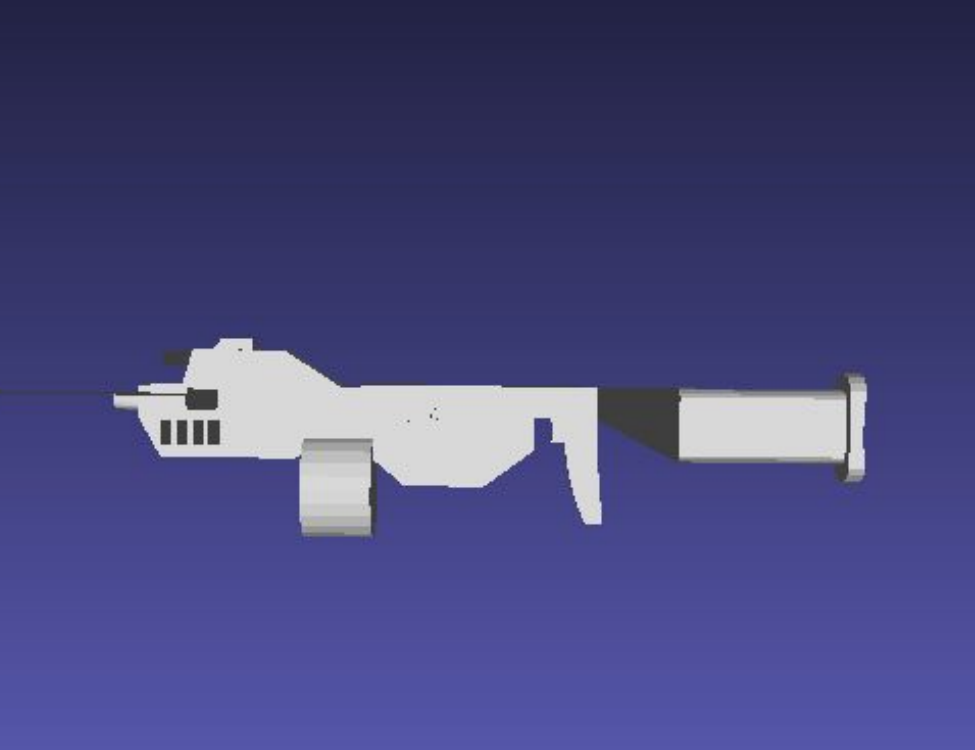}}\hfill
\subfigure[]{\includegraphics[width=0.19\linewidth]{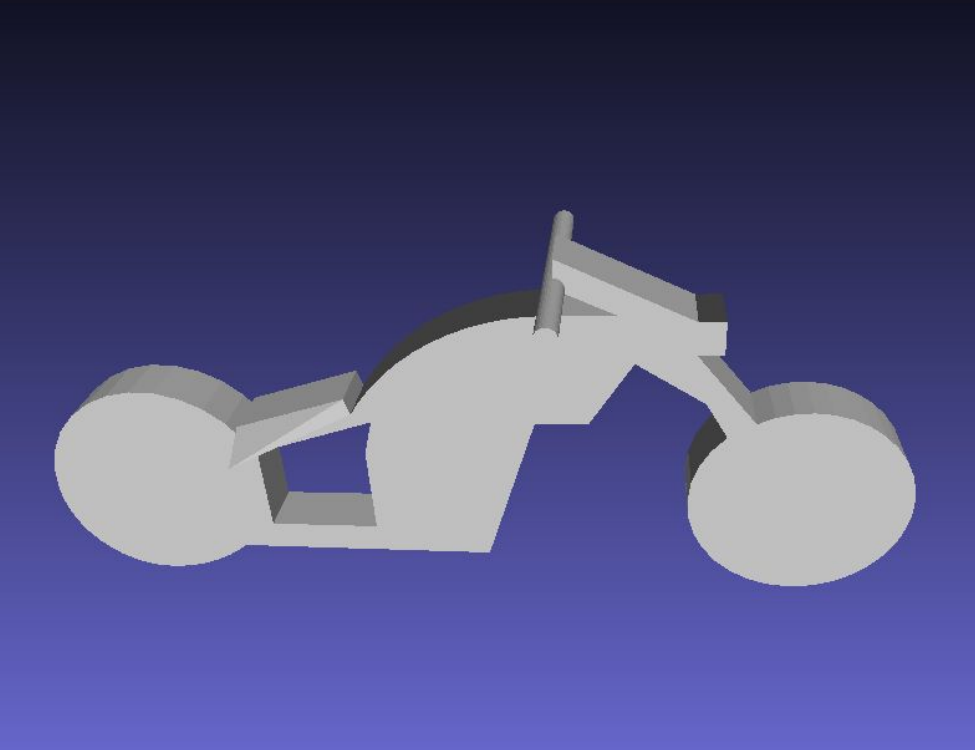}}\hfill
\subfigure[]{\includegraphics[width=0.19\linewidth]{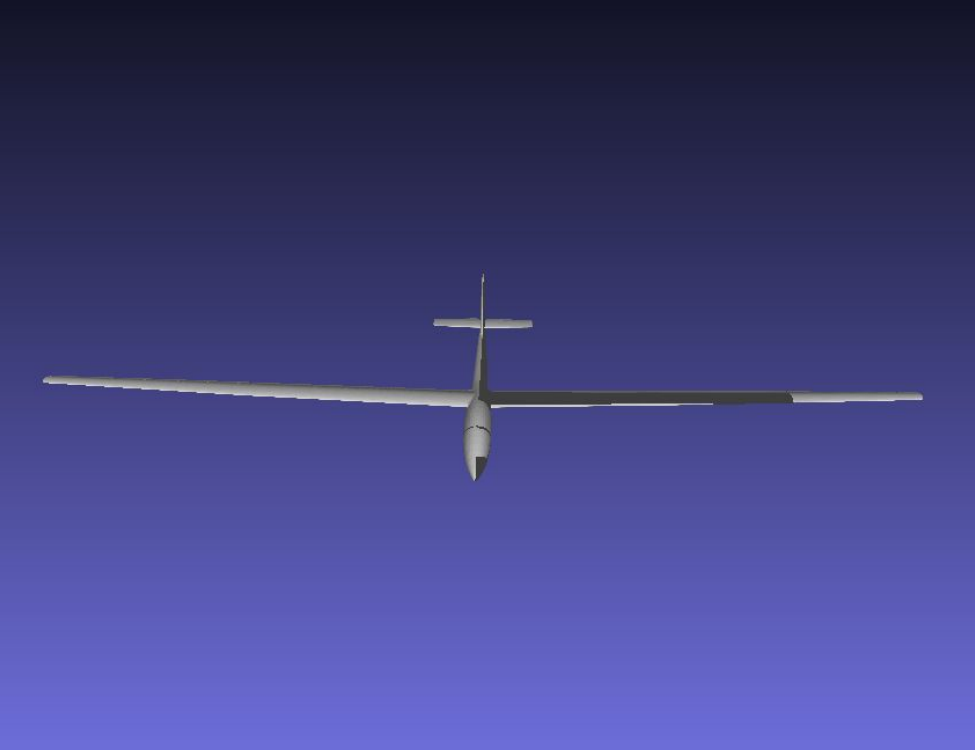}}
\\
\vspace{-5pt}
\subfigure[]{\includegraphics[width=0.19\linewidth]{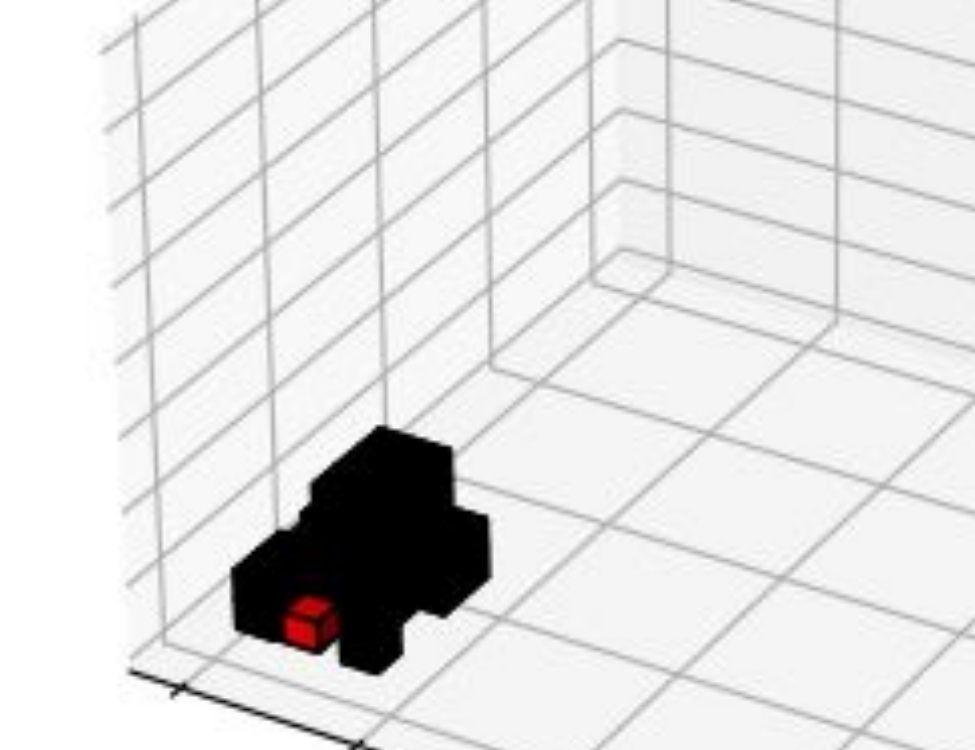}}\hfill
\subfigure[]{\includegraphics[width=0.19\linewidth]{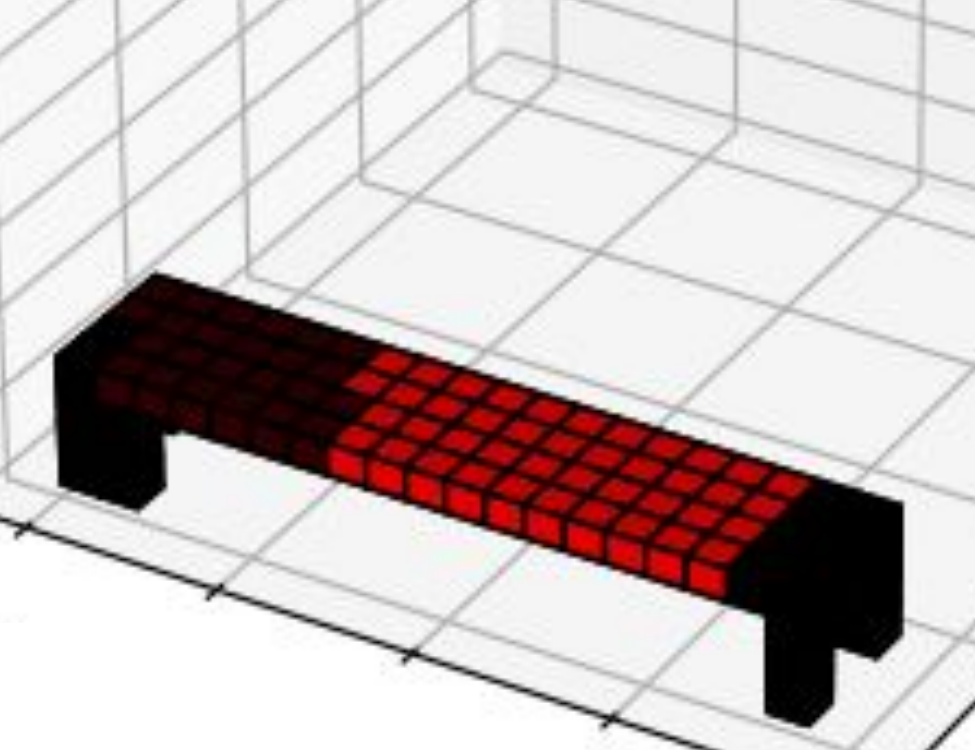}}\hfill
\subfigure[]{\includegraphics[width=0.19\linewidth]{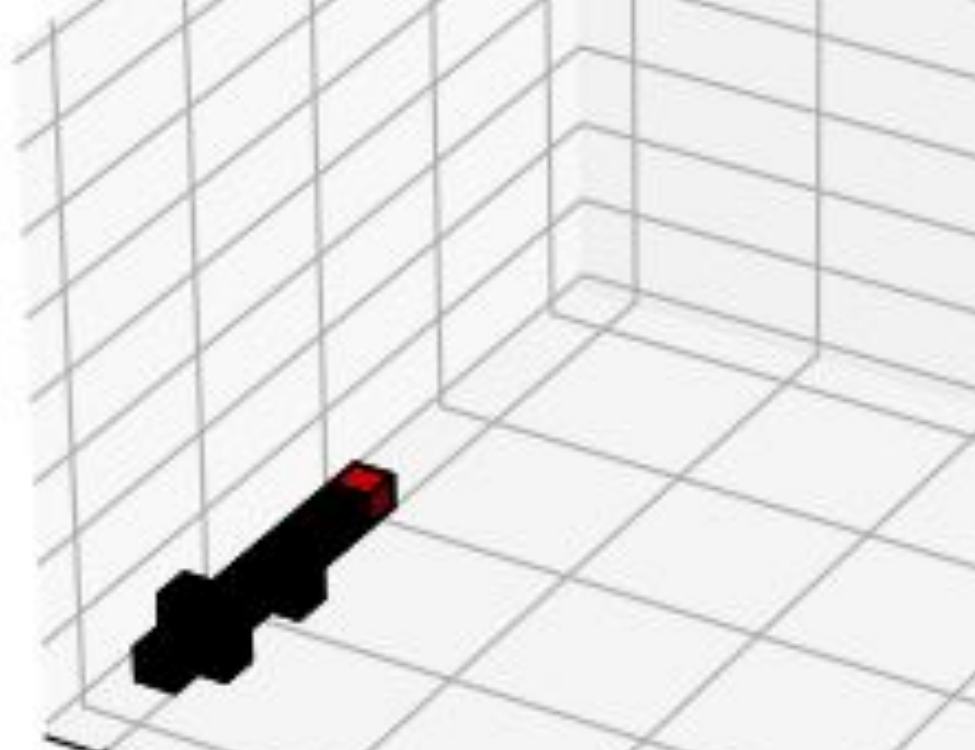}}\hfill
\subfigure[]{\includegraphics[width=0.19\linewidth]{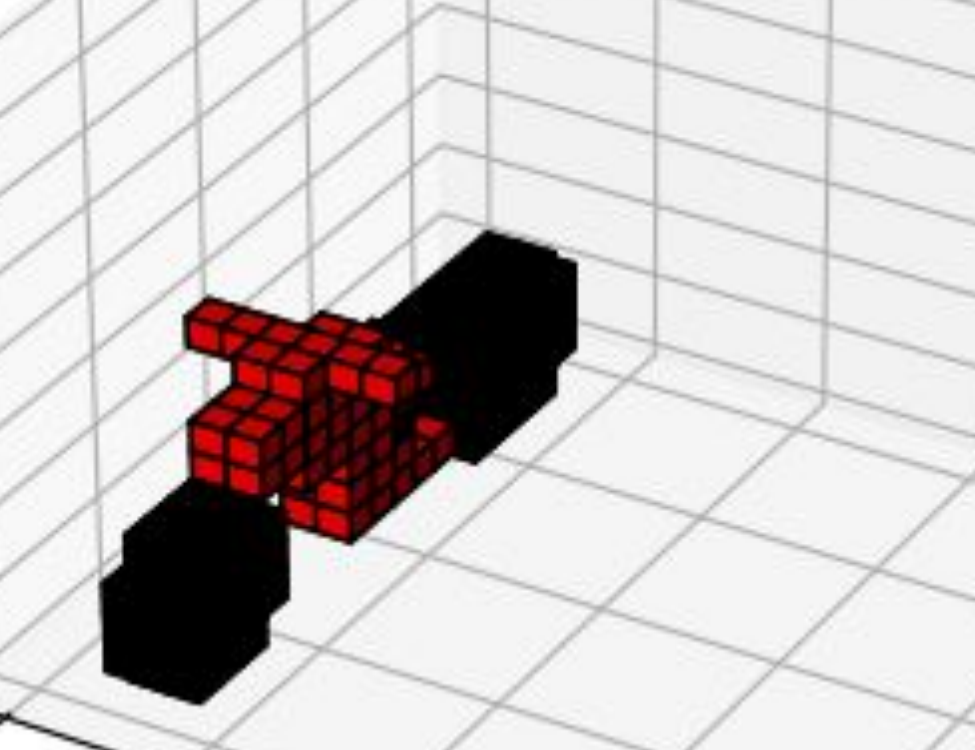}}\hfill
\subfigure[]{\includegraphics[width=0.19\linewidth]{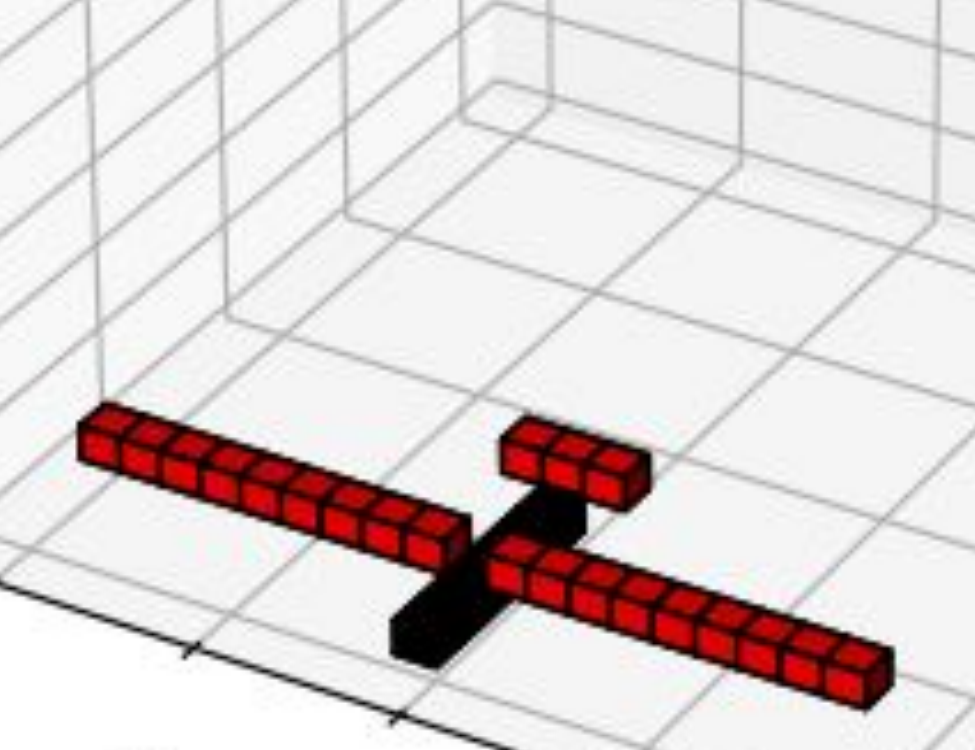}}
\vspace{-5pt}\\
\subfigure[]{\includegraphics[width=0.19\linewidth]{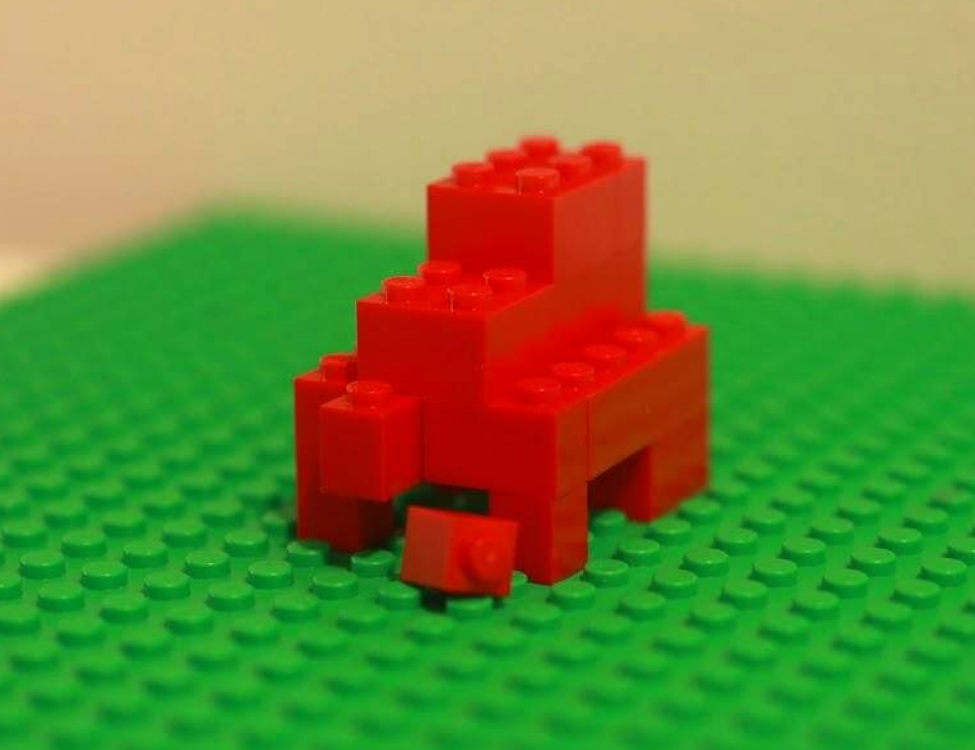}}\hfill
\subfigure[]{\includegraphics[width=0.19\linewidth]{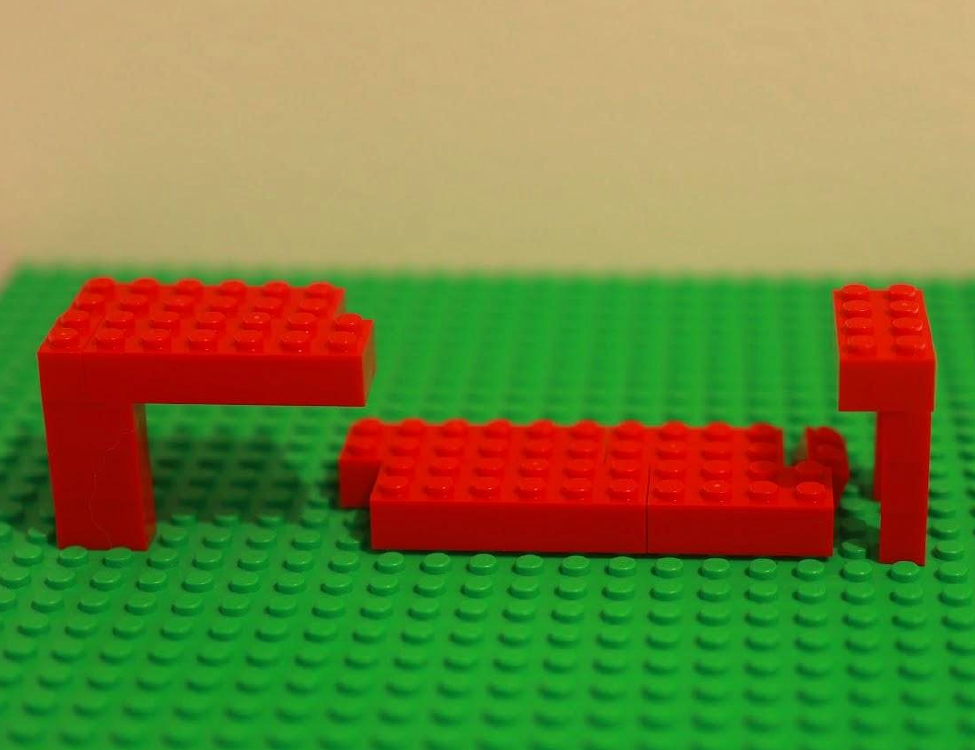}}\hfill
\subfigure[]{\includegraphics[width=0.19\linewidth]{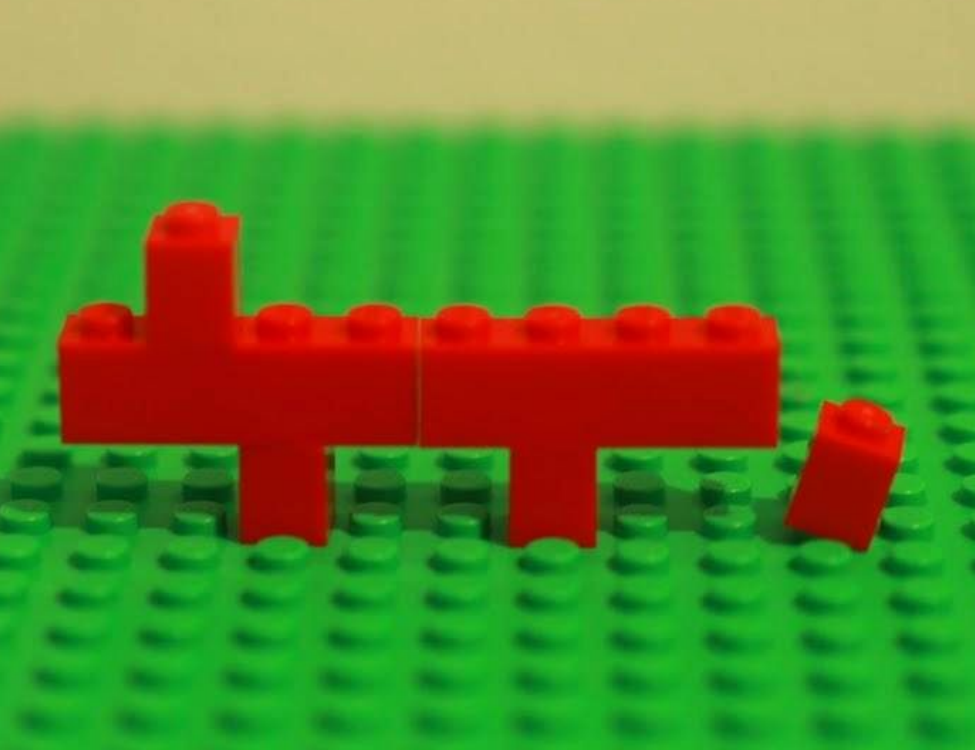}}\hfill
\subfigure[]{\includegraphics[width=0.19\linewidth]{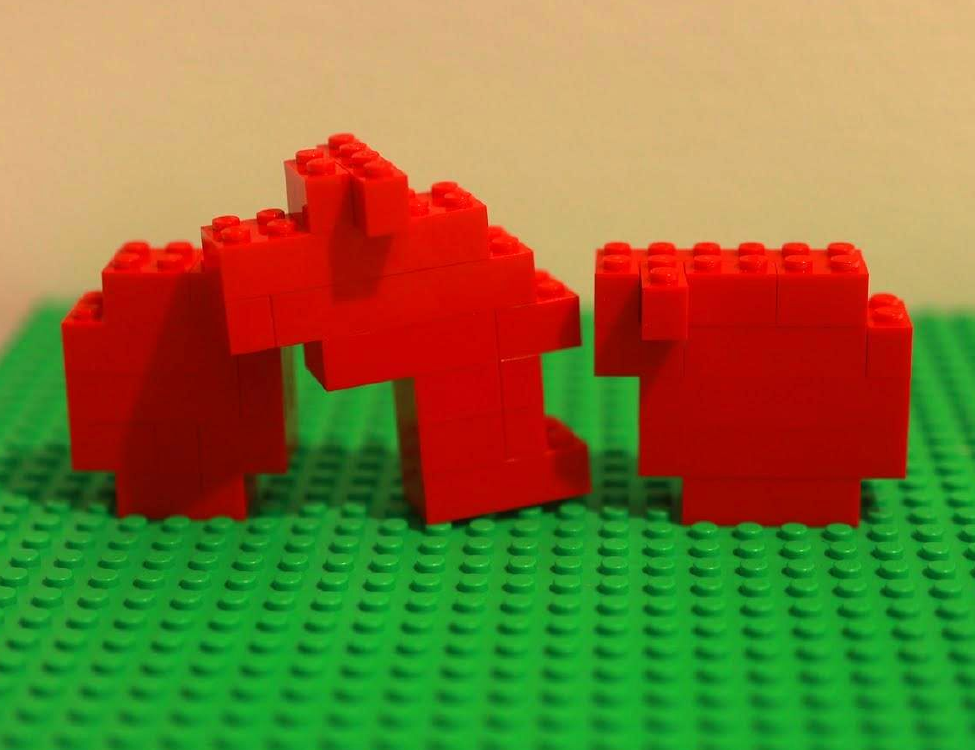}}\hfill
\subfigure[]{\includegraphics[width=0.19\linewidth]{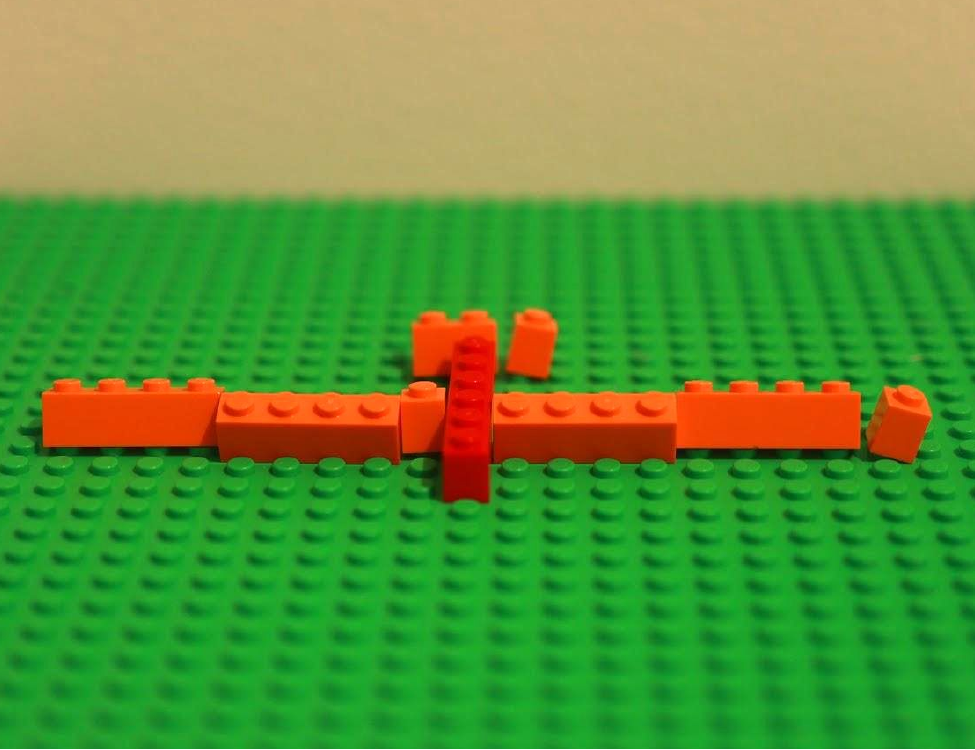}}
\vspace{-5pt}
    \caption{\footnotesize Example invalid designs in the StableLEGO dataset. \label{fig:invalid_dataset}}
    \vspace{-15pt}
    % \vspace{-15pt}
\end{figure}

\Cref{fig:dataset} illustrates the stability analysis results of our method on example valid designs in StableLego. 
The top row shows the original object, and the middle row depicts the stability analysis of the given Lego brick layout.
The third row shows the Lego structure built in real following the given brick layout.
We can see that our method correctly predicts that all structures are stable. 
\Cref{fig:invalid_dataset} shows examples of invalid Lego designs.
Our method correctly indicates that the structures will collapse (\ie the middle row).
In addition, the structures collapse at the predicted collapsing points as shown in the third row.

\begin{table}
\centering
\begin{tabular}{c  c  c c} 
\hline
 & Baseline \cite{10.1145/2816795.2818091} & Enhanced baseline & Ours  \\    
\hline
Solvable Count & 128 & 116 & \textbf{225}\\
Solvability &   $57.92\%$  & $51.56\%$ & $\mathbf{100\%}$\\
\hline
False Count & 12 &  \textbf{1}   & \textbf{1} \\
Stability Accuracy  & $90.63\%$ & $99.14\%$ &   $\mathbf{99.56\%}$\\
% Total Accuracy &  $51.56\%$  & $51.11\%$ & $\mathbf{99.56\%}$\\
\hline
\end{tabular}
\caption{\footnotesize Stability analysis results on a subset of StableLego dataset, \ie 225 Lego objects. \label{table:quatitative}}
\vspace{-15pt}
\end{table}

\begin{figure}
\centering
\subfigure[Lego structure.]{\includegraphics[width=0.3\linewidth]{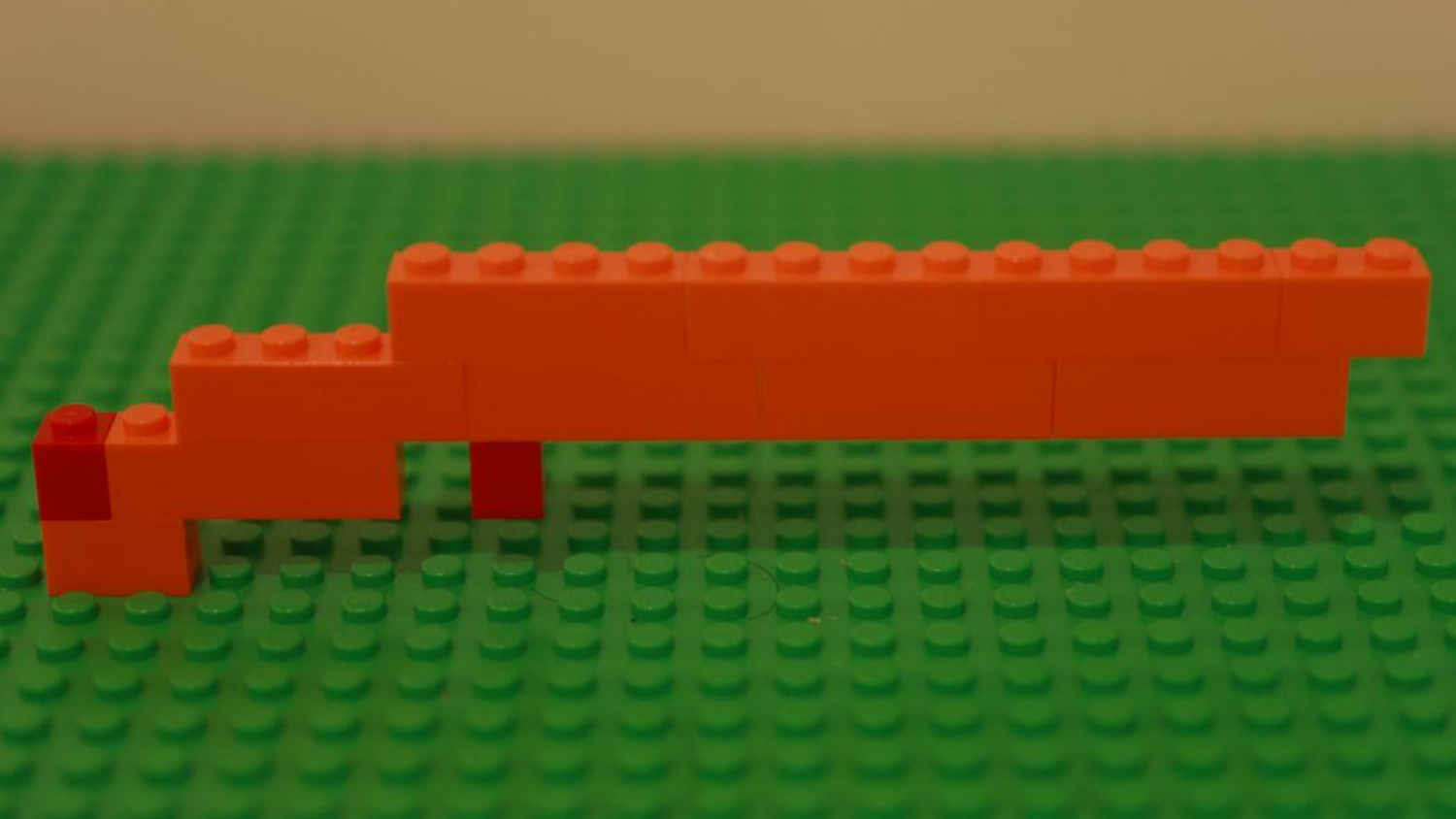}\label{fig:false_real}}\hfill
\subfigure[Our analysis.]{\includegraphics[width=0.3\linewidth]{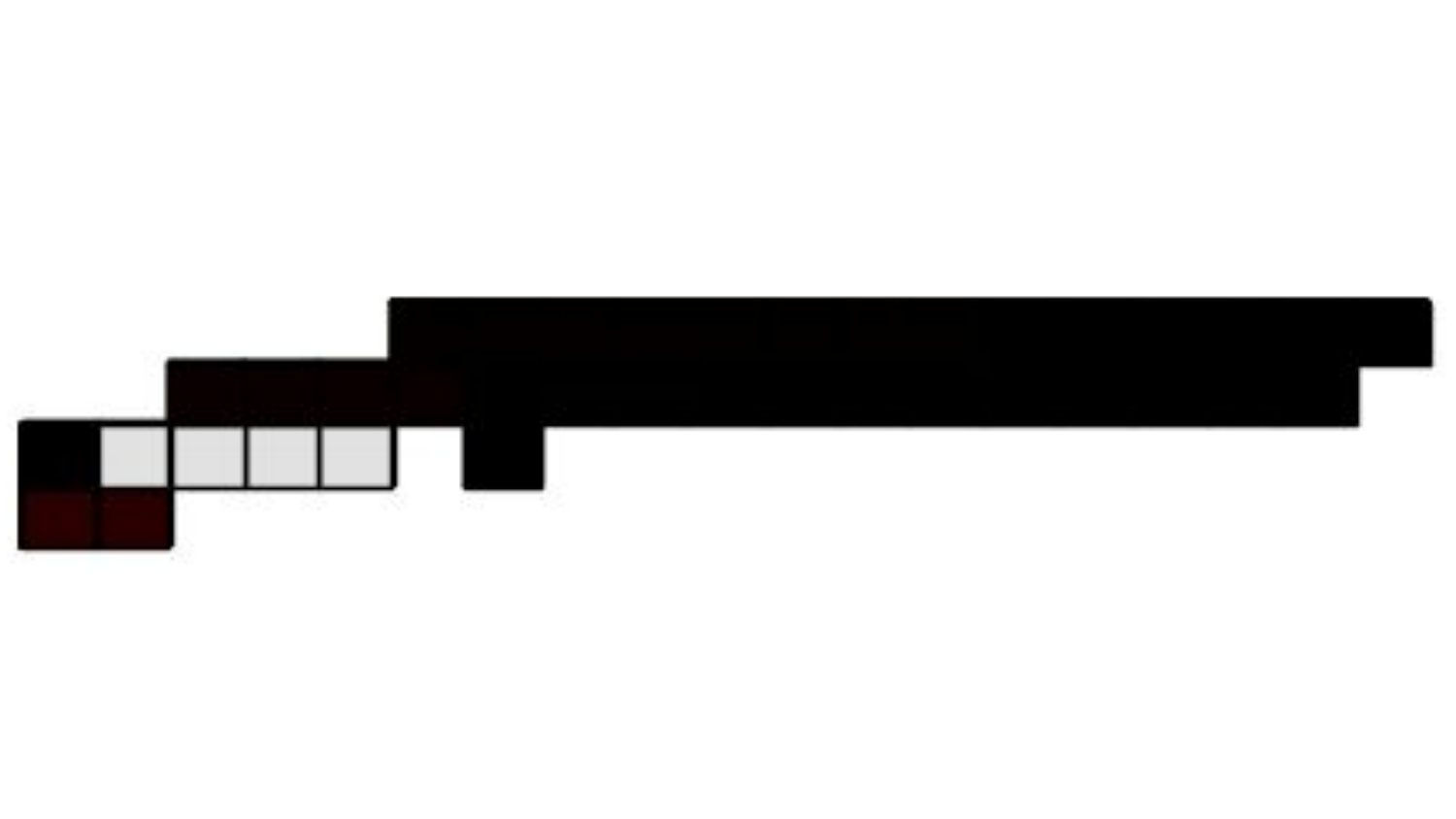}\label{fig:our_false}}\hfill
\subfigure[EB's analysis.]{\includegraphics[width=0.3\linewidth]{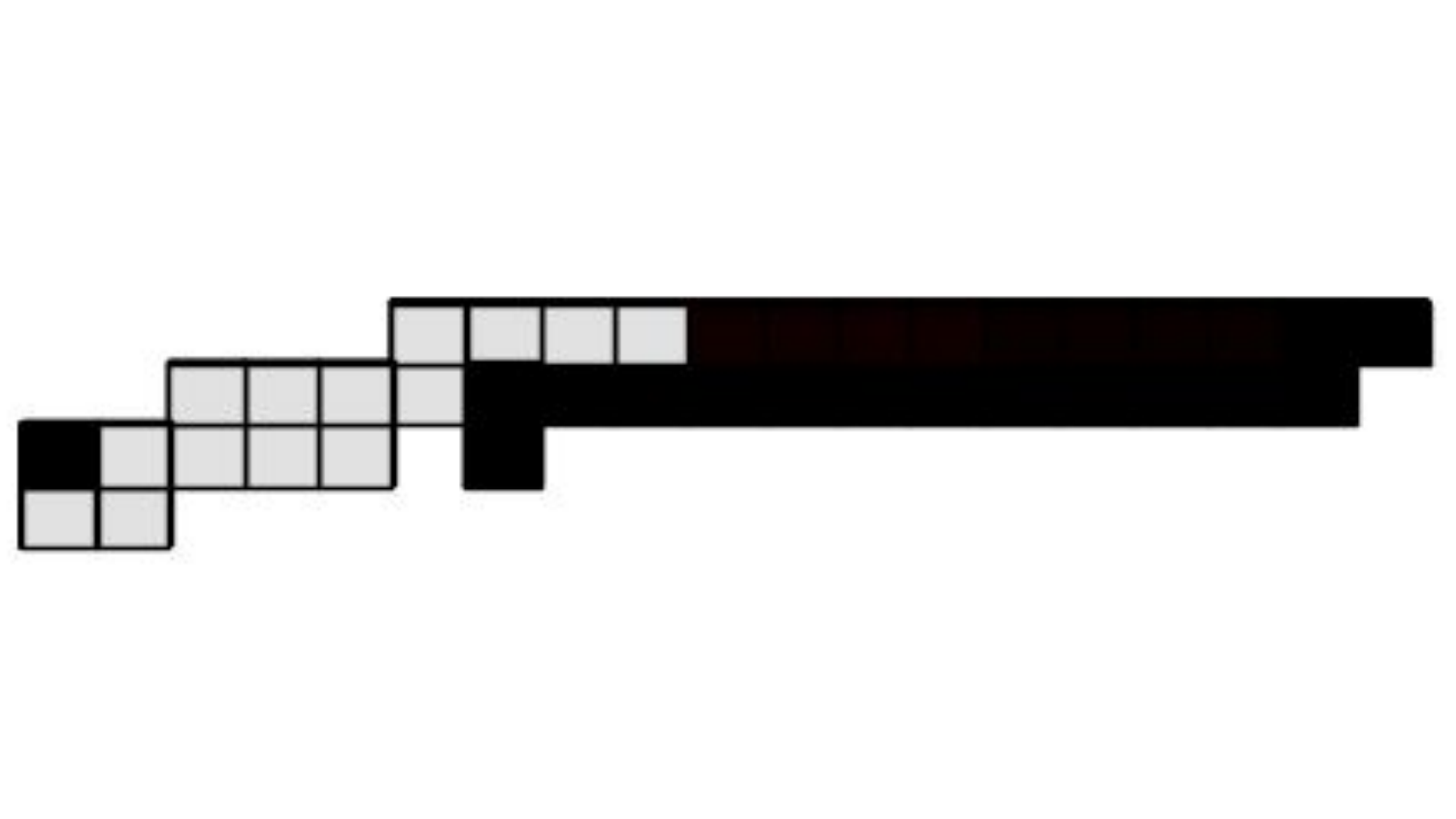}\label{fig:eb_false}}
\vspace{-5pt}
    \caption{\footnotesize The false prediction from ours and EB in \cref{table:quatitative}. \label{fig:false_example}}
    \vspace{-15pt}
\end{figure}

We use StableLego to conduct a thorough comparison between our proposed method and the baselines.
Due to the large scale of StableLego (\ie $50$k+ objects), it is practically impossible to physically build and verify each of them.
Thus, we randomly sampled a subset (\ie $225$ objects) and physically build each of them to verify the stability analysis result.
\Cref{table:quatitative} demonstrates the numerical comparison between our method and the baselines.
First, we compare the solvability of each method.
Given a structure, it is solvable if the algorithm can give a solution.
We can see our method achieves $100\%$ solvability, which means it successfully generates predictions for all of the 225 objects.
However, the baseline can only solve $57.92\%$ of the test set since it imposes the equilibrium conditions (\ie \cref{eq:force_eq,eq:torque_eq}) as constraints.
% If there exists no force distribution that satisfies the equilibrium constraints, the analysis will have no solution.
Similarly, the EB can solve $51.56\%$ of the test set, which is even lower due to the additional constraints (\ie \cref{eq:non-coexist,eq:equality}).
When applying the analysis methods to the entire StableLego, the baselines can only solve approximately $33\%$ (\ie $\sim$17k) of the entire dataset, while our proposed method can solve all of them.
% This is because the baseline enforces the static equilibrium as a hard constraint.
% When the structure violates the single-connected assumption, the baseline cannot estimate the stability of the given structure.
Second, we analyze the accuracy of each stability analysis algorithm.
The stability accuracy is defined as $\frac{N_{s}-N_{f}}{N_{s}}$.
$N_{f}$ is the number of incorrect predictions, \ie false count, and
$N_{s}$ is the number of solvable samples, \ie solvable count.
We can see that our method achieves the highest prediction accuracy.
It only has one false prediction out of 225 solvable structures. 
The false example is shown in \cref{fig:false_example}, in which our method predicts that the structure will collapse (\ie \cref{fig:our_false}) but turns out to be stable (\ie \cref{fig:false_real}).
This might be due to the model mismatch since the force model (\ie \cref{fig:force_model}) is a simplified approximation.
By imposing additional constraints, the EB also only has one false prediction, which is the identical one in our method (\ie \cref{fig:eb_false}).
However, by comparing \cref{fig:our_false,fig:eb_false}, our method gives a more realistic force distribution.
On the other hand, the baseline has significantly more incorrect predictions, and thus, has the lowest stability accuracy.
Due to the lack of constraints, it incorrectly predicts unstable structures to be stable.
% Note that the baseline has higher total accuracy than the EB.
% This is because we count unsolvable samples as incorrect predictions when calculating the metric.

\begin{figure}
    \centering
    \input{plot/time_plot}
    \vspace{-15pt}
    \caption{\footnotesize Computation time for the stability analysis. Blue: 50 percentile. Yellow: 75 percentile. Red: 25 percentile. Dashed: overall time. Solid: time for solving \cref{eq:optimization}. \label{fig:time}}
    \vspace{-15pt}
\end{figure}

\subsection{Stability Analysis Computation Time}
% Existing methods, \eg FEM \cite{pletzbrickfem}, can estimate structural stability, but the computation time is expensive.
It is desired that the stability can be efficiently estimated.
For the structures in \cref{fig:block_examples}, on average, the analysis results shown in \cref{fig:accuracy} are solved within $0.1\text{s}$ overall by our method (\ie constructing the optimization problem \cref{eq:optimization} and then solving it), and $0.01\text{s}$ if we only count the time for solving \cref{eq:optimization}.
To quantitatively evaluate the computation efficiency, we test our stability analysis in a controlled setting.
In particular, we use unit Lego bricks (\ie $1\times 1$) to build cuboids with different dimensions up to $10\times 10\times 10$.
The top figure in \cref{fig:time} shows the computation time for solving the structural stability with different numbers of bricks.
We can see that our method is efficient since it can estimate the stability within 1s, and mostly even within 0.5s.
The overall computation time (\ie the dashed lines) is less than 1.5s, and mostly within 1s.
As the size of the structure grows, it contains more bricks and takes longer to estimate the stability.
To further evaluate the computation efficiency, we show the computation time of solving the stability of StableLego in the bottom figure of \cref{fig:time}.
In general, it takes longer time for the structures in StableLego since it contains a wider variety of Lego bricks and the structures are more complex.
However, our method is still able to solve efficiently.
When having less than 300 bricks, our method can solve within 1s. 
As the number of bricks grows, the complexity increases, and it takes a longer time. 
But we can still expect it to estimate the stability within several seconds.

\subsection{Extension}
Our experiment only uses the highlighted bricks in \cref{table:brick_weight}, but the proposed stability analysis formulation applies to other bricks as well.
Moreover, the proposed method is applicable to other real-world applications.

\begin{figure}
\centering
\subfigure[]{\includegraphics[width=0.165\linewidth]{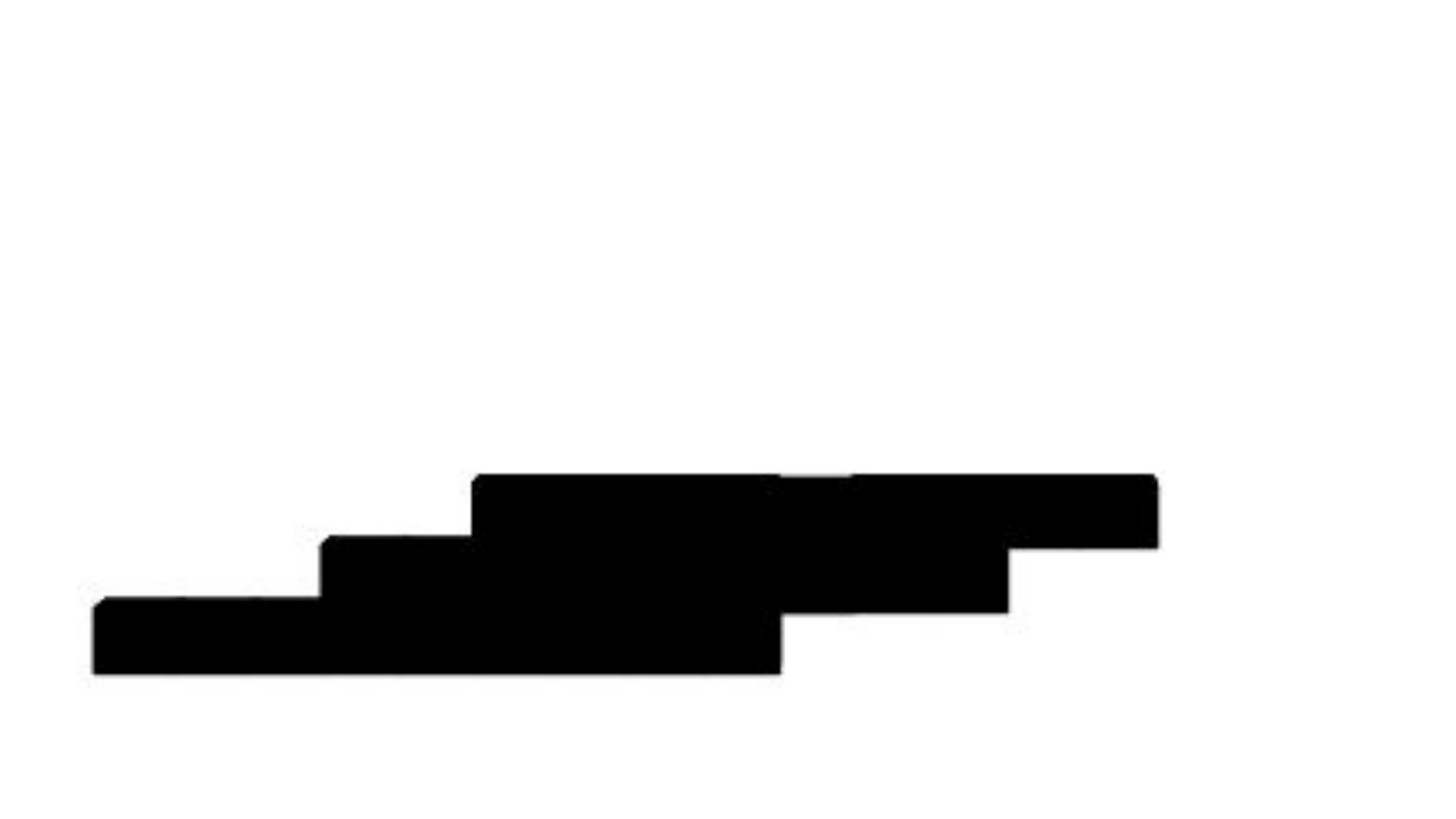}\label{fig:stack1_analysis}}\hfill
\subfigure[]{\includegraphics[width=0.165\linewidth]{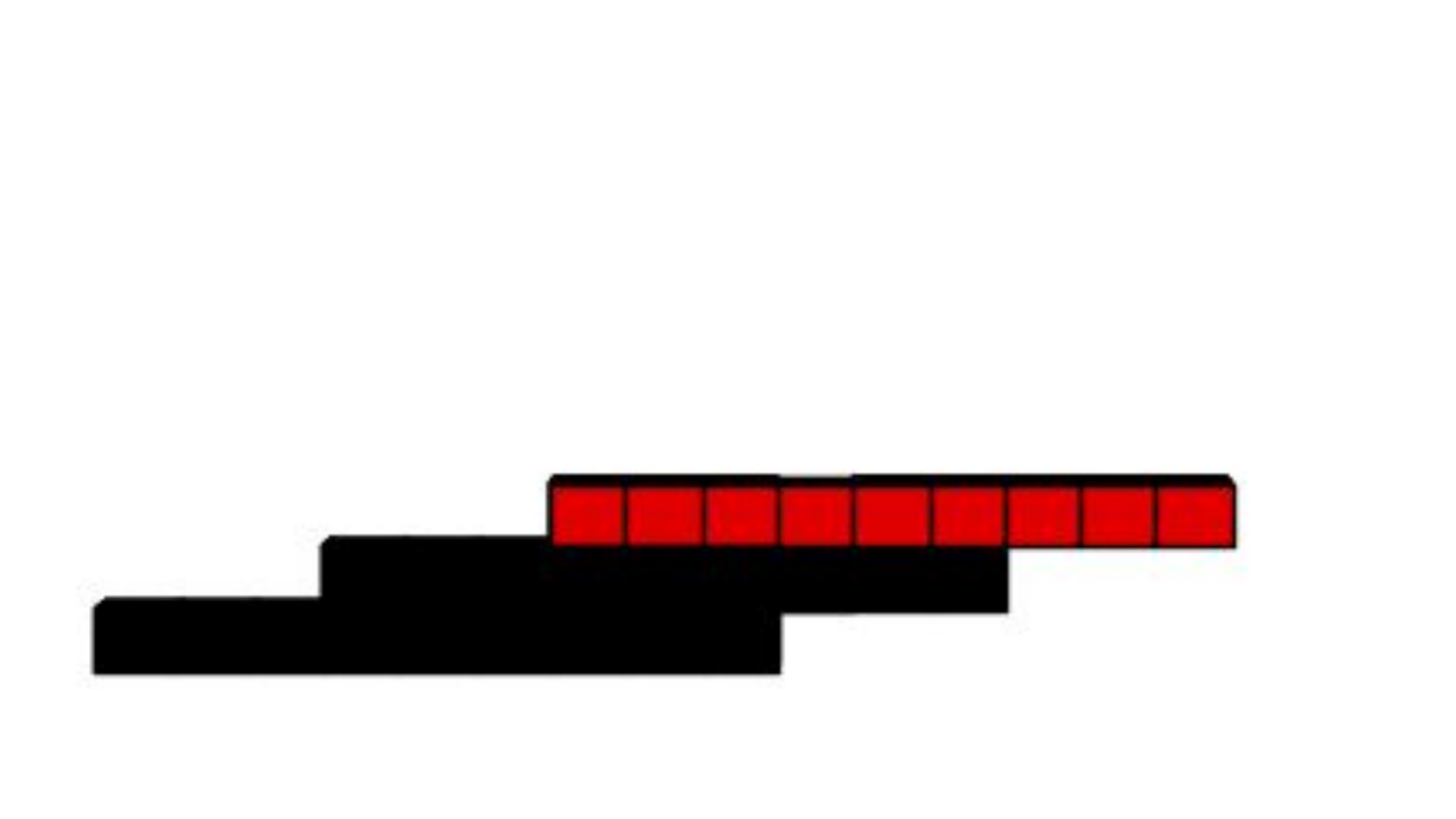}\label{fig:stack2_analysis}}\hfill
\subfigure[]{\includegraphics[width=0.165\linewidth]{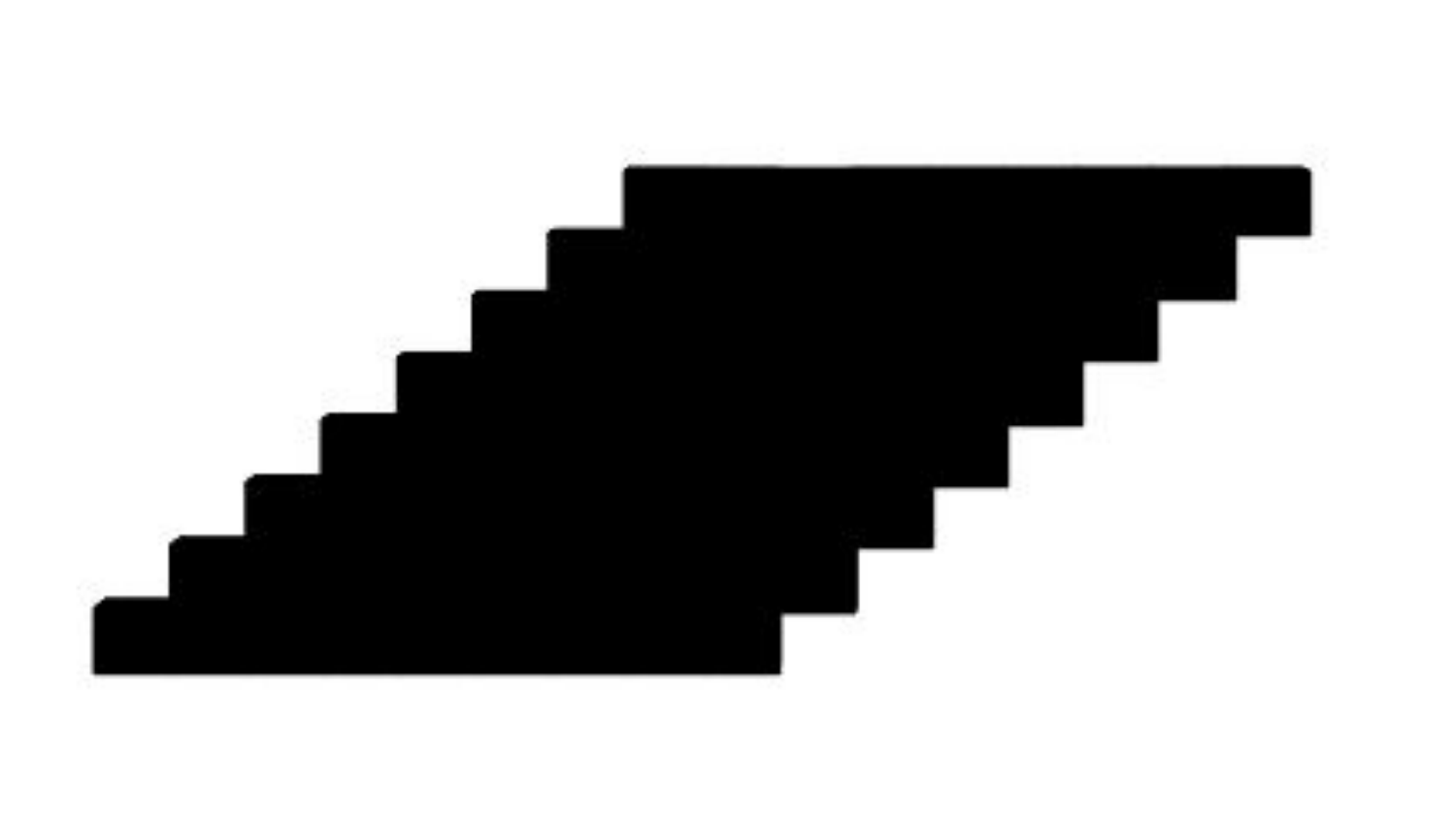}\label{fig:stack3_analysis}}\hfill
\subfigure[]{\includegraphics[width=0.165\linewidth]{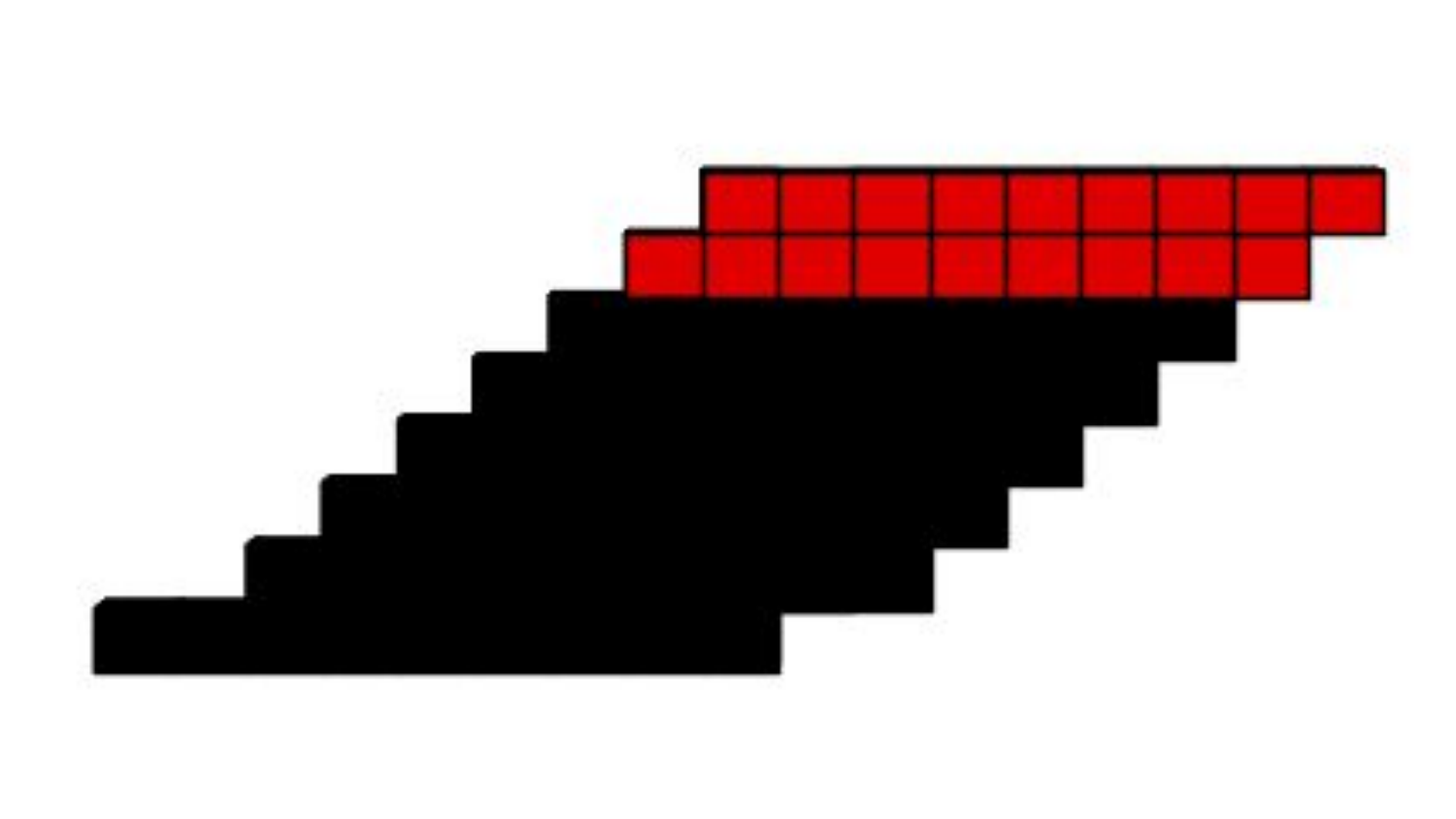}\label{fig:stack4_analysis}}\hfill
\subfigure[]{\includegraphics[width=0.165\linewidth]{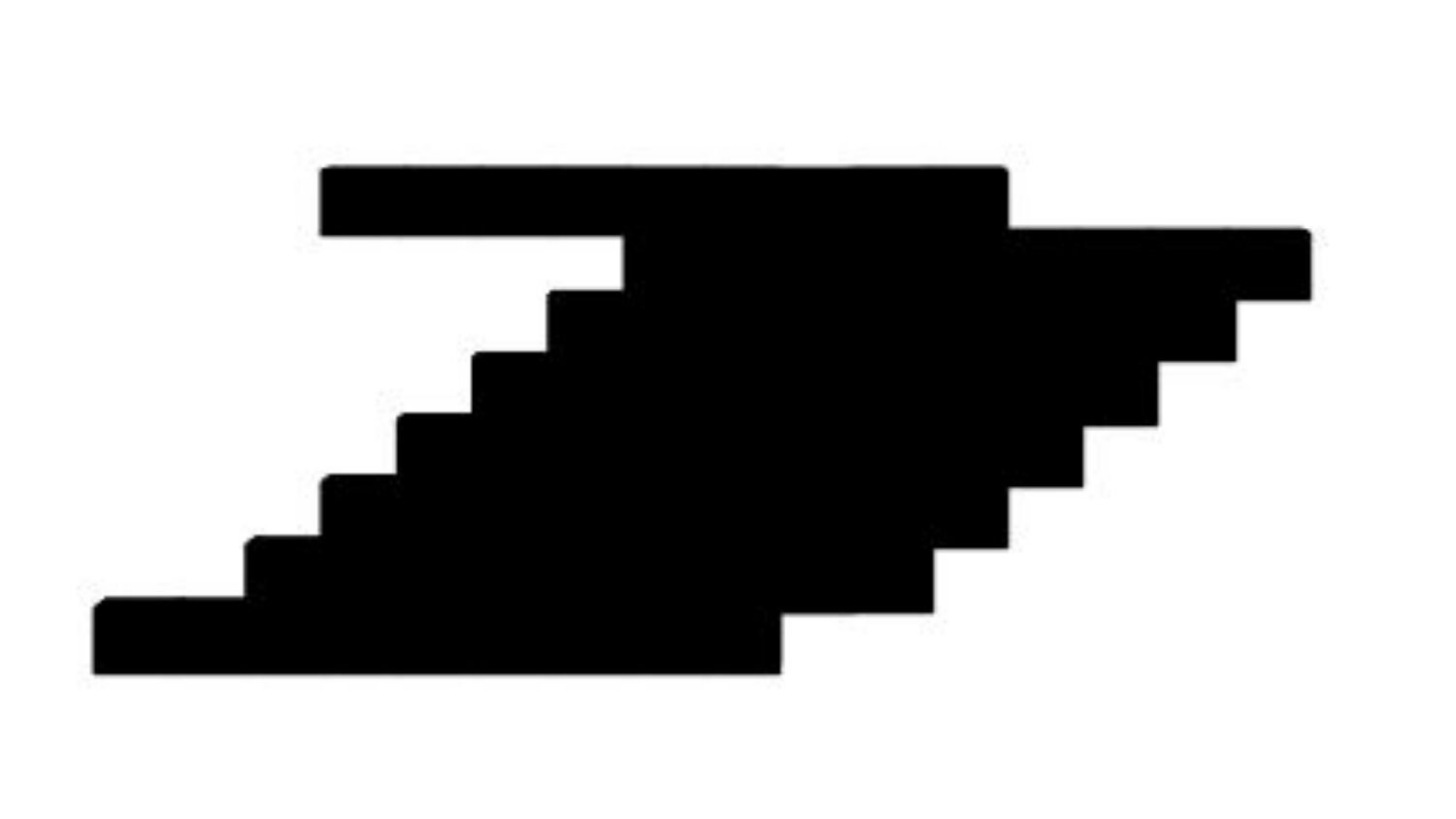}\label{fig:stack5_analysis}}\hfill
\subfigure[]{\includegraphics[width=0.165\linewidth]{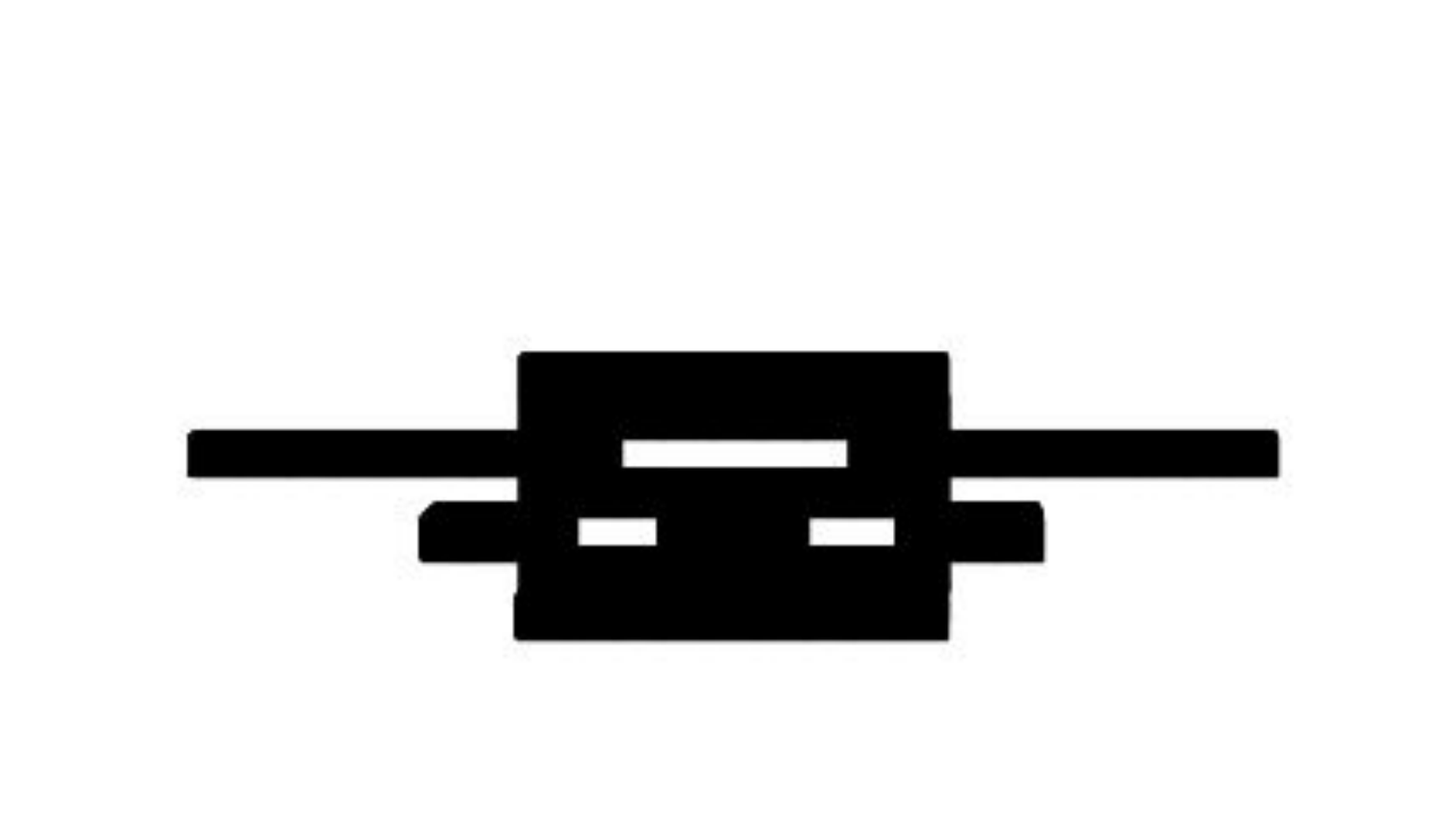}\label{fig:stack6_analysis}}
\vspace{-5pt}\\
\subfigure[]{\includegraphics[width=0.165\linewidth]{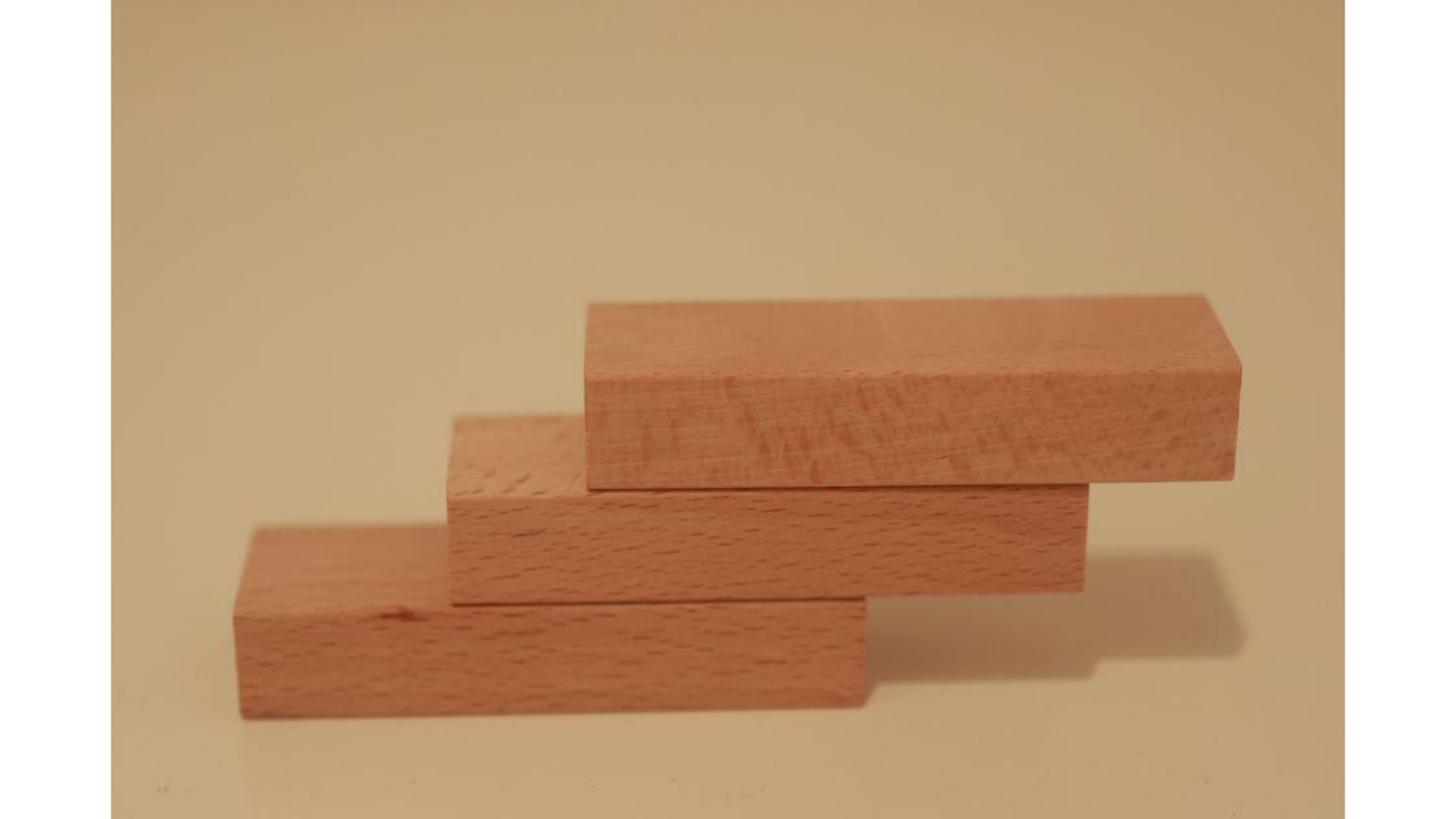}\label{fig:stack1}}\hfill
\subfigure[]{\includegraphics[width=0.165\linewidth]{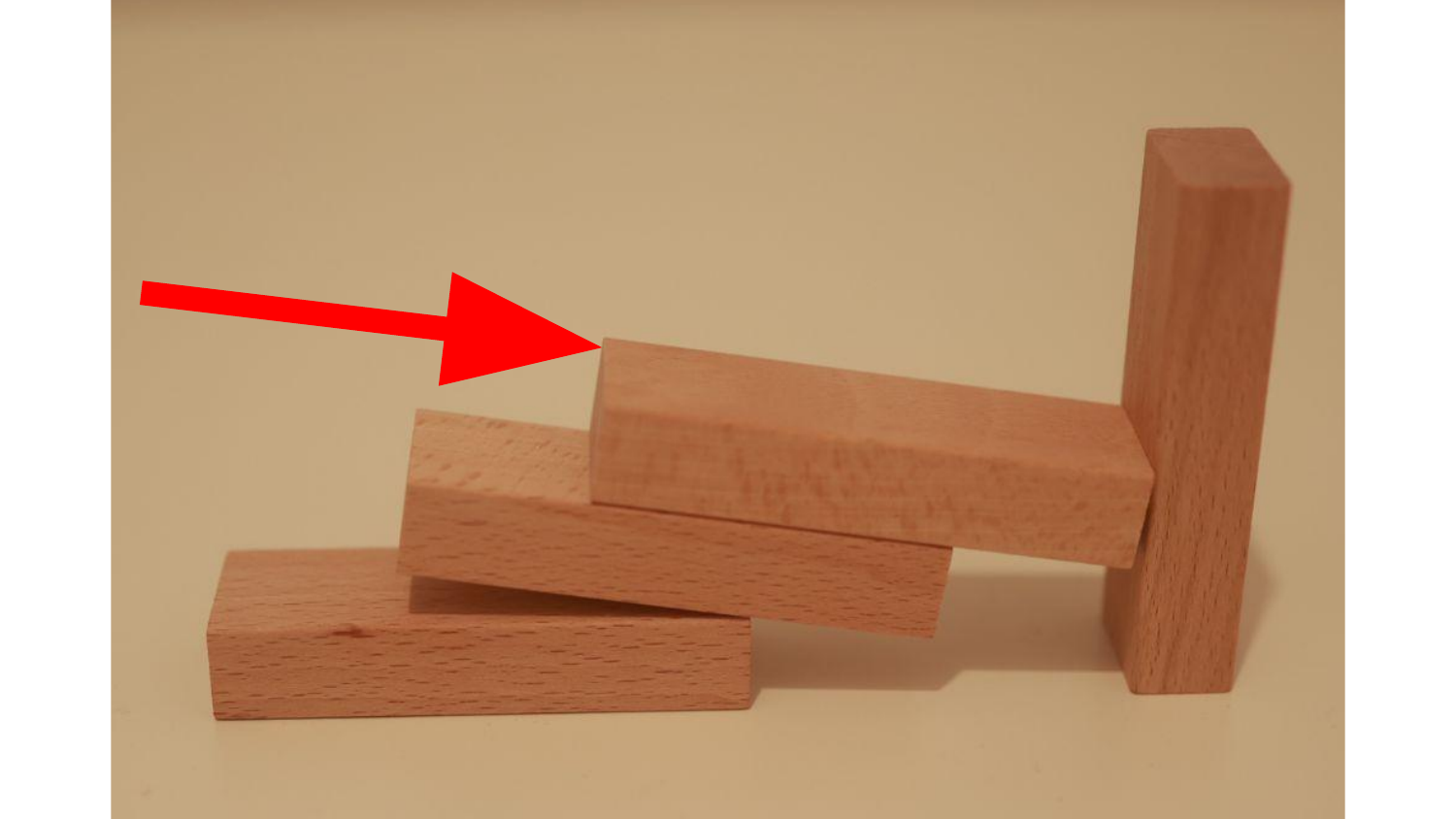}\label{fig:stack2}}\hfill
\subfigure[]{\includegraphics[width=0.165\linewidth]{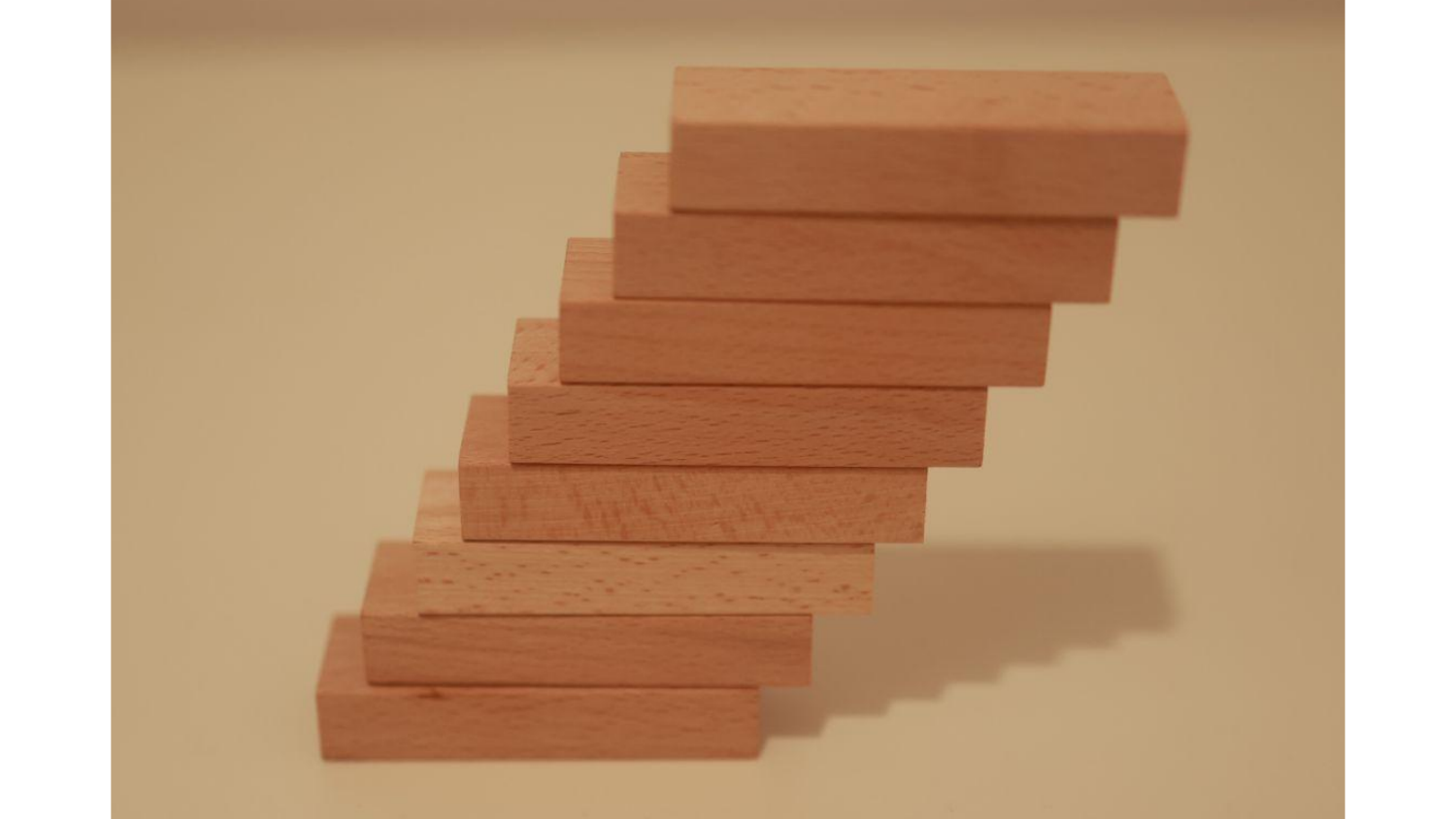}\label{fig:stack3}}\hfill
\subfigure[]{\includegraphics[width=0.165\linewidth]{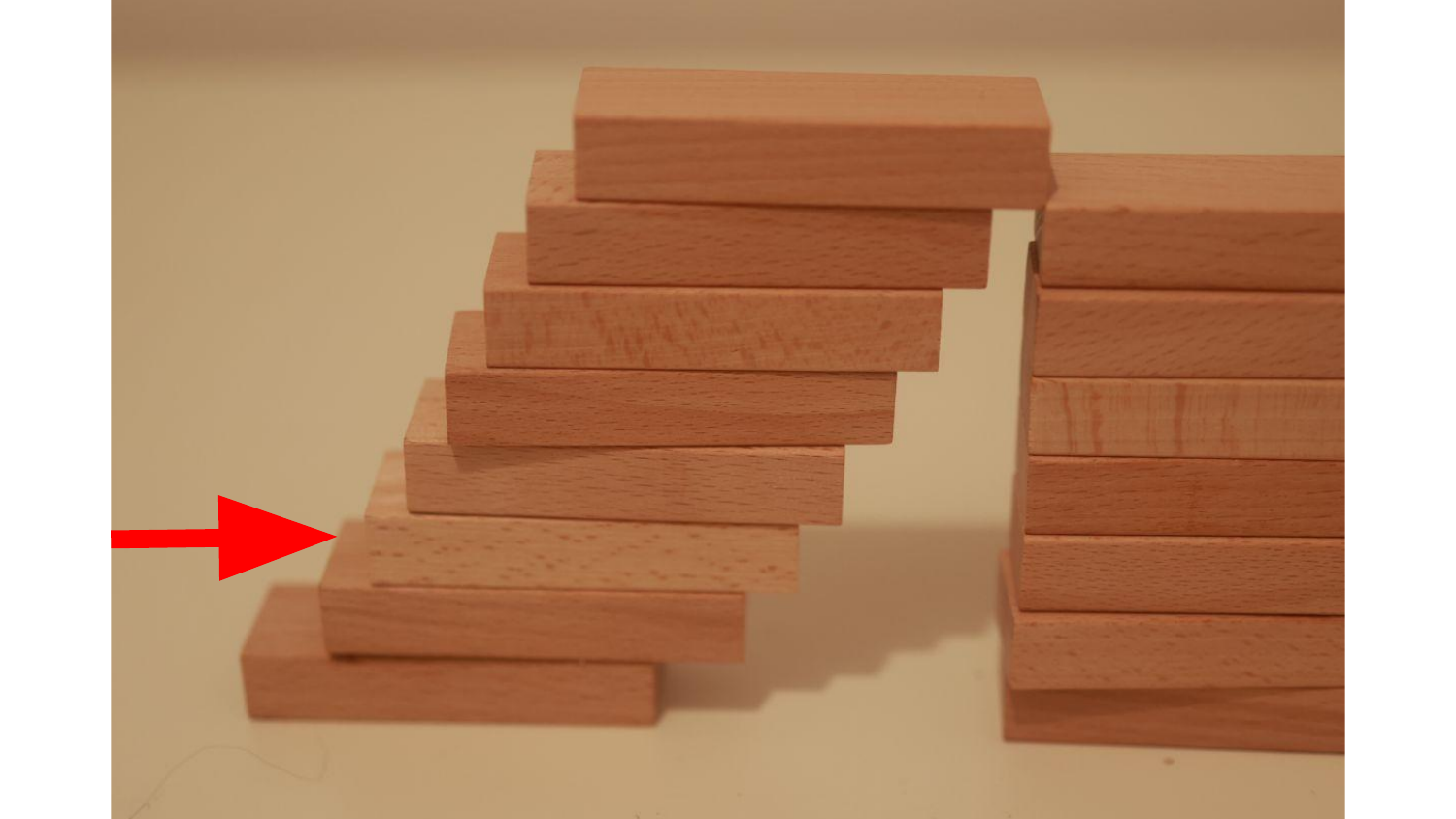}\label{fig:stack4}}\hfill
\subfigure[]{\includegraphics[width=0.165\linewidth]{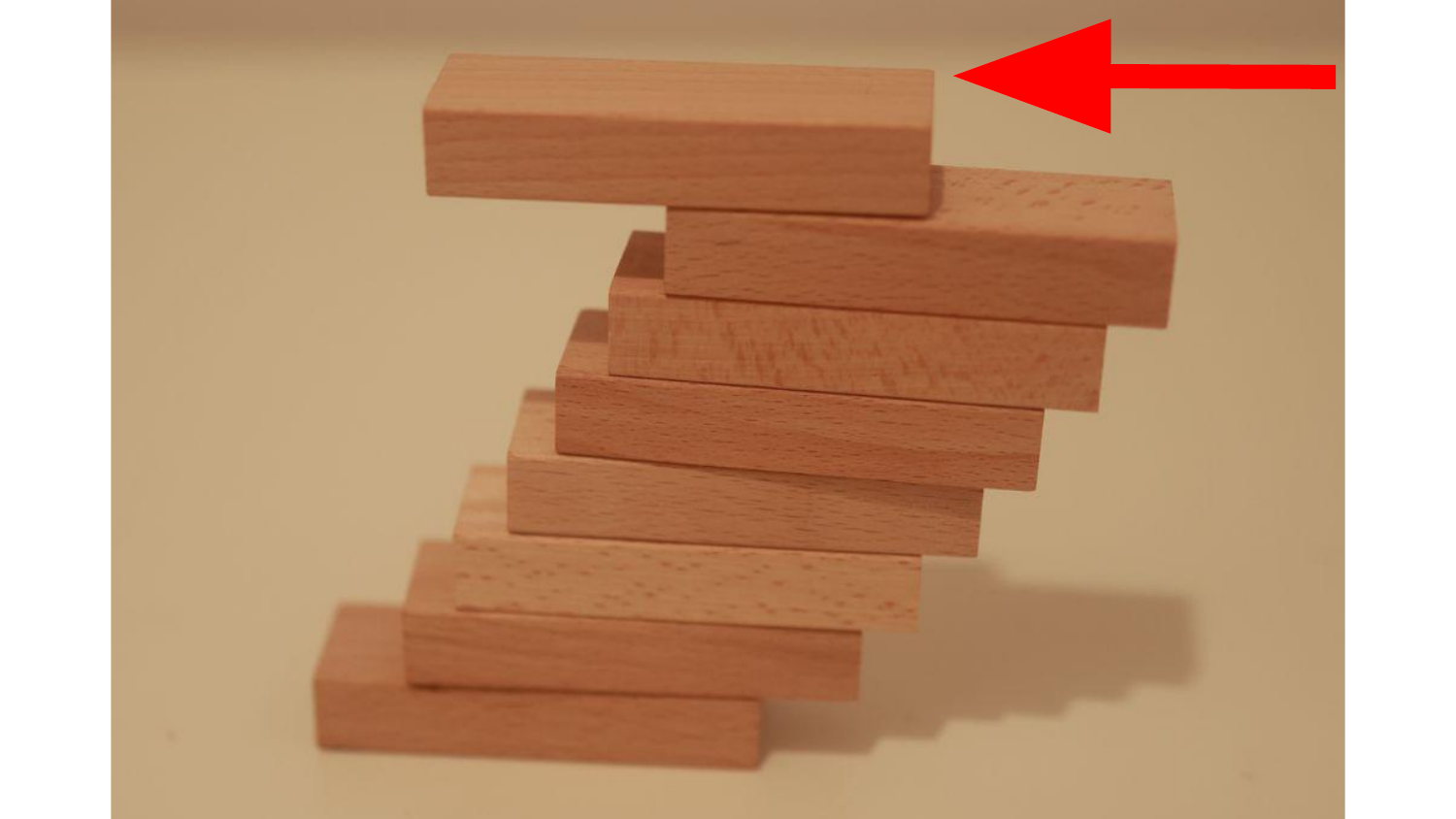}\label{fig:stack5}}\hfill
\subfigure[]{\includegraphics[width=0.165\linewidth]{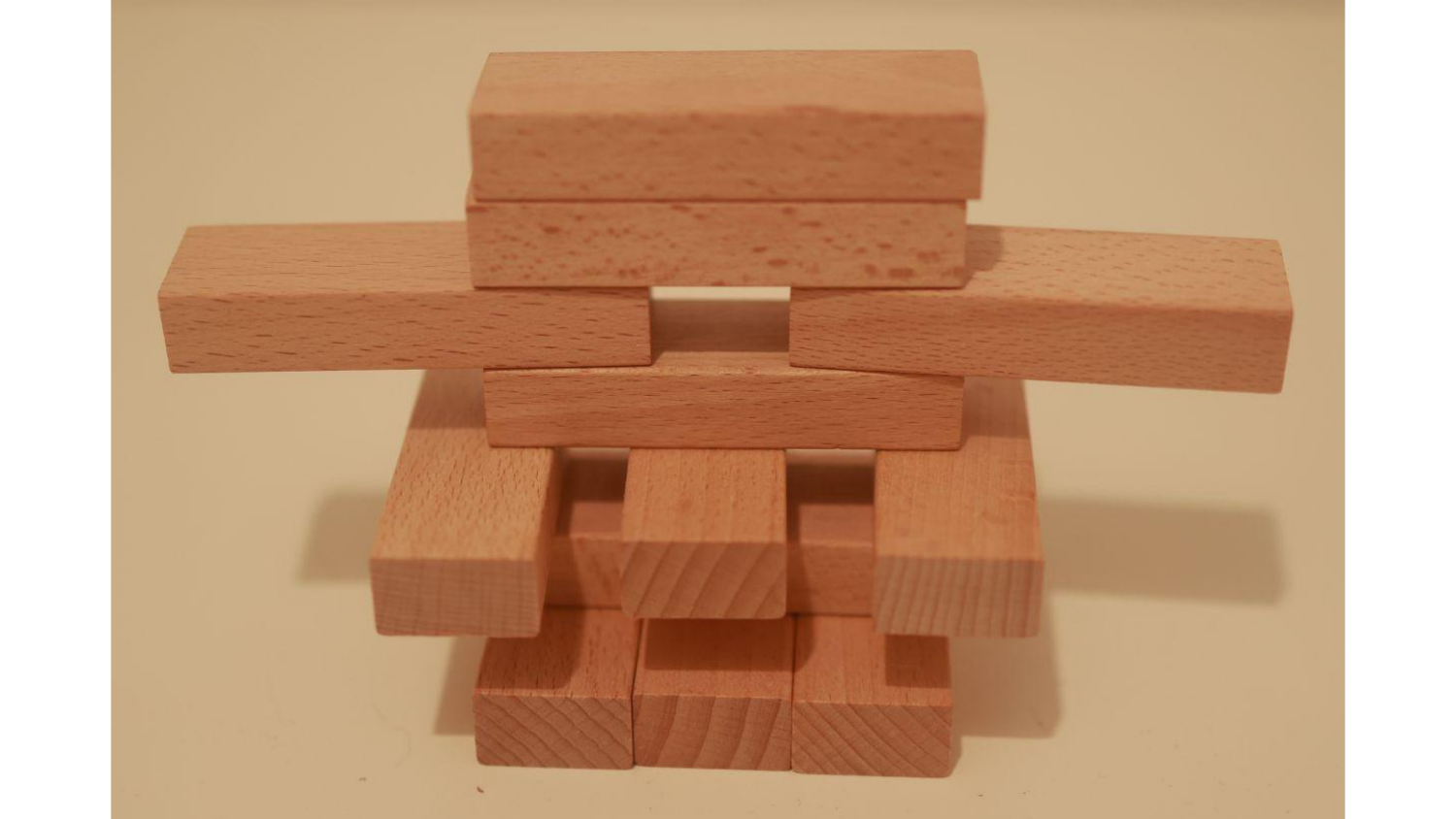}\label{fig:stack6}}
\vspace{-5pt}
    \caption{\footnotesize Stability analysis for regular block stacking. The top row 
    % of \cref{fig:regular_block_stacking}, \ie \cref{fig:stack1_analysis,fig:stack2_analysis,fig:stack3_analysis,fig:stack4_analysis,fig:stack5_analysis,fig:stack6_analysis}, 
    illustrates the stability analysis results corresponding to the structures shown in the bottom row respectively. Black: stable bricks. Red: collapsing bricks.
    % \ie \cref{fig:stack1,fig:stack2,fig:stack3,fig:stack4,fig:stack5,fig:stack6}. 
    \label{fig:regular_block_stacking}}
    \vspace{-15pt}
    % \vspace{-15pt}
\end{figure}

\paragraph{Block stacking}
The proposed formulation can be easily extended to regular block stacking assembly, in which the bricks have smooth surfaces as shown in \cref{fig:regular_block_stacking}.
We use the identical force model as shown in \cref{fig:force_model} except for removing the frictions from the Lego interlocking mechanism.
% The top row of \cref{fig:regular_block_stacking}, \ie \cref{fig:stack1_analysis,fig:stack2_analysis,fig:stack3_analysis,fig:stack4_analysis,fig:stack5_analysis,fig:stack6_analysis}, illustrates the stability analysis results corresponding to the structures shown in the bottom row, \ie \cref{fig:stack1,fig:stack2,fig:stack3,fig:stack4,fig:stack5,fig:stack6}.
In \cref{fig:stack1}, the offset between the bottom and middle bricks is larger than the offset between the middle and top bricks.
Our method predicts that the structure is stable as shown in \cref{fig:stack1_analysis}, and it is indeed stable when built in real as shown in \cref{fig:stack1}.
However, when the top brick is pushed to the right as shown in \cref{fig:stack2}, the structure collapses.
Note that the vertical brick on the right is only for the purpose of photo shooting.
Despite the slight change in the assembly, our stability analysis can reliably indicate that the structure will collapse as shown in \cref{fig:stack2_analysis}.
When we decrease the offset between adjacent bricks, we can build a structure with 8 levels as shown in \cref{fig:stack3} and our algorithm also predicts that the structure is stable as shown in \cref{fig:stack3_analysis}.
However, when the second from the bottom brick is moved slightly to the right while all the other offsets between bricks remain the same, the stability analysis indicates that the structure will collapse as shown in \cref{fig:stack4_analysis} due to the slight modification.
And \cref{fig:stack4} shows that the structure collapses as expected.
To bring the collapsing structure back to stable, our stability analysis indicates that we can just move the top brick to the left as shown in \cref{fig:stack5_analysis}.
When we build the corresponding structure, it is stable without the need of external support as shown in \cref{fig:stack5}.
\Cref{fig:stack6_analysis,fig:stack6} illustrate a random complex example with more bricks and orientations, which demonstrates that our method is applicable to regular block stacking assembly.

\begin{figure}
\centering
\subfigure[]{\includegraphics[width=0.25\linewidth]{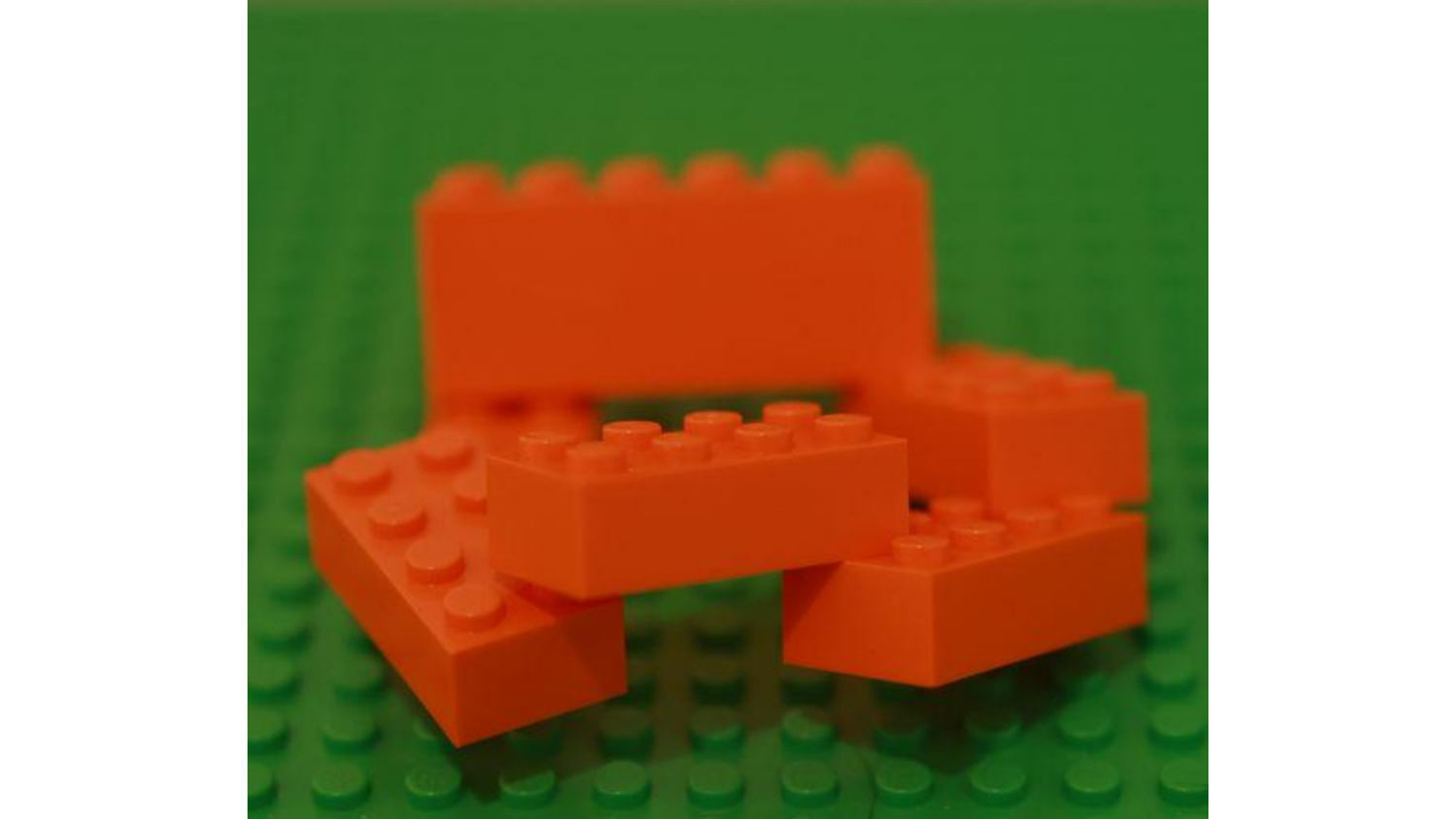}\label{fig:orilego1}}\hfill
\subfigure[]{\includegraphics[width=0.25\linewidth]{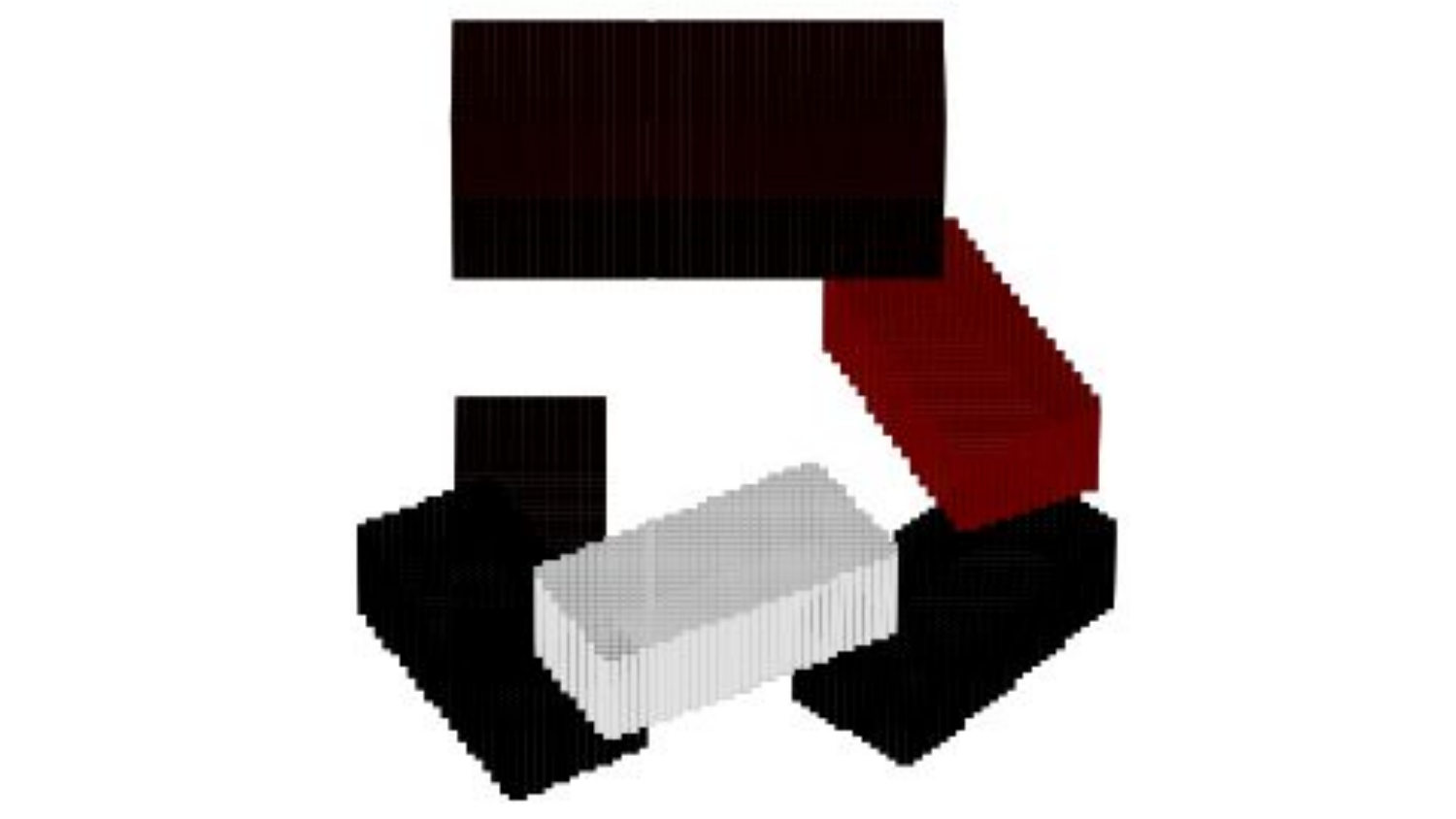}\label{fig:orilego1_analysis}}\hfill
\subfigure[]{\includegraphics[width=0.25\linewidth]{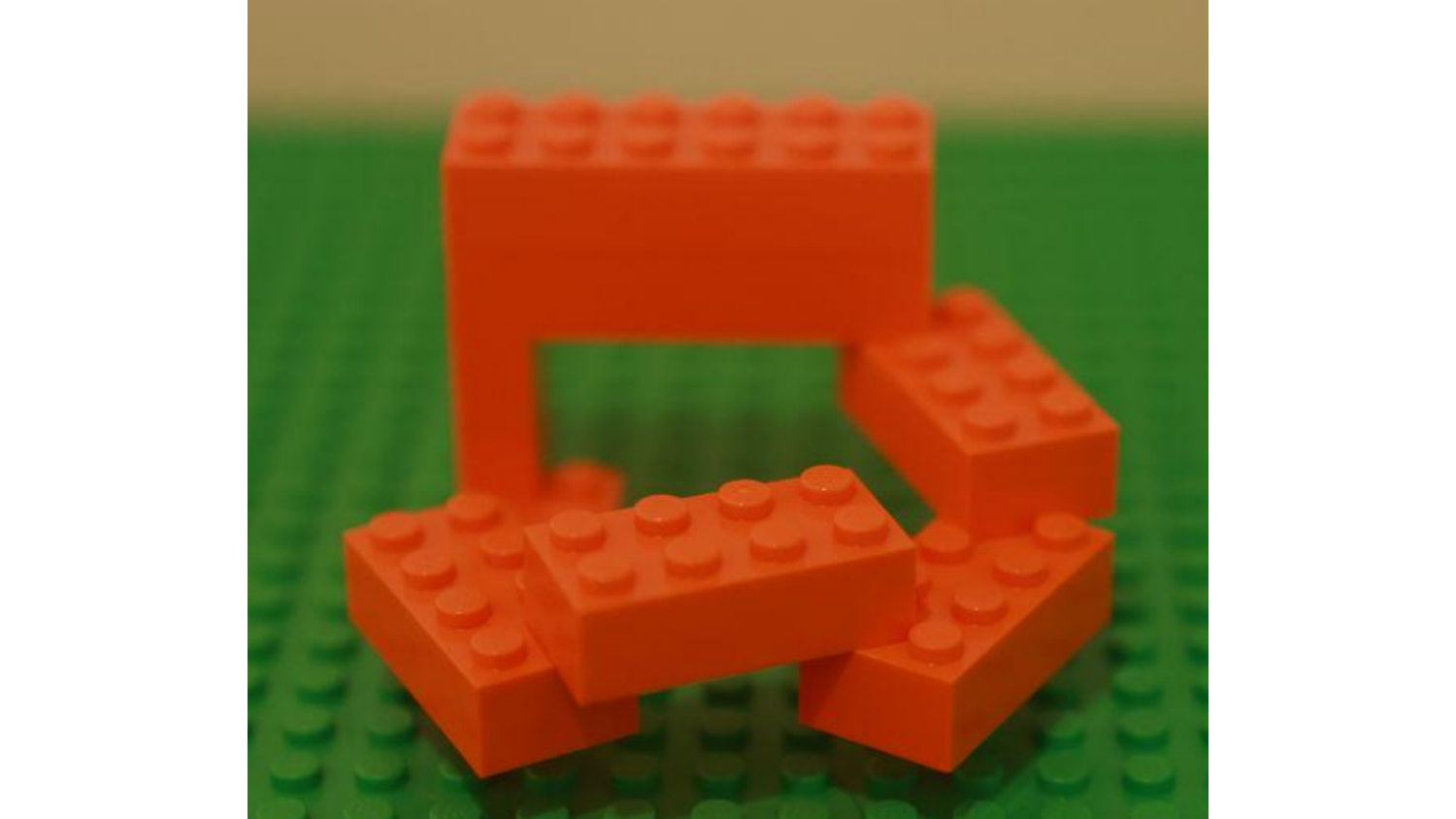}\label{fig:orilego2}}\hfill
\subfigure[]{\includegraphics[width=0.25\linewidth]{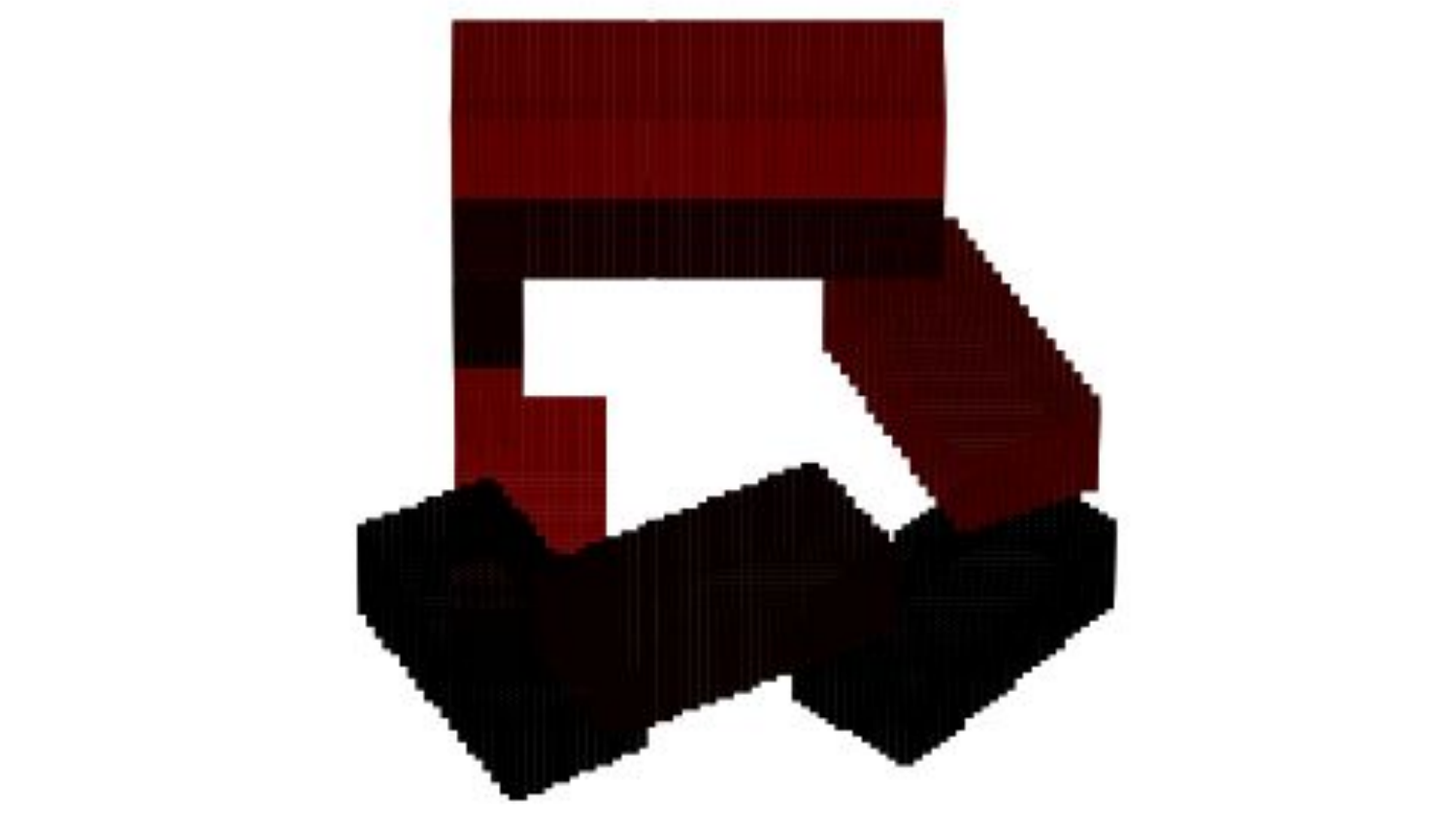}\label{fig:orilego2_analysis}}\\
\subfigure[]{\includegraphics[width=0.25\linewidth]{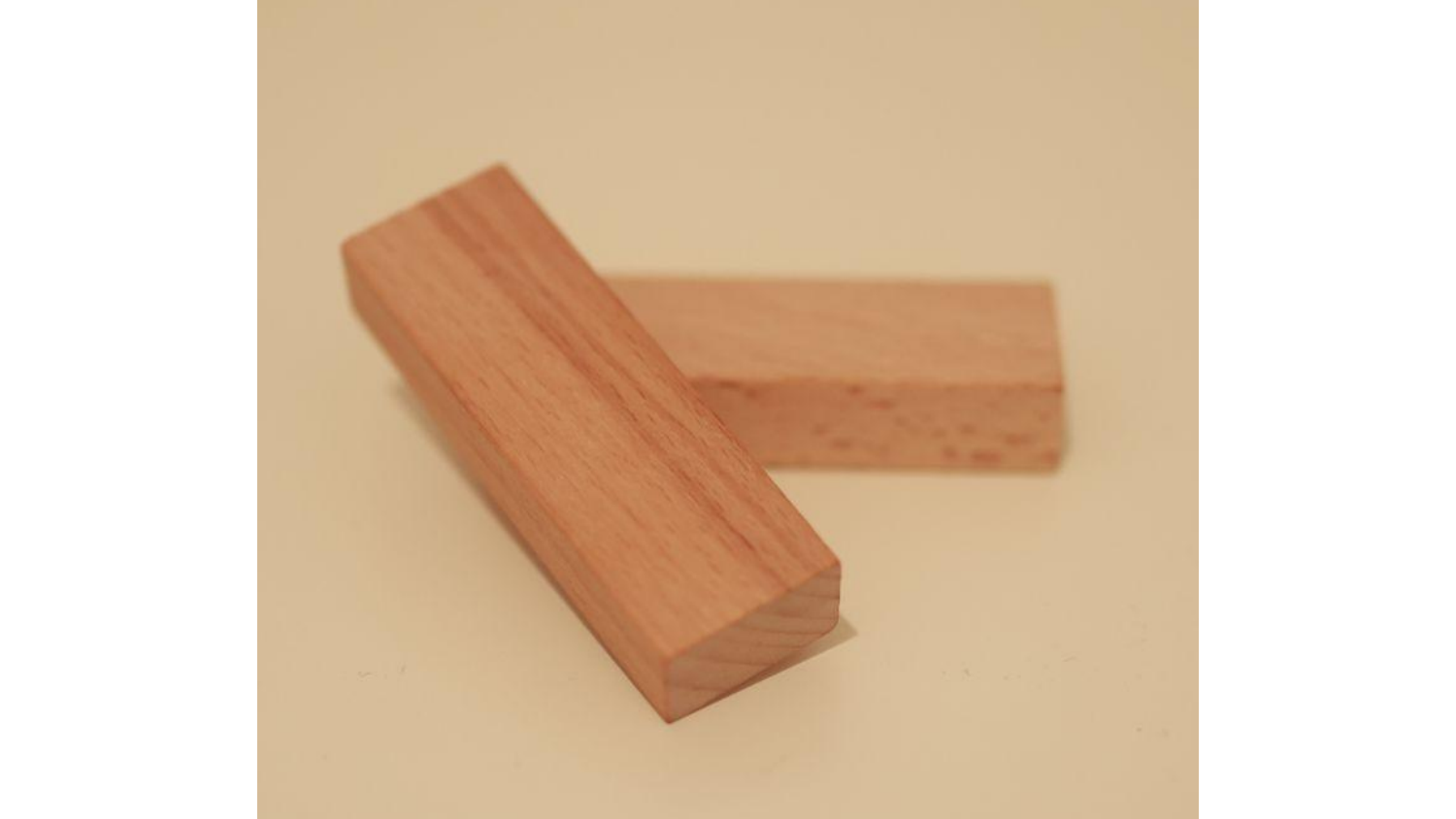}\label{fig:ori1}}\hfill
\subfigure[]{\includegraphics[width=0.25\linewidth]{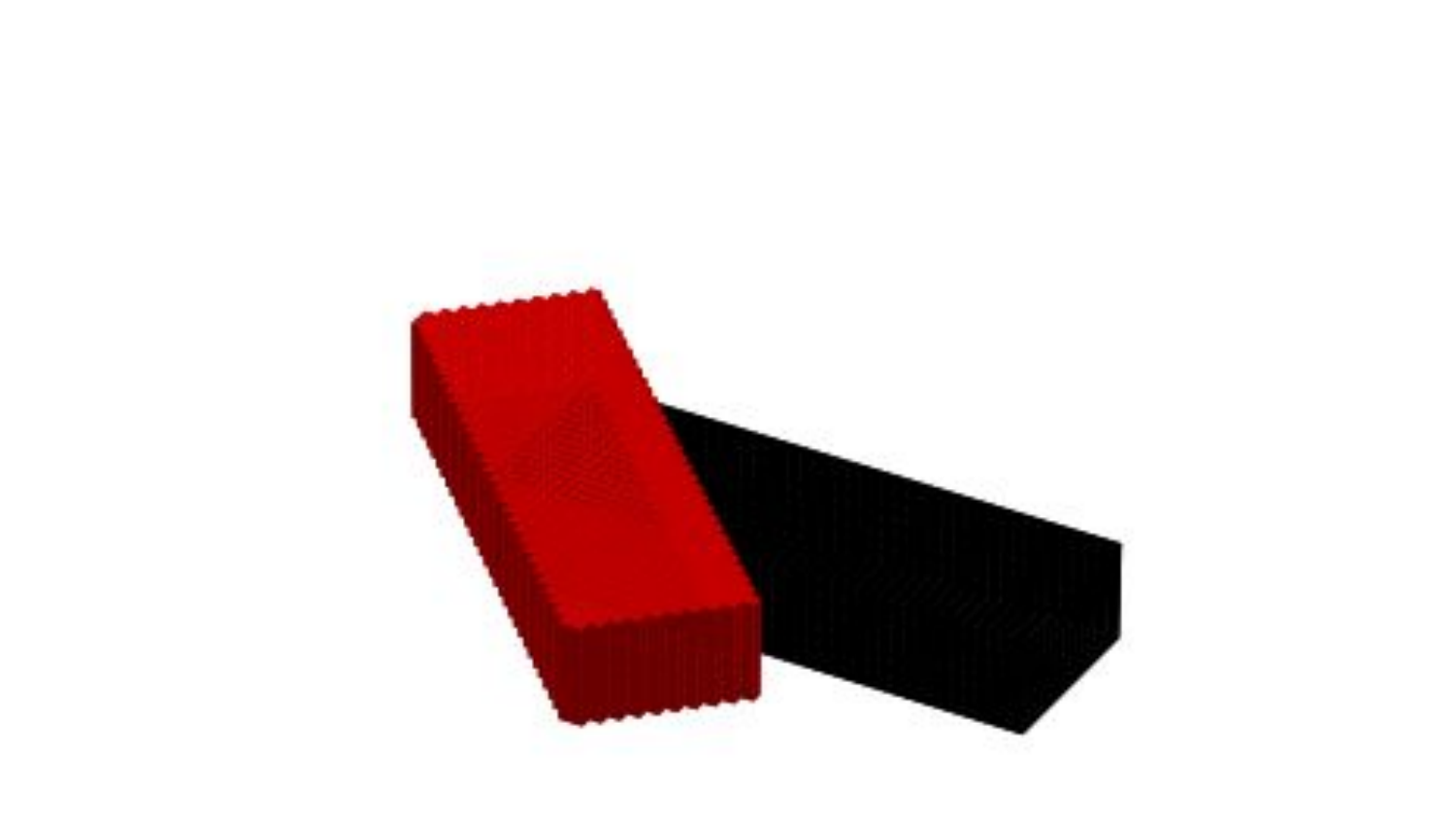}\label{fig:ori1_analysis}}\hfill
\subfigure[]{\includegraphics[width=0.25\linewidth]{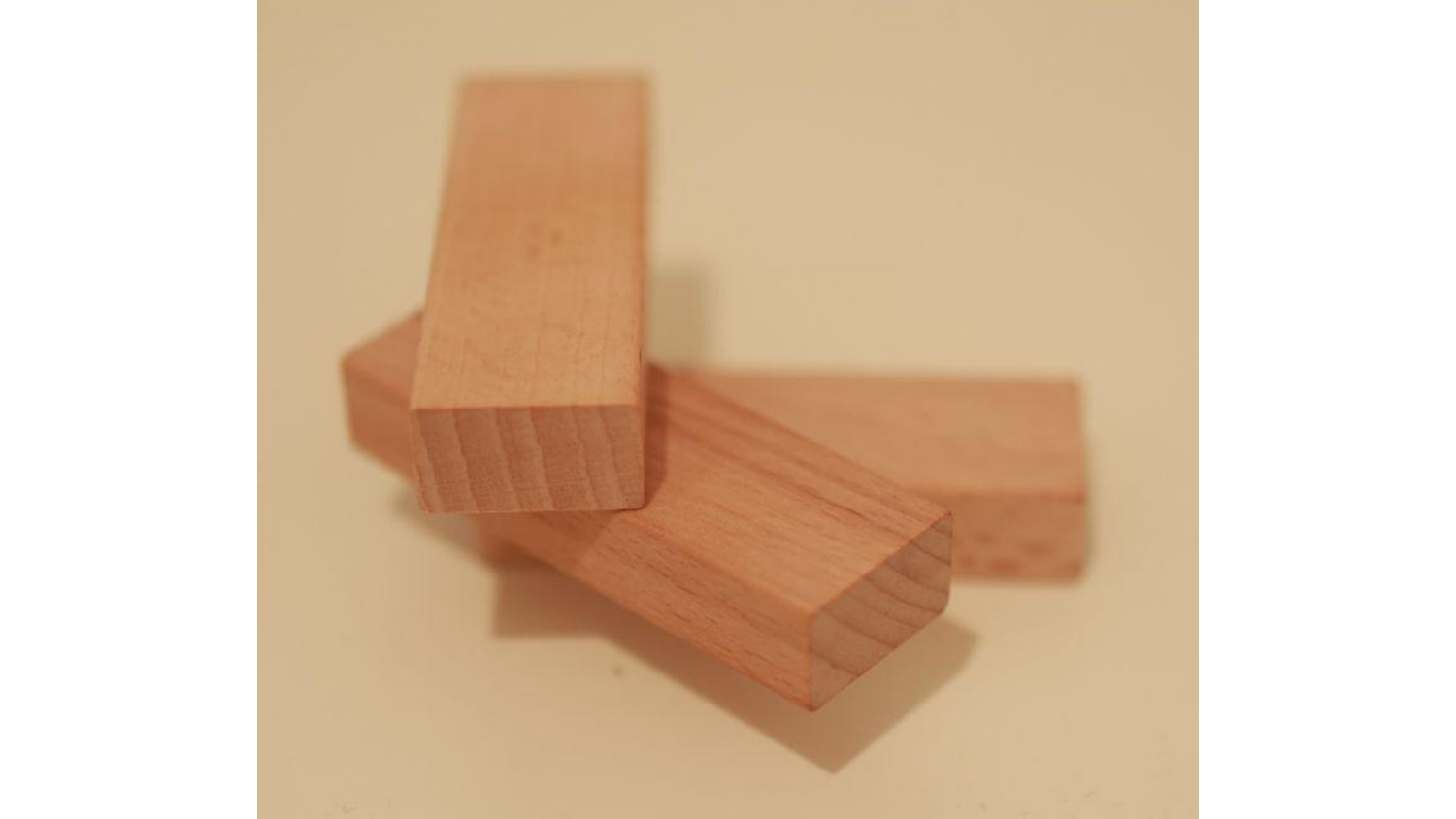}\label{fig:ori2}}\hfill
\subfigure[]{\includegraphics[width=0.25\linewidth]{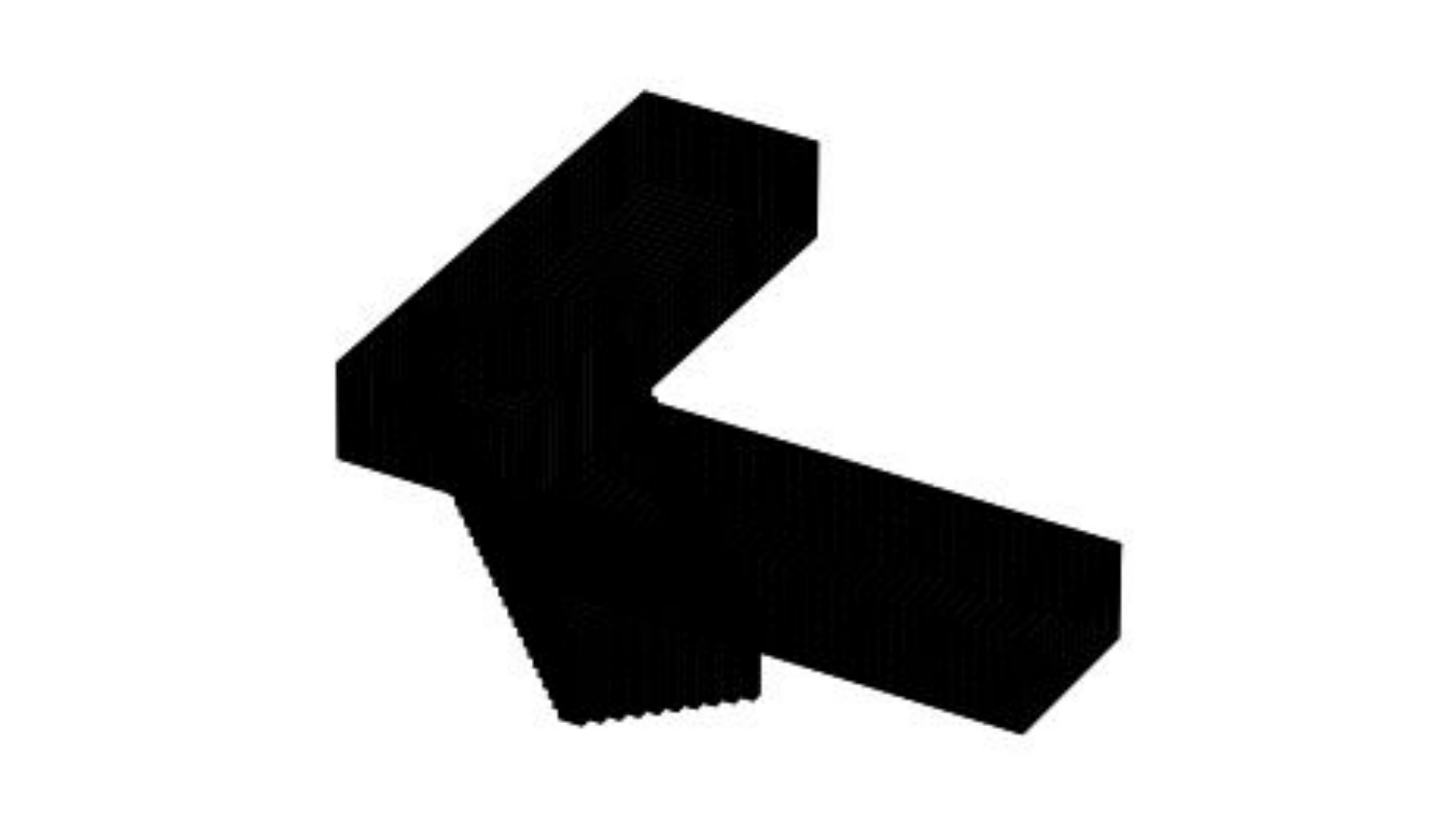}\label{fig:ori2_analysis}}
\vspace{-5pt}
    \caption{\footnotesize Stability analysis with arbitrary block orientations. Black: stable bricks. Red/White: collapsing bricks. \label{fig:arbitrary_ori}}
    \vspace{-15pt}
\end{figure}

\paragraph{Stacking with arbitrary orientation}
The proposed method can be applied to structures with arbitrarily oriented blocks.
\Cref{fig:orilego1,fig:orilego1_analysis,fig:orilego2,fig:orilego2_analysis} demonstrate applying our method to Lego structures with arbitrary orientations, in which bricks are only connected by one knob.
\Cref{fig:orilego1_analysis} correctly predicts that the structure in \cref{fig:orilego1} collapses at the predicted position, \ie the white brick in \cref{fig:orilego1_analysis}.
By adding two $1\times 2$ bricks under the $2\times 6$ brick, the structure in \cref{fig:orilego2} becomes stable, which is also indicated in our analysis as shown in \cref{fig:orilego2_analysis}.
Besides Lego, we also apply our stability analysis to regular blocks with arbitrary orientations.
The upper block in \cref{fig:ori1} has an approximately $45^{\circ}$ angle between the bottom brick.
Thus, it collapses as the stability analysis predicts as shown in \cref{fig:ori1_analysis}.
When a block is added above the collapsing block, our analysis indicates that the structure will be stable as shown in \cref{fig:ori2_analysis}.
The assembled structure is indeed stable as shown in \cref{fig:ori2}.
Therefore, we can see the proposed method is capable of solving structures with arbitrary orientation.

\begin{figure}
\centering
\subfigure[]{\includegraphics[width=0.25\linewidth]{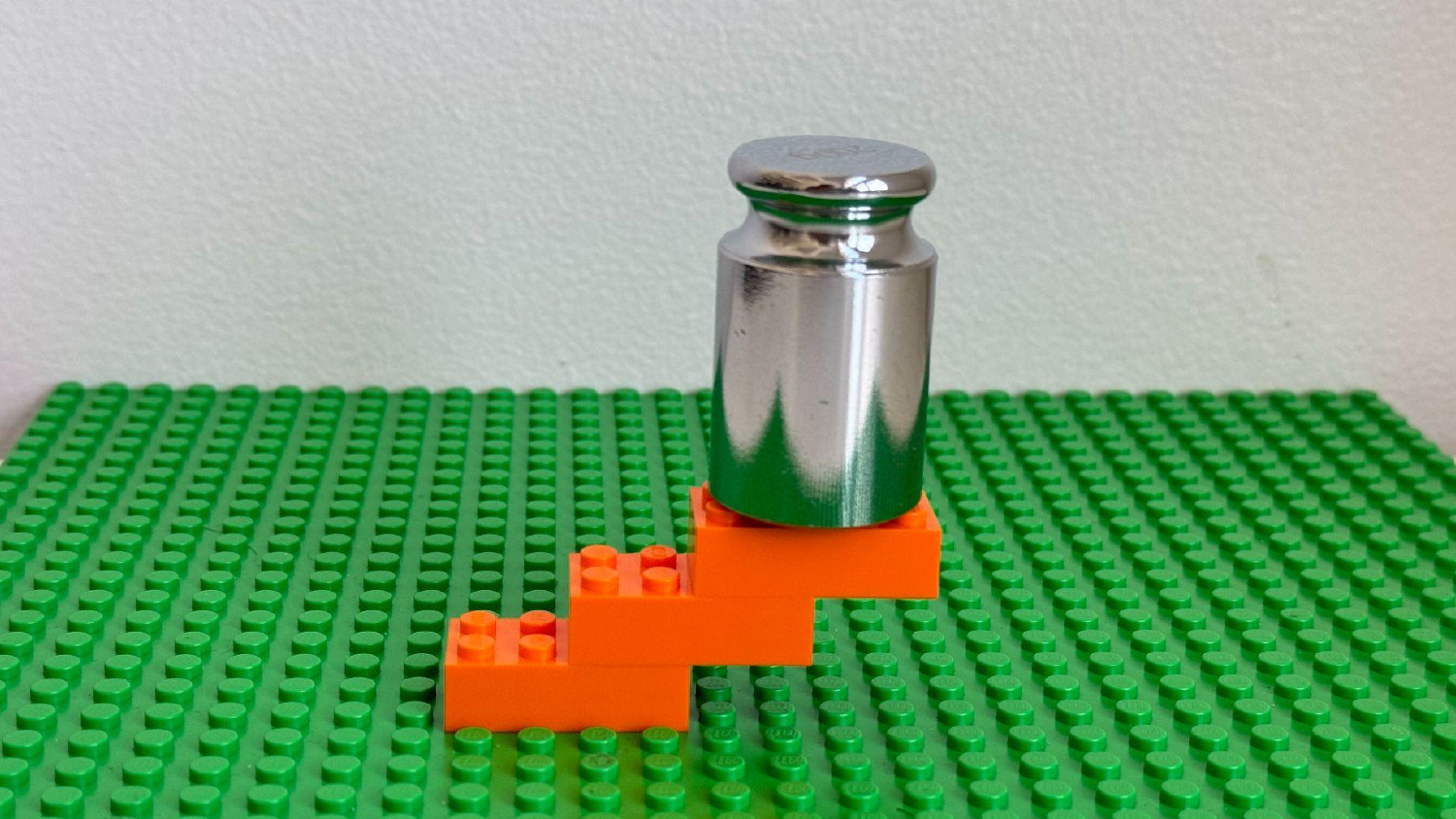}\label{fig:external_stable}}\hfill
\subfigure[]{\includegraphics[width=0.25\linewidth]{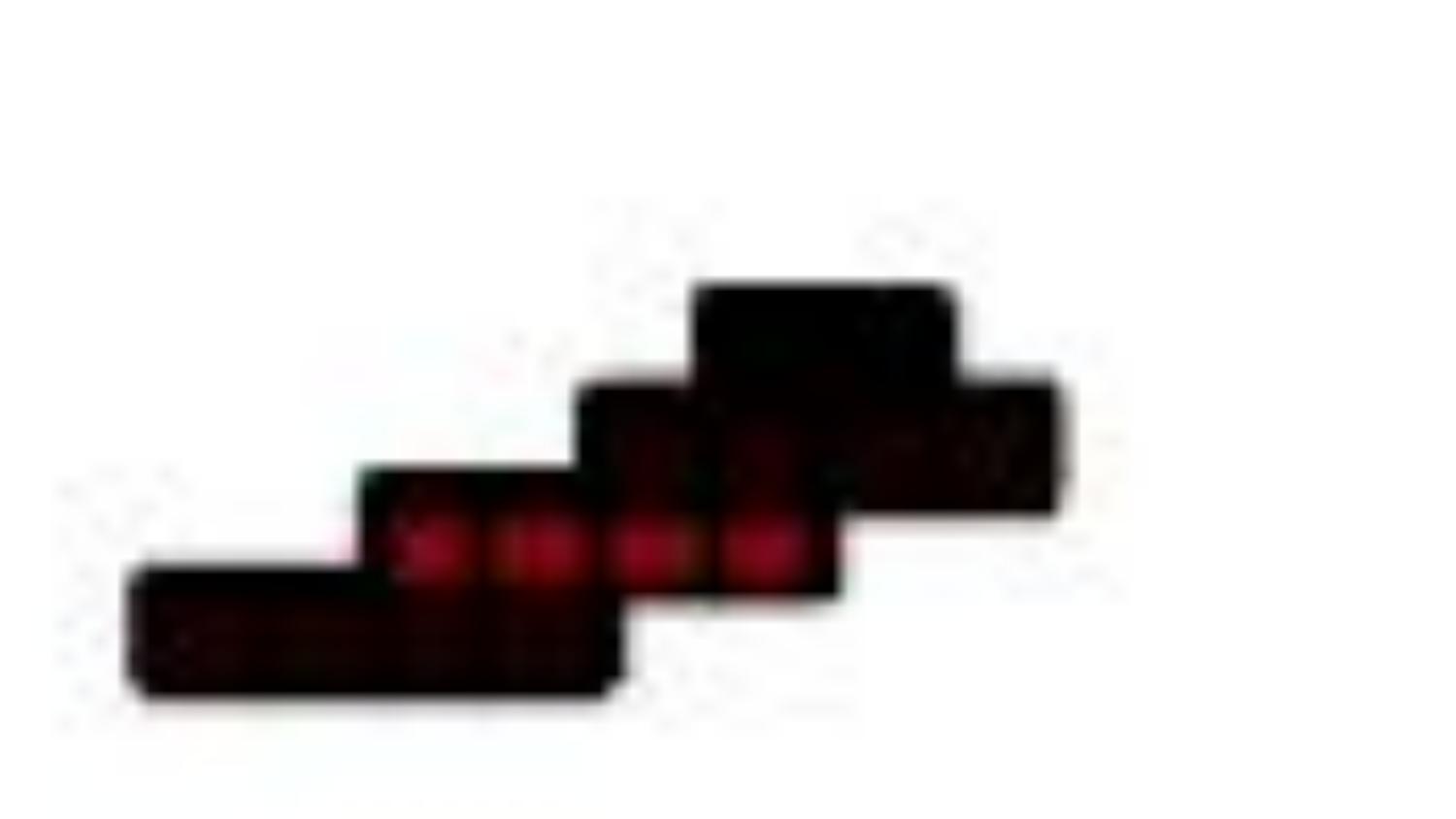}\label{fig:external_stable_analysis}}\hfill
\subfigure[]{\includegraphics[width=0.25\linewidth]{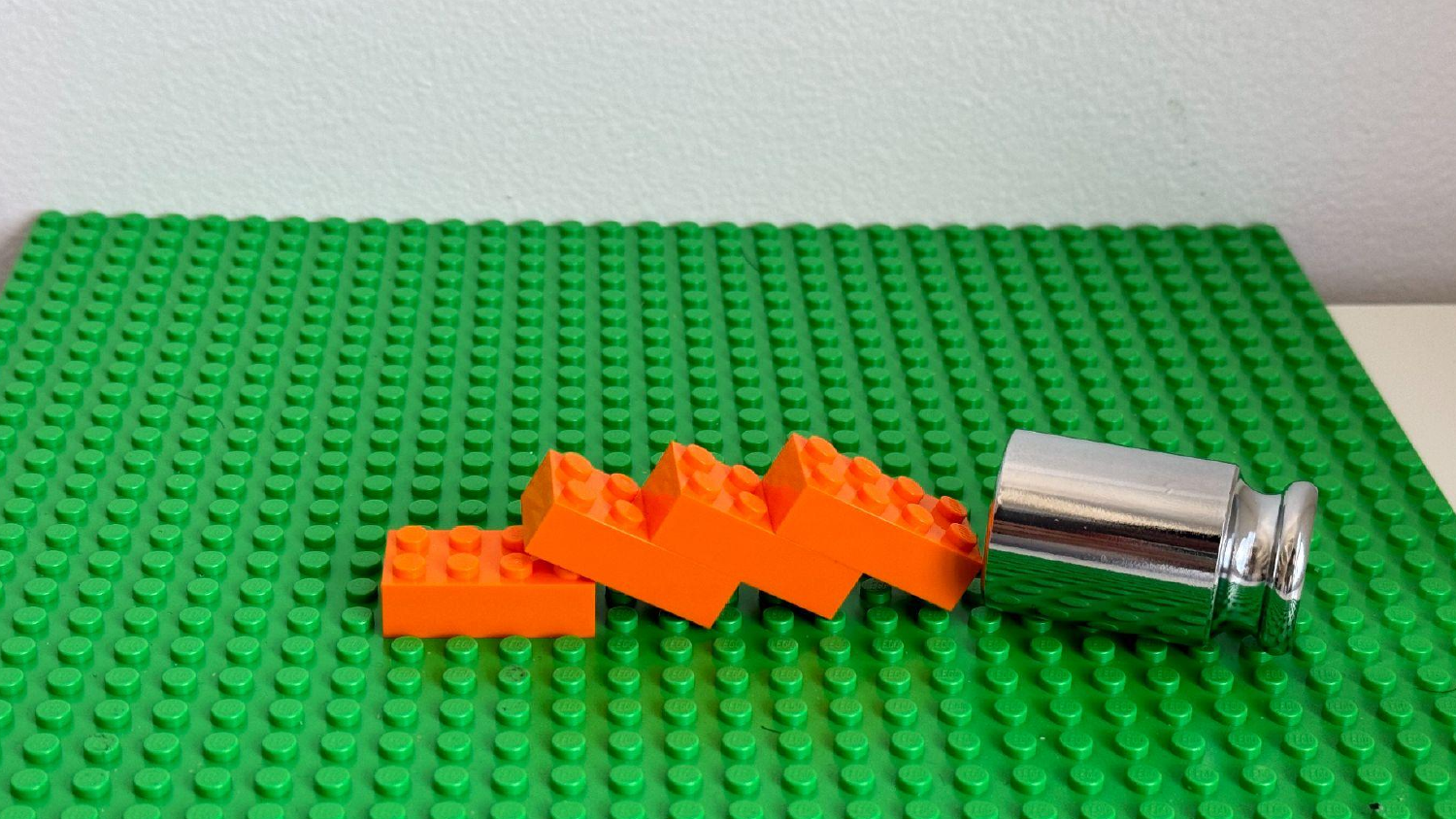}\label{fig:external_unstable}}\hfill
\subfigure[]{\includegraphics[width=0.25\linewidth]{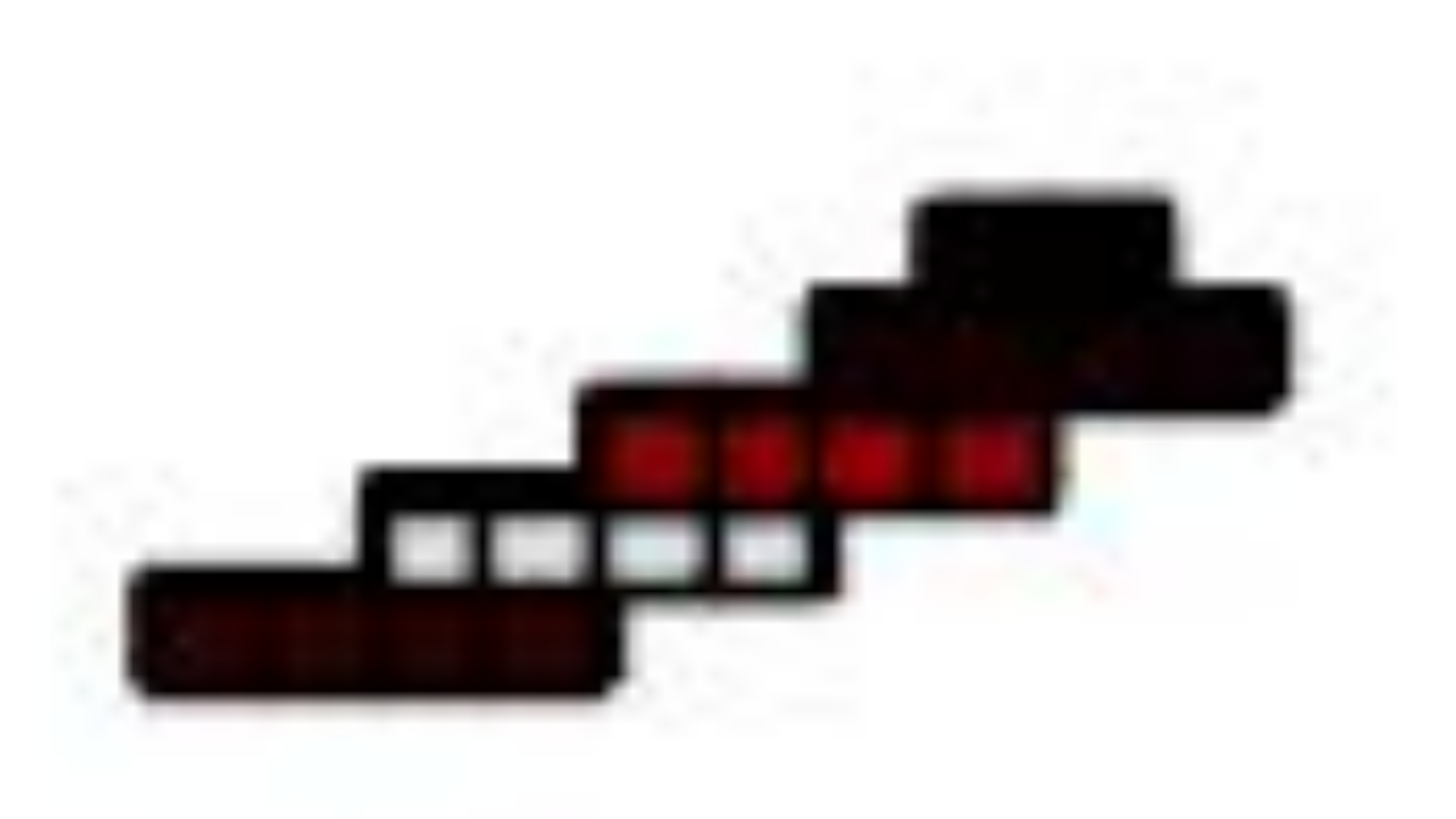}\label{fig:external_unstable_analysis}}
\vspace{-5pt}
    \caption{\footnotesize Stability analysis with external loads. Black: less internal stress. Red: higher internal stress. White: collapsing bricks.\label{fig:external_force}}
    % \vspace{-15pt}
    \vspace{-15pt}
\end{figure}

\paragraph{External weight}
The proposed stability analysis can also be extended to account for external forces by specifying large weights for specific bricks.
\Cref{fig:external_force} illustrates examples of accounting for external loads. 
A 200g weight is put on the Lego stairs. 
In the stability analysis, we use a $2\times 2$ brick with 200g to approximate the weight.
Our method indicates that a 3-level stair can support the weight (\cref{fig:external_stable_analysis}) while a 4-level stair cannot (\cref{fig:external_unstable_analysis}).
\Cref{fig:external_stable,fig:external_unstable} demonstrate the corresponding stable and unstable structures in real.
And the unstable structure indeed breaks at the predicted weakest point, \ie the white brick in \cref{fig:external_unstable_analysis}.

\begin{figure}
\centering
\subfigure[]{\includegraphics[width=0.25\linewidth]{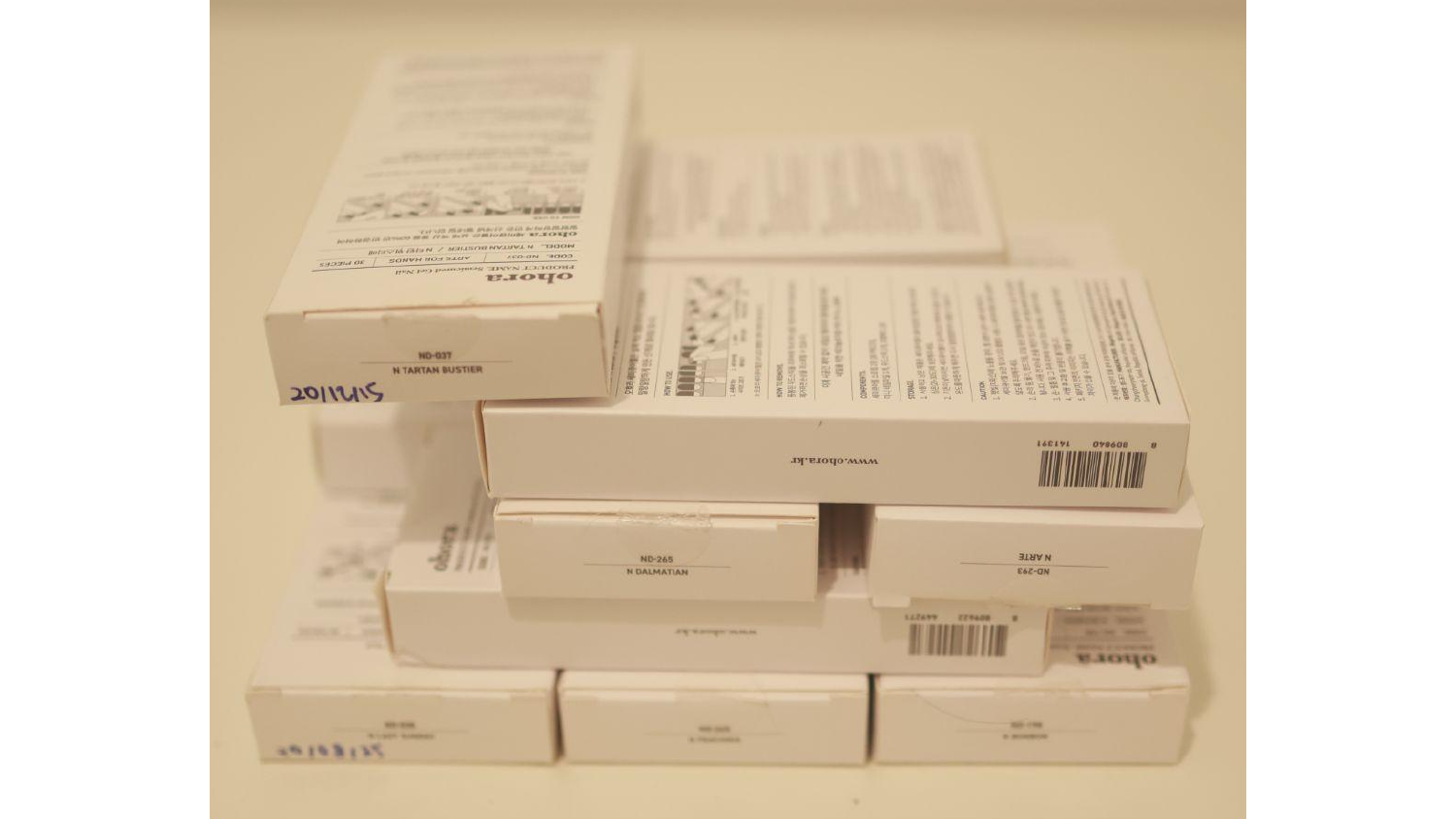}\label{fig:pallet1}}\hfill
\subfigure[]{\includegraphics[width=0.25\linewidth]{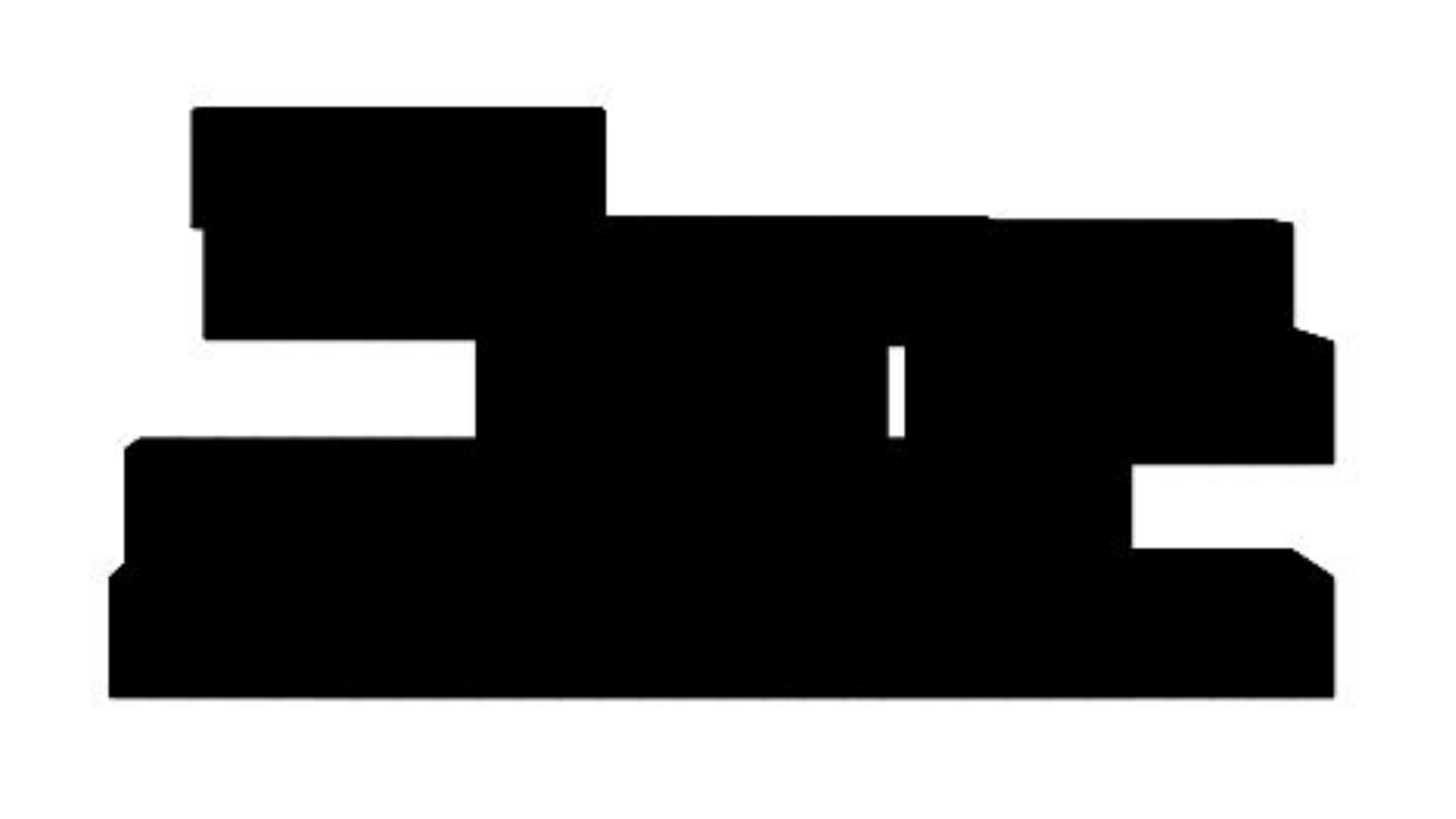}\label{fig:pallet1_analysis}}\hfill
\subfigure[]{\includegraphics[width=0.25\linewidth]{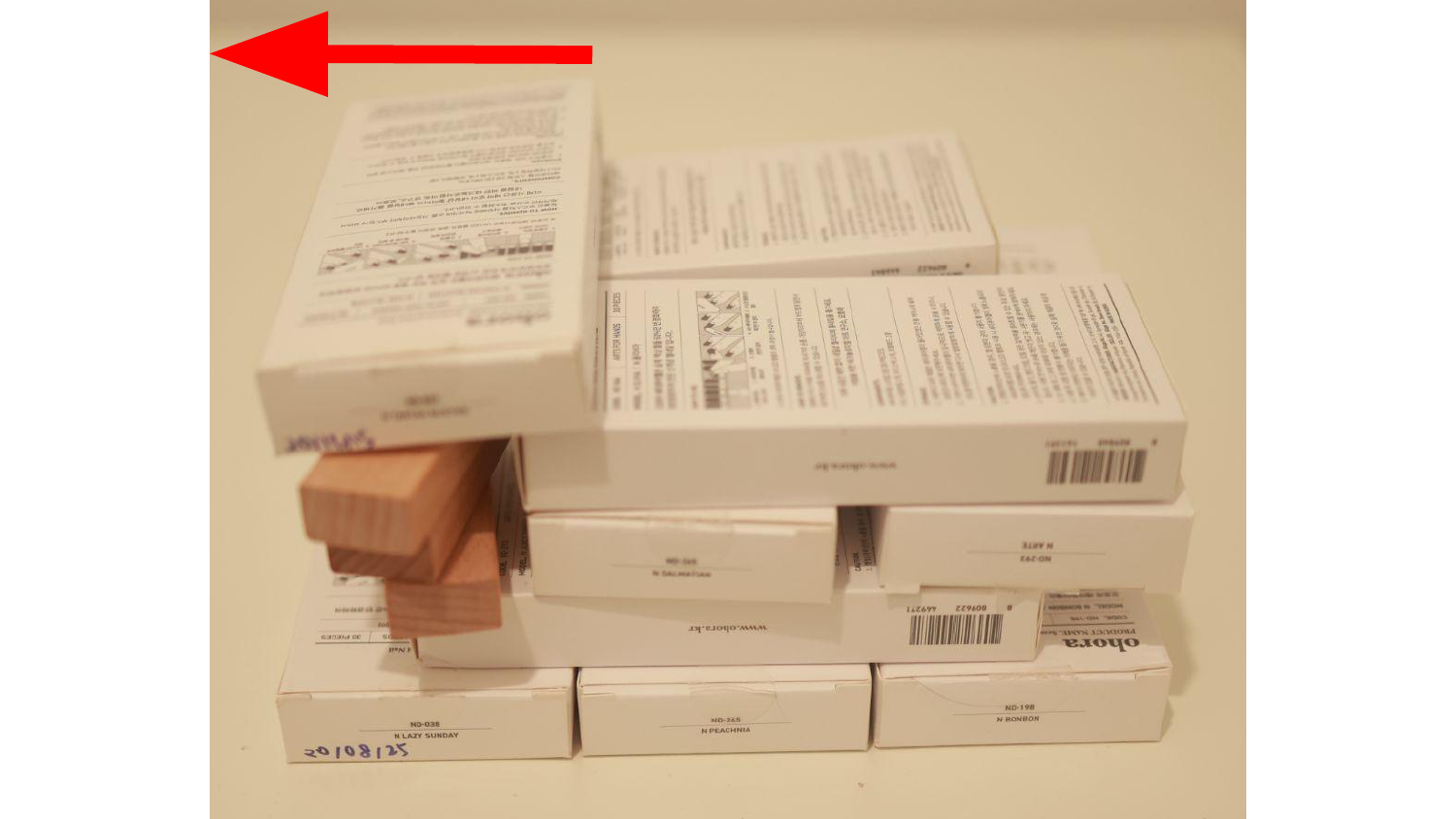}\label{fig:pallet2}}\hfill
\subfigure[]{\includegraphics[width=0.25\linewidth]{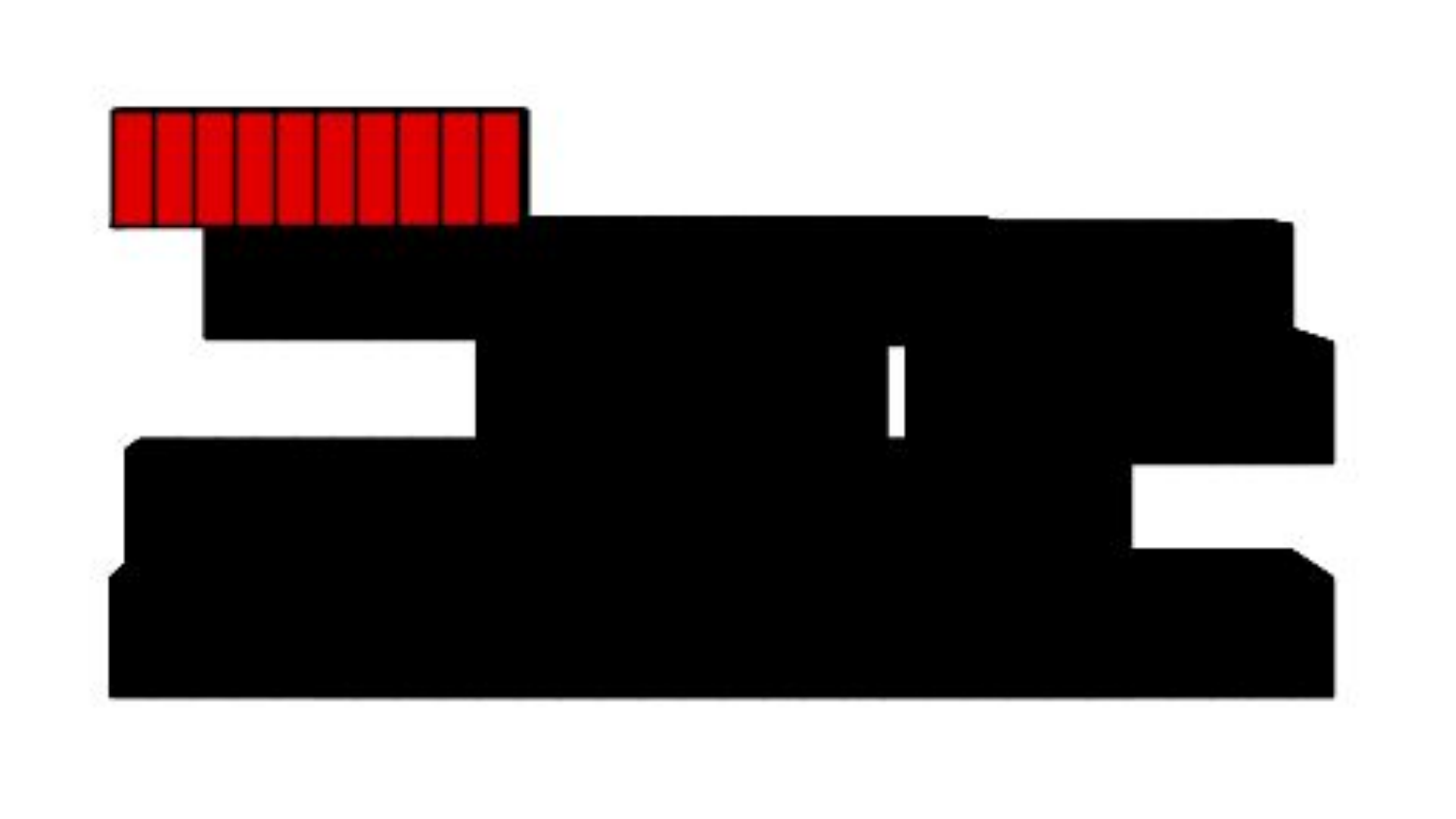}\label{fig:pallet2_analysis}}
\vspace{-5pt}
    \caption{\footnotesize Stability analysis for package palletization. Black: stable bricks. Red: collapsing bricks. \label{fig:palletization}}
    \vspace{-15pt}
\end{figure}

\paragraph{Palletization}
Package palletization is an important application in manufacturing \cite{wu2024efficient}.
\Cref{fig:palletization} illustrates that the proposed stability analysis can be applied to predict whether the pallets are stable.
\Cref{fig:pallet1,fig:pallet1_analysis} demonstrate a stable pallet of package boxes.
When slightly moving the top package to the left, the pallet will collapse as shown in \cref{fig:pallet2,fig:pallet2_analysis}.
We add blocks to support the collapsing package box for the purpose of photo shooting.
Therefore, the proposed method can be deployed to real-world applications such as palletization and manufacturing.

\subsection{Discussion}
There are several limitations to the proposed method.
First, the current implementation only considers cubic blocks, while the proposed formulation is generalizable to generic components.
Thus, we aim to improve the implementation by considering assembly components with more generic shapes.
Second, the current framework requires the user to provide the assembly configuration, which is time-consuming and prone to error.
In the future, we aim to enable the robot to inspect the structure and auto-generate the assembly configuration so that the analysis process can be automated.

Despite the limitations, there are numerous future directions that we can pursue.
Accounting for external forces can be helpful when planning dual-arm Lego assembly using the manipulation strategy in \cite{liu2023lightweight}.
The stability analysis can indicate \textit{whether} and \textit{where} a supporting arm is needed.
In addition, we can use our approach to efficiently guide generative AI (as shown in \cref{fig:generative_ai,fig:generative_ai_analysis}) to improve the imperfect design.
Moreover, we aim to integrate dynamic forces into the optimization, which enables the robot to understand the impact of its action to the world and potentially outperform in complex manipulation tasks, \eg Jenga extraction \cite{doi:10.1126/scirobotics.aav3123}.

% Moreover, the proposed stability analysis can be integrated into sequence planning algorithms to generate a feasible assembly plan so that all the intermediate steps are physically valid.

\section{Conclusion}

This paper studies the structural stability of block stacking structures.
In particular, this paper leverages the RBE method and proposes a new optimization formulation,
which optimizes over force balancing equations, for inferring
the stability of 3D structures. 
To benchmark the performance, we provide StableLego: a dataset of 3D objects with their Lego layouts and the corresponding stability inferences. 
The dataset includes a wide variety of assembly configurations (\ie more than 50k structures) using standardized Lego bricks. 
The proposed formulation is verified using 1) hand-crafted Lego designs, 2) the StableLego dataset, and 3) regular building blocks.

\bibliographystyle{IEEEtran}
\bibliography{bibliography}      
\end{document}

%% file: plot/time_plot.tex

%
\begin{tikzpicture}
\definecolor{mycolor1}{rgb}{0.00000,0.0,1}%
\definecolor{mycolor2}{rgb}{1, 0, 0}%
\definecolor{mycolor3}{rgb}{0.92900,0.69400,0.12500}%
\definecolor{mycolor4}{rgb}{0.49400,0.18400,0.55600}%
\pgfplotsset{compat=1.5}
\begin{groupplot}[group style={group size=1 by 2,horizontal sep=20pt, vertical sep=35pt}]
\nextgroupplot[
width = 9cm,
height = 3cm,
font = \footnotesize,
% xmode=log,
legend cell align={left},
legend style={nodes={scale=0.58, transform shape}, fill opacity=0.8, draw opacity=1, text opacity=1, draw=white!80!black},
tick align=inside,
tick pos=left,
x grid style={white!69.0196078431373!black},
xmin=0, xmax=1000,
xtick style={color=black},
y grid style={white!69.0196078431373!black},
ylabel={Time(s)},
% ymode=log,
ymin=0,
ymax=1.5,
ytick style={color=black},
title = {\textbf{Unit Brick ($1\times 1$) Structure Computation Time}}
]
\addplot [very thick, color=mycolor1]
  table{%
50 0.00631248950958252
100 0.01557457447052
150 0.0285160541534424
200 0.0448329448699951
250 0.0584995746612549
300 0.076019287109375
350 0.0950822830200195
400 0.122810363769531
450 0.146926522254944
500 0.16923189163208
550 0.200401544570923
600 0.227420806884766
650 0.262317180633545
700 0.289595484733582
750 0.332412481307983
800 0.350093245506287
850 0.413101315498352
900 0.451053857803345
950 0.530991077423096
1000 0.546321630477905
1050 0.593316078186035
};
% \addlegendentry{$50$ Percentile}

\addplot [dashed, very thick, color=mycolor1]
  table{%
50 0.0581138134002686
100 0.0945707559585571
150 0.139809131622314
200 0.189546823501587
250 0.24744987487793
300 0.306450843811035
350 0.361968636512756
400 0.430717945098877
450 0.489970803260803
500 0.544354677200317
550 0.631056785583496
600 0.684999704360962
650 0.764039278030396
700 0.816408276557922
750 0.885432481765747
800 0.942053556442261
850 1.04051506519318
900 1.14725756645203
950 1.24711871147156
1000 1.30094075202942
1050 1.37839150428772
};
% \addlegendentry{$50$ Percentile}

\addplot [very thick, color=mycolor2]
  table{%
50 0.004158616065979
100 0.0127078294754028
150 0.0234034061431885
200 0.0379014611244202
250 0.0504372119903564
300 0.06208336353302
350 0.0740261077880859
400 0.0917527675628662
450 0.109533369541168
500 0.127002954483032
550 0.150264382362366
600 0.172934532165527
650 0.196953594684601
700 0.21637237071991
750 0.24118447303772
800 0.266403079032898
850 0.316877245903015
900 0.368041753768921
950 0.372544646263123
1000 0.406235694885254
1050 0.452087640762329
};
% \addlegendentry{$25$ Percentile}

\addplot [dashed, very thick, color=mycolor2]
  table{%
50 0.0505334734916687
100 0.0837504863739014
150 0.125482320785522
200 0.166907429695129
250 0.216639280319214
300 0.258181273937225
350 0.32032322883606
400 0.383080959320068
450 0.437840640544891
500 0.492494940757751
550 0.552129030227661
600 0.624895930290222
650 0.687521636486053
700 0.738727927207947
750 0.792122602462769
800 0.850027680397034
850 0.942563951015472
900 1.01035523414612
950 1.0491783618927
1000 1.14946937561035
1050 1.219322681427
};
% \addlegendentry{$25$ Percentile}

\addplot [very thick, color=mycolor3]
  table{%
50 0.00834363698959351
100 0.0197643041610718
150 0.0355546474456787
200 0.0545567274093628
250 0.0719952583312988
300 0.097663402557373
350 0.117933869361877
400 0.151631116867065
450 0.183130264282227
500 0.218772530555725
550 0.25903308391571
600 0.290729522705078
650 0.326954483985901
700 0.358808755874634
750 0.403199434280396
800 0.443490743637085
850 0.512248814105988
900 0.559995174407959
950 0.593779683113098
1000 0.633782625198364
1050 0.686394214630127
};

\addplot [dashed, very thick, color=mycolor3]
  table{%
50 0.0661818981170654
100 0.109899580478668
150 0.159783124923706
200 0.228971123695374
250 0.286486625671387
300 0.346527993679047
350 0.397938311100006
400 0.472303152084351
450 0.533262252807617
500 0.605176687240601
550 0.689528107643127
600 0.75951087474823
650 0.833701848983765
700 0.89275997877121
750 0.962834119796753
800 1.04167723655701
850 1.17465949058533
900 1.23808646202087
950 1.3066531419754
1000 1.40555167198181
1050 1.4858067035675
};

\nextgroupplot[
width = 9cm,
height = 3cm,
font = \footnotesize,
% xmode=log,
legend cell align={left},
legend style={nodes={scale=0.58, transform shape}, fill opacity=0.8, draw opacity=1, text opacity=1, draw=white!80!black},
tick align=inside,
tick pos=left,
x grid style={white!69.0196078431373!black},
xmin=0, xmax=600,
xtick style={color=black},
y grid style={white!69.0196078431373!black},
ylabel={Time(s)},
% ymode=log,
ymin=0,
ymax=6,
ytick style={color=black},
xlabel={Number of Bricks},
title = {\textbf{StableLego Computation Time}}
]
\addplot [very thick, color=mycolor1]
  table{%
50 0.0100762844085693
100 0.0359817743301392
150 0.108963489532471
200 0.208039045333862
250 0.349682092666626
300 0.536070346832275
350 0.814275741577148
400 1.06191396713257
450 1.43914580345154
500 1.51359820365906
550 2.04006338119507
600 2.53019499778748
650 3.06352257728577
700 3.69459176063538
750 4.41786634922028
800 5.37392425537109
850 5.83497381210327
900 7.10917973518372
950 11.0502586364746
1000 2.1585054397583
1050 7.52259111404419
1100 3.27371382713318
1150 3.65926778316498
1200 1.91187858581543
};
% \addlegendentry{$50$ Percentile}

\addplot [dashed, very thick, color=mycolor1]
  table{%
50 0.0482096672058105
100 0.115421772003174
150 0.249437093734741
200 0.412884950637817
250 0.632223606109619
300 0.875682353973389
350 1.23173689842224
400 1.52679371833801
450 1.98162996768951
500 2.15710711479187
550 2.7184431552887
600 3.30811953544617
650 3.92198371887207
700 4.55862641334534
750 5.3148068189621
800 6.53522562980652
850 6.70547485351562
900 8.69550037384033
950 12.1176538467407
1000 3.48441600799561
1050 8.40163898468018
1100 4.43033361434937
1150 5.08627820014954
1200 3.41803860664368
};

\addplot [very thick, color=mycolor2]
  table{%
50 0.00658702850341797
100 0.0219574570655823
150 0.0622954368591309
200 0.110219717025757
250 0.187815308570862
300 0.281978130340576
350 0.385824918746948
400 0.5465247631073
450 0.628765940666199
500 0.762811183929443
550 0.99198043346405
600 1.18716323375702
650 1.71220278739929
700 1.64858722686768
750 1.98041951656342
800 1.89695596694946
850 2.95524156093597
900 3.22519755363464
950 4.94199705123901
1000 1.62112855911255
1050 6.30975556373596
1100 3.27371382713318
1150 3.6393169760704
1200 1.91187858581543
};
% \addlegendentry{$25$ Percentile}

\addplot [dashed, very thick, color=mycolor2]
  table{%
50 0.0385191440582275
100 0.0783628225326538
150 0.185591578483582
200 0.309904813766479
250 0.466704368591309
300 0.647763133049011
350 0.871321201324463
400 1.05096507072449
450 1.25936502218246
500 1.49457931518555
550 1.73711097240448
600 1.99559092521667
650 2.60288643836975
700 2.6929258108139
750 3.04590201377869
800 2.86864972114563
850 4.32361268997192
900 4.37676787376404
950 6.00429224967957
1000 2.97517317533493
1050 7.44742488861084
1100 4.43033361434937
1150 5.06865644454956
1200 3.41803860664368
};

\addplot [very thick, color=mycolor3]
  table{%
50 0.0156706571578979
100 0.0632146596908569
150 0.186530470848083
200 0.40823769569397
250 0.809308171272278
300 1.34615409374237
350 2.11177659034729
400 2.63889026641846
450 3.77367627620697
500 4.84383082389832
550 4.88434445858002
600 4.9787272810936
650 5.01839852333069
700 6.56611764431
750 6.02414071559906
800 8.94912815093994
850 8.2930788397789
900 9.8484160900116
950 27.440812587738
1000 2.82062864303589
1050 9.25816011428833
1100 3.27371382713318
1150 3.67921859025955
1200 1.91187858581543
};
% \addlegendentry{$75$ Percentile}

\addplot [dashed, very thick, color=mycolor3]
  table{%
50 0.062034010887146
100 0.162479937076569
150 0.330980777740479
200 0.60413384437561
250 1.07721889019012
300 1.65205216407776
350 2.4941291809082
400 3.14029026031494
450 4.26051646471024
500 5.38888955116272
550 5.60273051261902
600 5.8156498670578
650 5.81660747528076
700 7.34256649017334
750 7.13943898677826
800 9.90055561065674
850 9.28544747829437
900 10.874427318573
950 28.6694693565369
1000 4.08304506540298
1050 10.2136080265045
1100 4.43033361434937
1150 5.10389995574951
1200 3.41803860664368
};

\end{groupplot}
\end{tikzpicture}%